\documentclass{article}

\usepackage{cua_gym_arxiv}

\usepackage[utf8]{inputenc} % allow utf-8 input
\usepackage[T1]{fontenc}    % use 8-bit T1 fonts
\usepackage{hyperref}       % hyperlinks
\usepackage{url}            % simple URL typesetting
\usepackage{booktabs}       % professional-quality tables
\usepackage{amsfonts}       % blackboard math symbols
\usepackage{nicefrac}       % compact symbols for 1/2, etc.
\usepackage{microtype}      % microtypography
\usepackage{xcolor}         % colors
\usepackage{subcaption}
\usepackage{graphicx} 
\usepackage{wrapfig}

\usepackage{amsmath}
\usepackage{etoc}
\usepackage{algorithm}
\usepackage{algpseudocode}
\usepackage{longtable}
\usepackage{listings}
\usepackage[most]{tcolorbox}
\tcbuselibrary{listings,breakable,skins}

\definecolor{promptheaderbg}{HTML}{2C3E50}
\definecolor{promptbodybg}{HTML}{F7F5F0}
\definecolor{codeheaderbg}{HTML}{37352F}
\definecolor{codebodybg}{HTML}{F7F6F3}
\definecolor{codeKwd}{HTML}{C9211E}
\definecolor{codeStr}{HTML}{067D17}
\definecolor{codeCom}{HTML}{6A737D}
\definecolor{codeNum}{HTML}{1750EB}

\newtcblisting{promptlisting}[1]{%
    enhanced, breakable, sharp corners,
    listing only,
    boxrule=0pt,
    colback=promptbodybg,
    colbacktitle=promptheaderbg,
    coltitle=white,
    fonttitle=\bfseries\sffamily,
    title={#1},
    listing options={
        basicstyle=\ttfamily\footnotesize,
        breaklines=true,
        breakatwhitespace=false,
        columns=fullflexible,
        keepspaces=true,
        showspaces=false,
        showstringspaces=false,
        upquote=true,
    },
    left=10pt, right=10pt, top=8pt, bottom=8pt,
    before skip=10pt, after skip=10pt,
}

\lstdefinestyle{cuapython}{
    language=Python,
    basicstyle=\ttfamily\footnotesize,
    keywordstyle=\color{codeKwd}\bfseries,
    stringstyle=\color{codeStr},
    commentstyle=\color{codeCom}\itshape,
    numberstyle=\color{codeNum},
    breaklines=true,
    breakatwhitespace=false,
    columns=fullflexible,
    keepspaces=true,
    showspaces=false,
    showstringspaces=false,
    upquote=true,
}

\newtcblisting{pythonlisting}[1]{%
    enhanced, breakable, sharp corners,
    listing only,
    boxrule=0pt,
    colback=codebodybg,
    colbacktitle=codeheaderbg,
    coltitle=white,
    fonttitle=\bfseries\sffamily,
    title={#1},
    listing options={style=cuapython},
    left=10pt, right=10pt, top=8pt, bottom=8pt,
    before skip=10pt, after skip=10pt,
}

\definecolor{actionheaderbg}{HTML}{8B1A1A}
\definecolor{actionbodybg}{HTML}{FDF2F2}

\newtcblisting{actionbox}[1][Action]{%
    enhanced, breakable, sharp corners,
    listing only,
    boxrule=0pt,
    colback=actionbodybg,
    colbacktitle=actionheaderbg,
    coltitle=white,
    fonttitle=\bfseries\sffamily,
    title={#1},
    listing options={
        basicstyle=\ttfamily\footnotesize,
        breaklines=true,
        breakatwhitespace=false,
        columns=fullflexible,
        keepspaces=true,
        showspaces=false,
        showstringspaces=false,
        upquote=true,
    },
    left=10pt, right=10pt, top=6pt, bottom=6pt,
    before skip=6pt, after skip=8pt,
}

\newcommand{\thought}[1]{\par\noindent\textit{Thought:}~#1\par}

\newcommand{\trajstep}[1]{\par\medskip\noindent\textbf{\large Step #1.}\par\smallskip}

\newtcblisting{plainlisting}[1]{%
    enhanced, breakable, sharp corners,
    listing only,
    boxrule=0pt,
    colback=codebodybg,
    colbacktitle=codeheaderbg,
    coltitle=white,
    fonttitle=\bfseries\sffamily,
    title={#1},
    listing options={
        basicstyle=\ttfamily\footnotesize,
        breaklines=true,
        breakatwhitespace=false,
        columns=fullflexible,
        keepspaces=true,
        showspaces=false,
        showstringspaces=false,
        upquote=true,
    },
    left=10pt, right=10pt, top=8pt, bottom=8pt,
    before skip=10pt, after skip=10pt,
}

\newcommand{\ourdata}{32{,}112}
\newcommand{\ourenv}{110}

\newcommand{\stub}[1]{\textcolor{gray}{\textit{[#1]}}}
\providecommand{\textrc}[1]{\textsc{#1}}
\newcommand{\ourwork}{\textrc{CUA-Gym}}
\newcommand{\ourhub}{\textrc{CUA-Gym-Hub}}
\newcommand{\cuagymtitleicon}{\raisebox{-0.12em}{\includegraphics[height=1.10em,trim=120 55 95 45,clip]{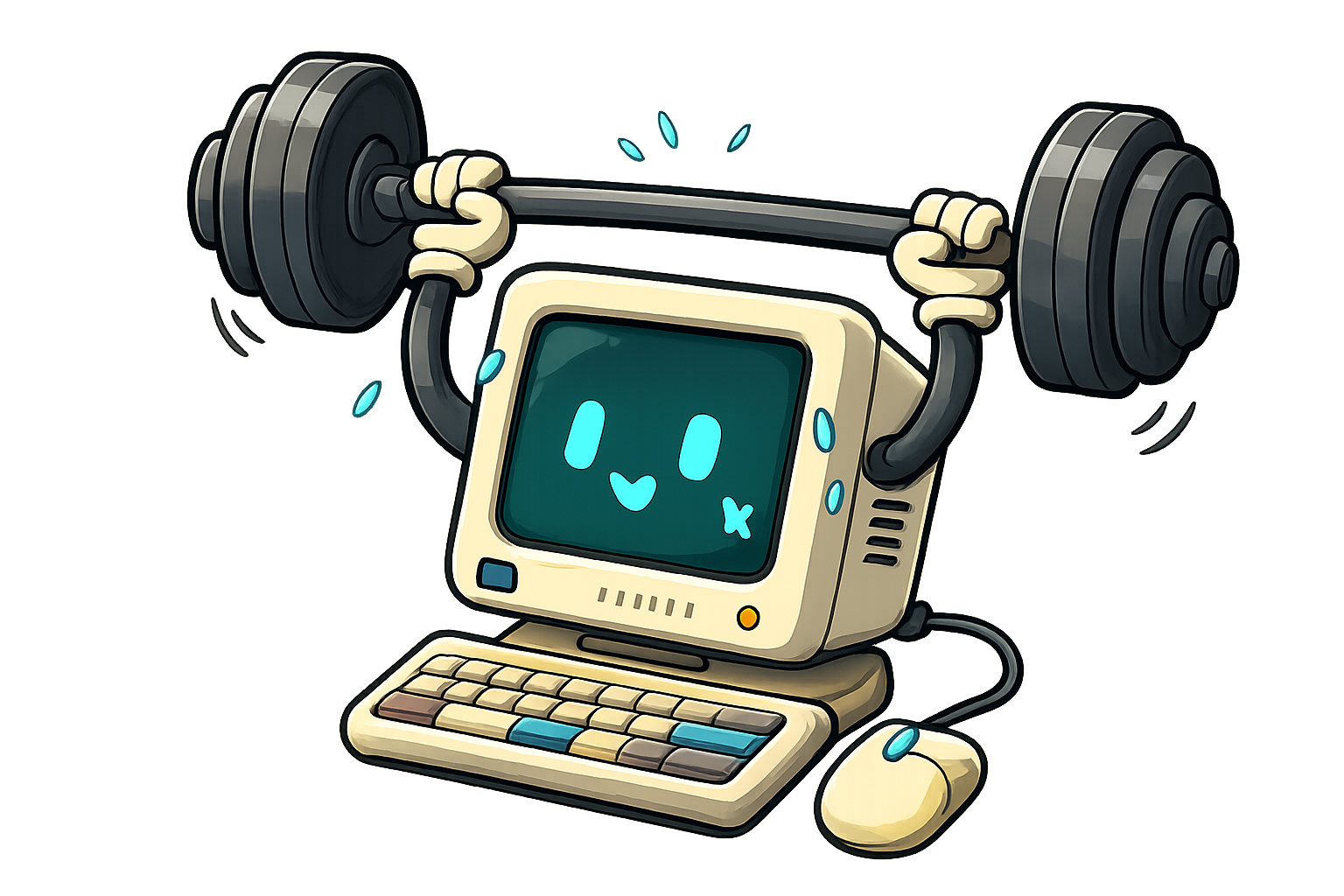}}\hspace{0.25em}}
\newcommand{\smallmodel}{Qwen3.5-35B-A3B}
\newcommand{\largemodel}{Qwen3.5-397B-A17B}
\newcommand{\smallmodelosw}[1]{\textbf{#1}}
\newcommand{\largemodelosw}[1]{\textbf{#1}}
\newcommand{\smallmodelweb}[1]{\textbf{#1}}
\newcommand{\largemodelweb}[1]{\textbf{#1}}
\newcommand{\numdesktop}{16}
\newcommand{\numweb}{94}
\newcommand{\ourweb}{\numweb}

\newcommand{\tasksperleaf}{25}                       % per-leaf task minimum
\newcommand{\templatecap}{3}                         % max instantiations / template / app
\newcommand{\tuplesperapp}{1{,}000}                  % per-app verified-tuple minimum
\newcommand{\rolloutspertask}{4}                     % teacher rollouts sampled per task
\newcommand{\teacherfilterrate}{50}                  % \% of teacher rollouts discarded
\newcommand{\sftcorpussize}{3{,}578}                 % retained SFT trajectories
\newcommand{\sftlr}{7\!\times\!10^{-6}}              % AdamW LR (cosine decay to 7e-7)
\newcommand{\sftbatch}{512}                          % prompts per batch
\newcommand{\sftepochs}{1}                           % epochs over SFT corpus
\newcommand{\vmpoolsize}{2{,}000}                    % parallel OSWorld VMs
\newcommand{\vmutil}{75}                             % avg VM utilization \%

\makeatletter
\expandafter\def\csname trajpath@01\endcsname{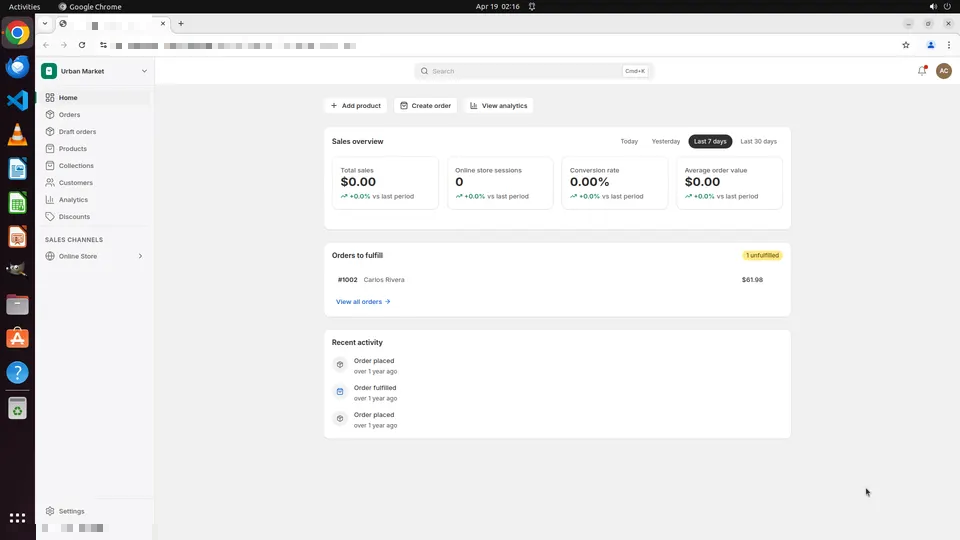}
\expandafter\def\csname trajpath@02\endcsname{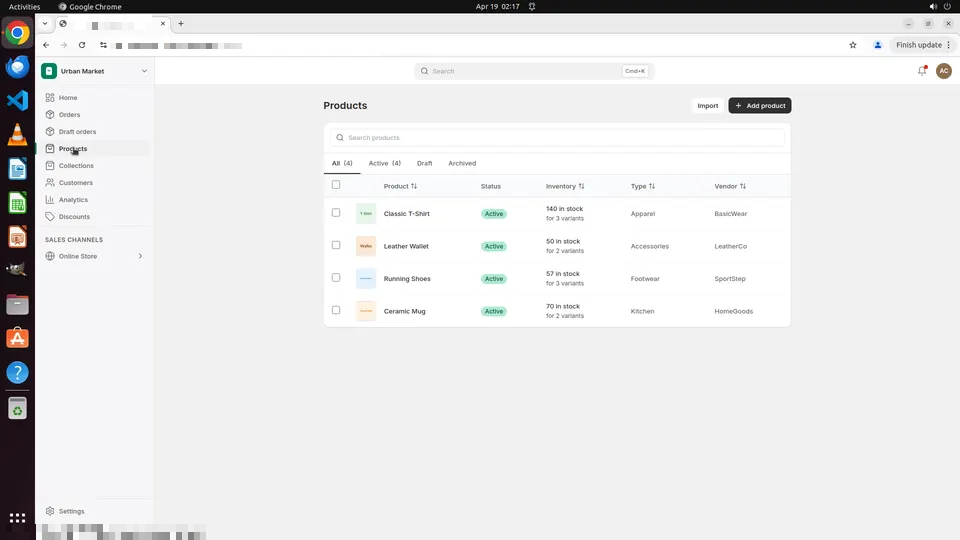}
\expandafter\def\csname trajpath@03\endcsname{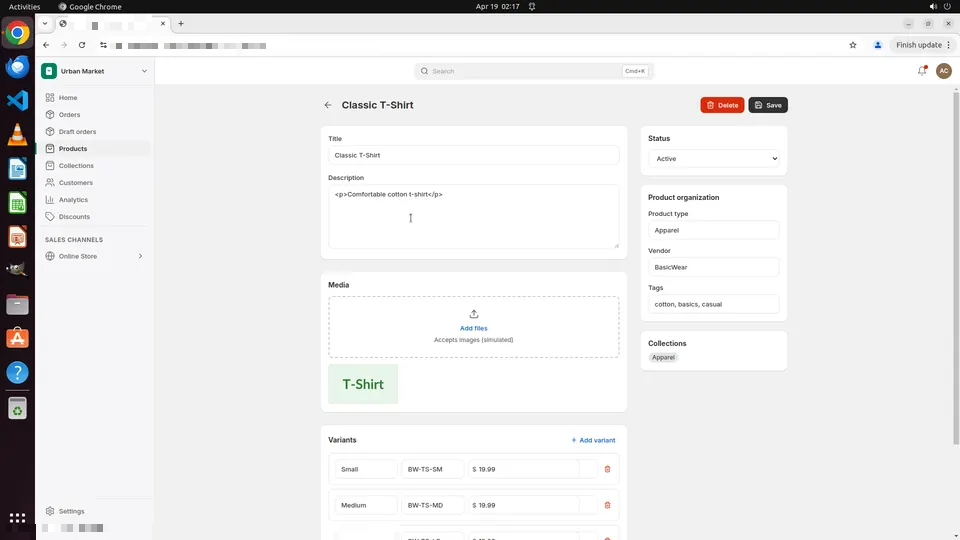}
\expandafter\def\csname trajpath@04\endcsname{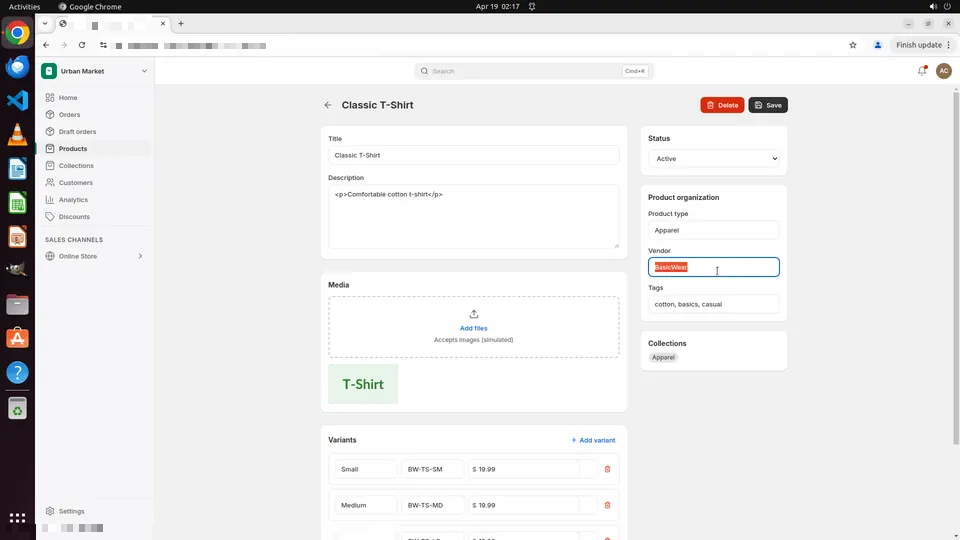}
\expandafter\def\csname trajpath@05\endcsname{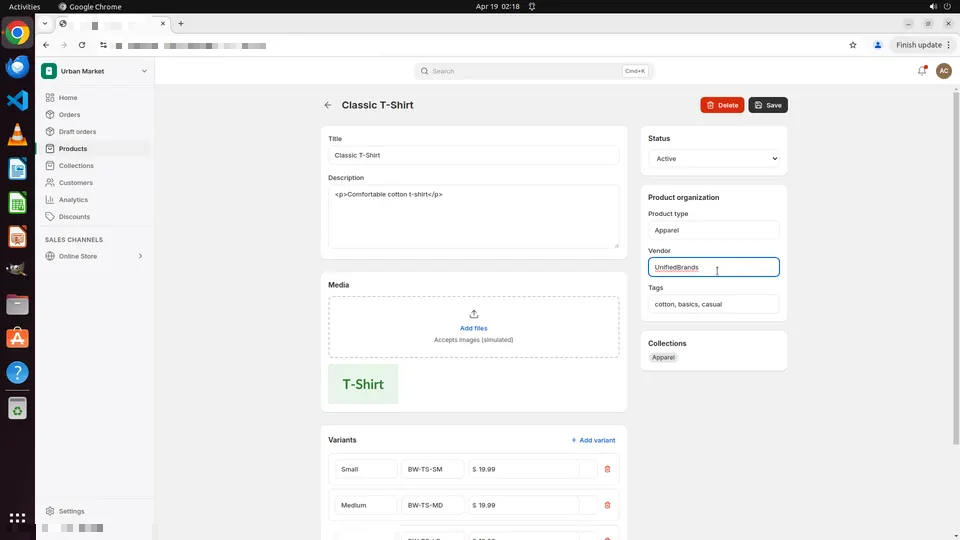}
\expandafter\def\csname trajpath@06\endcsname{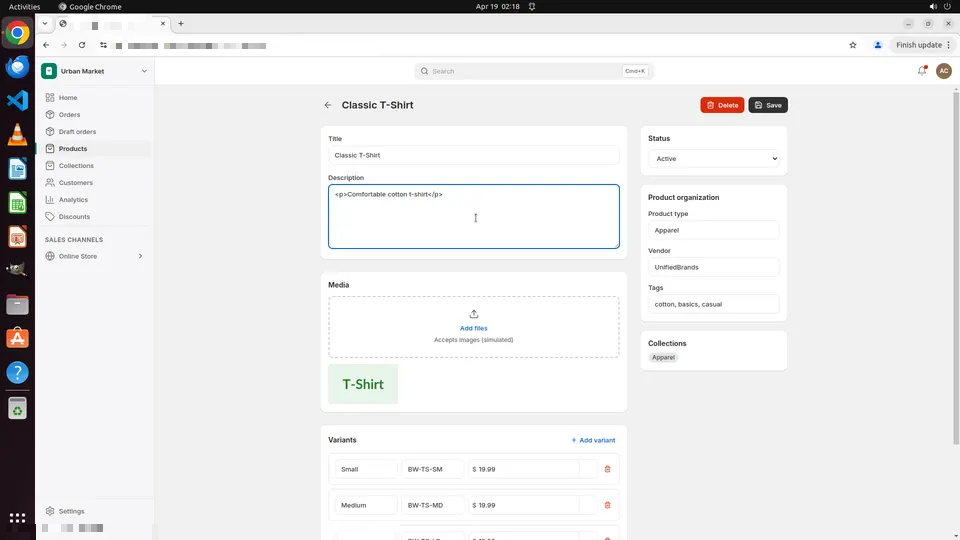}
\expandafter\def\csname trajpath@07\endcsname{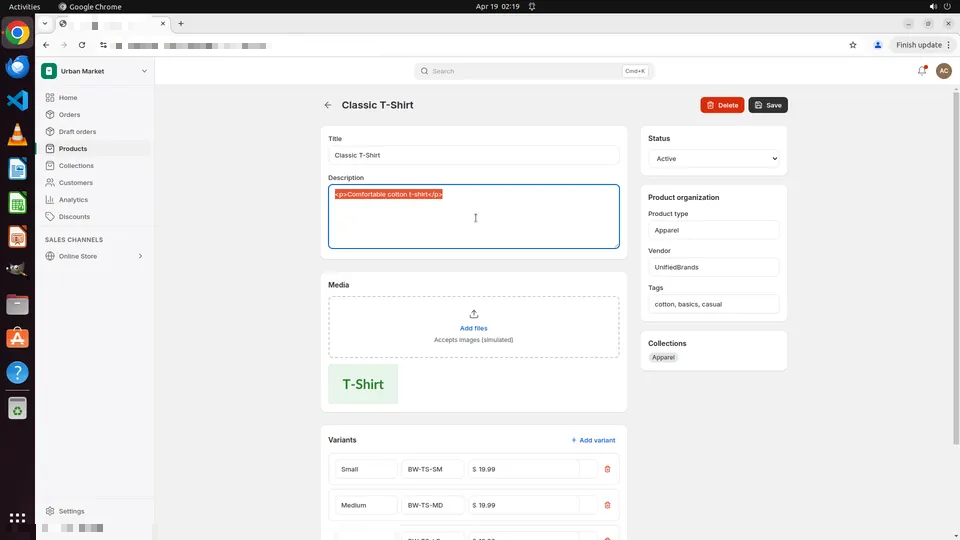}
\expandafter\def\csname trajpath@08\endcsname{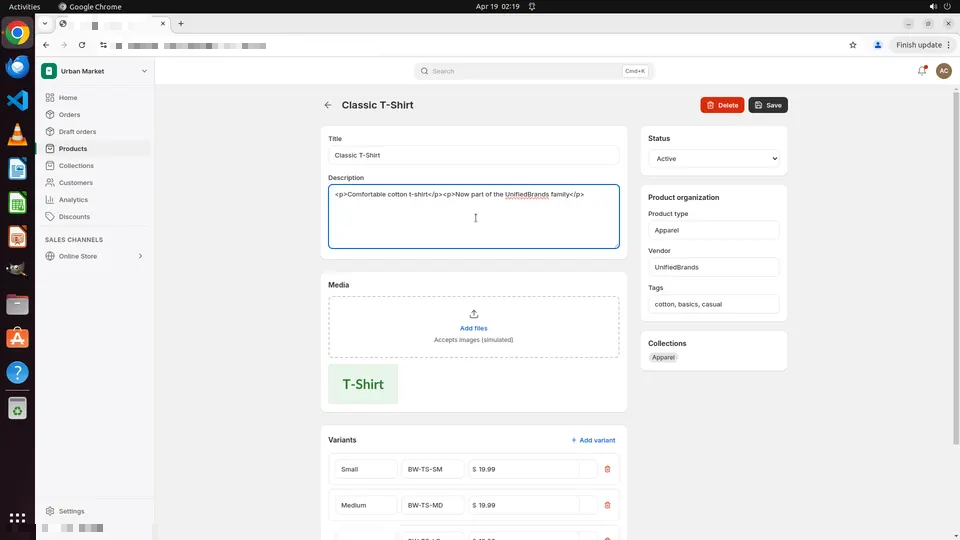}
\expandafter\def\csname trajpath@09\endcsname{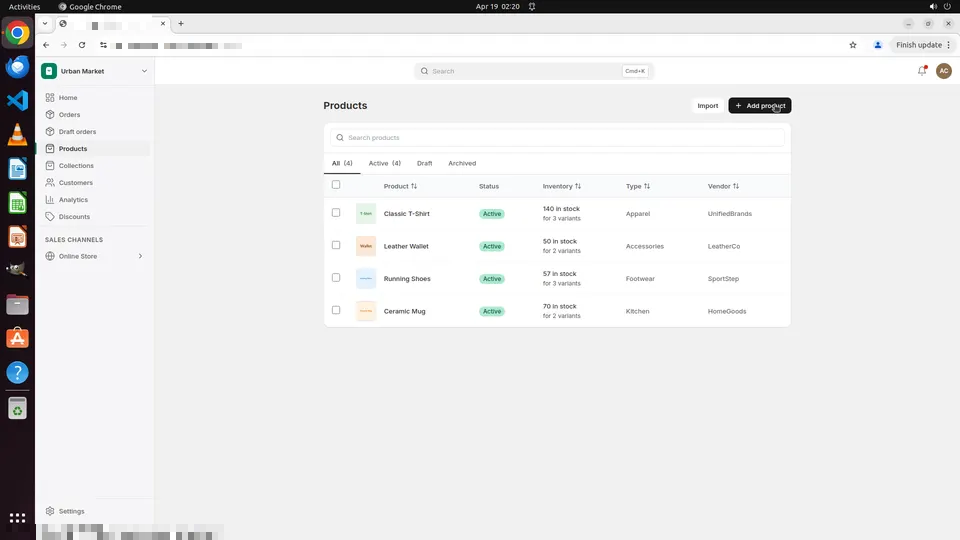}
\expandafter\def\csname trajpath@10\endcsname{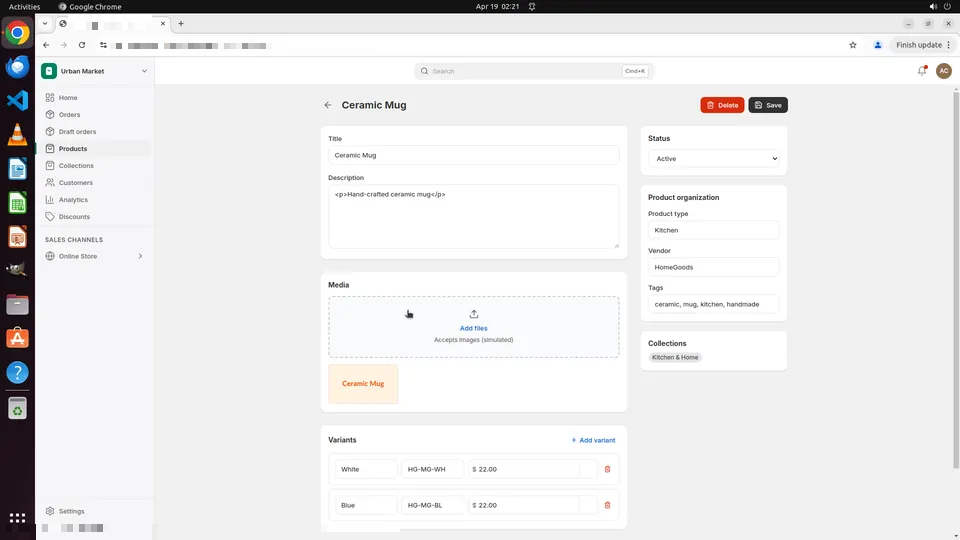}
\expandafter\def\csname trajpath@11\endcsname{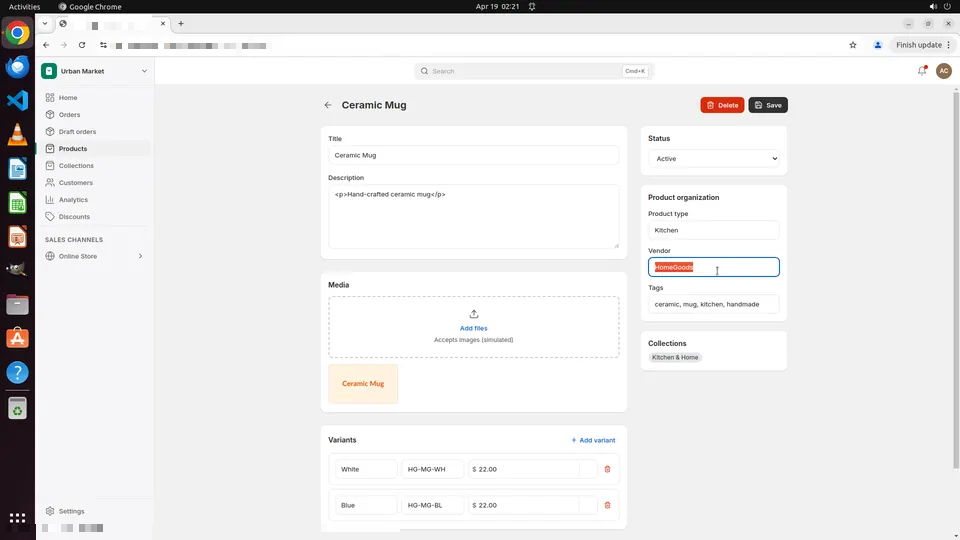}
\expandafter\def\csname trajpath@12\endcsname{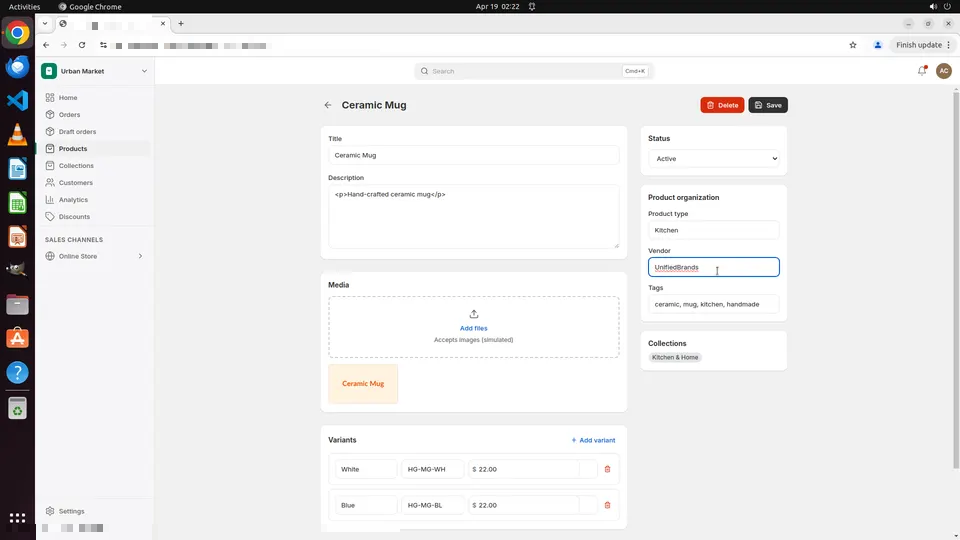}
\expandafter\def\csname trajpath@13\endcsname{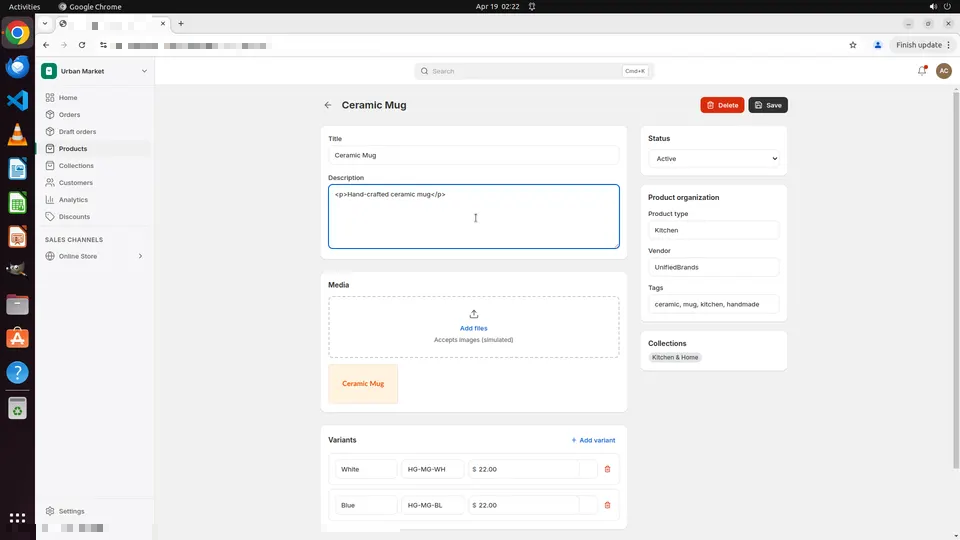}
\expandafter\def\csname trajpath@14\endcsname{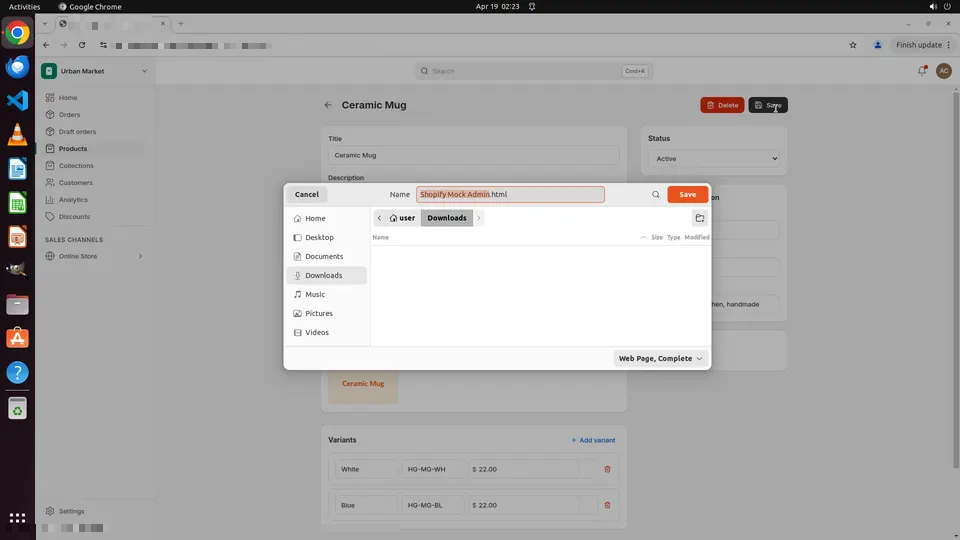}
\expandafter\def\csname trajpath@15\endcsname{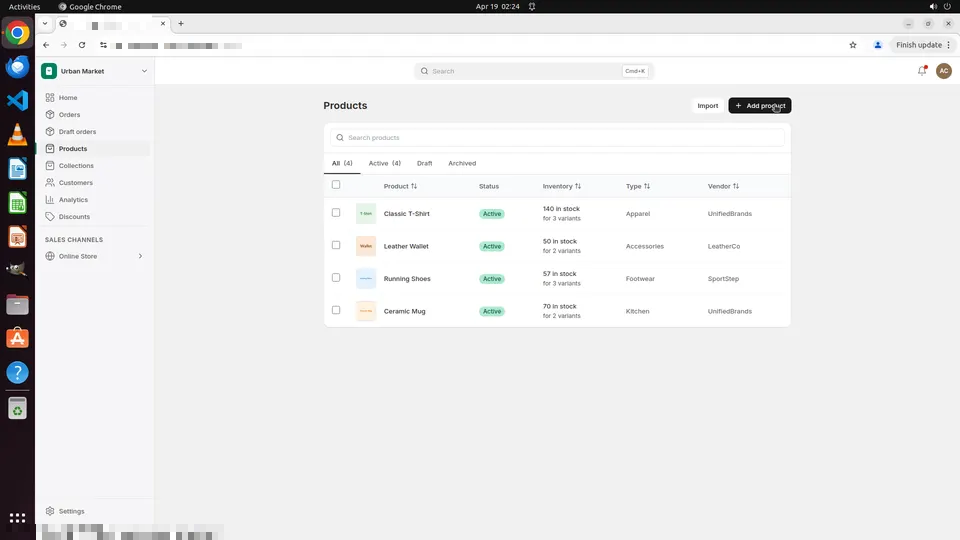}
\expandafter\def\csname trajpath@16\endcsname{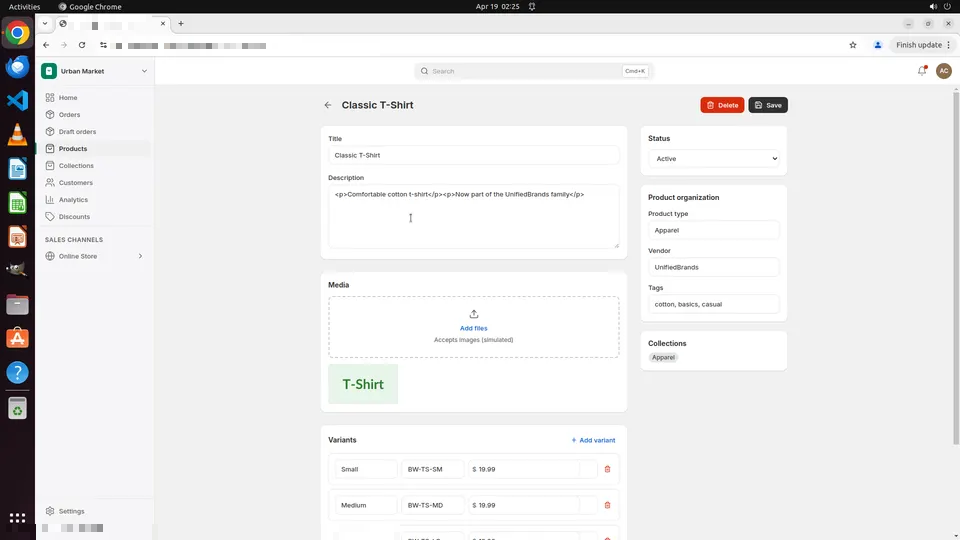}
\expandafter\def\csname trajpath@17\endcsname{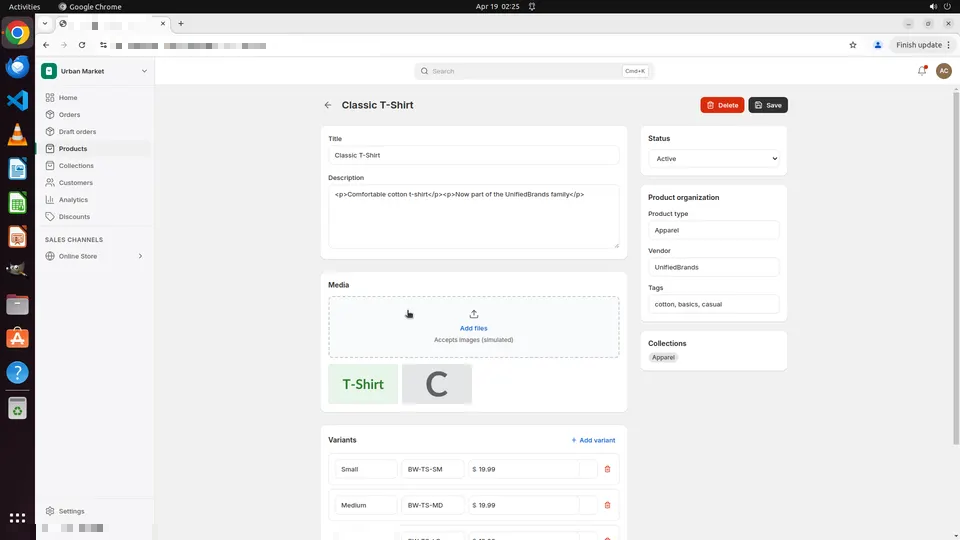}
\expandafter\def\csname trajpath@18\endcsname{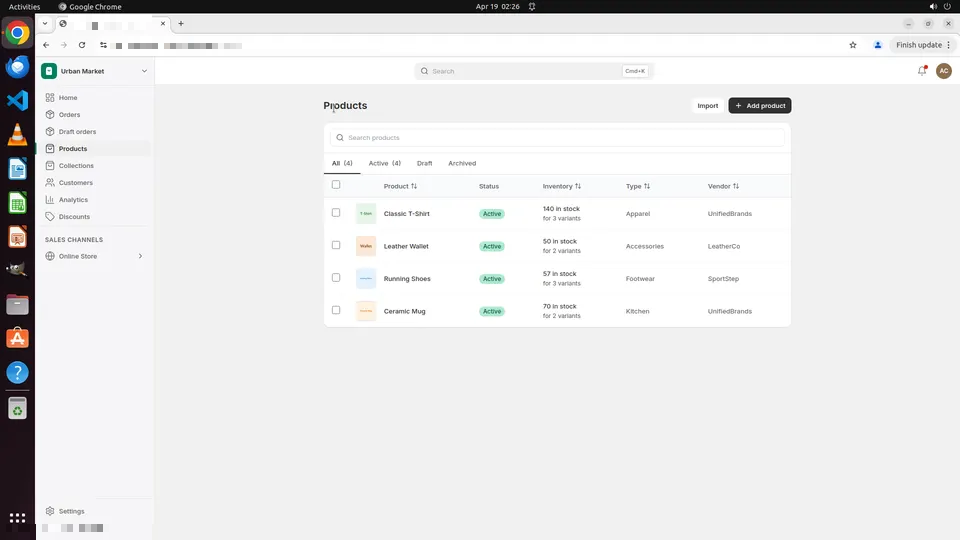}
\expandafter\def\csname trajpath@19\endcsname{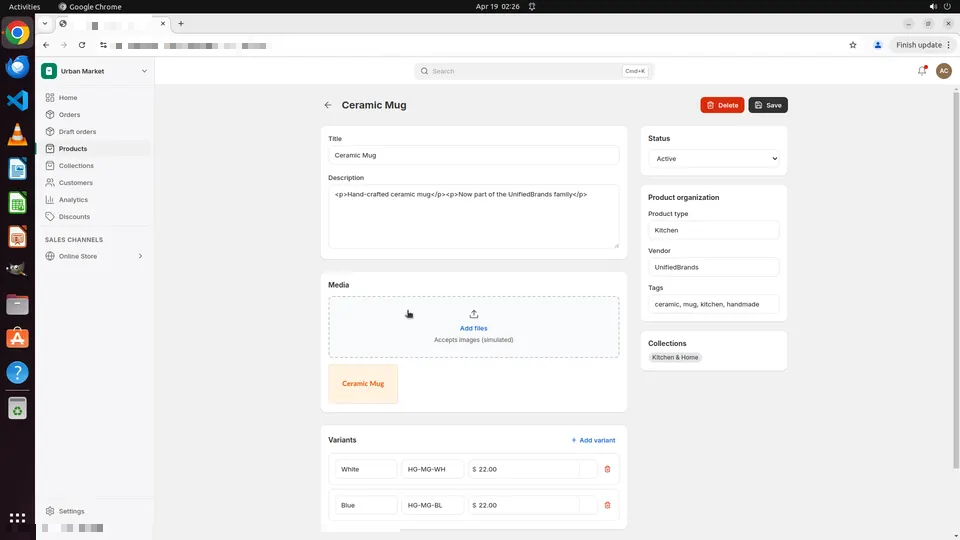}
\newcommand{\trajimg}[1]{%
  \begingroup
  \edef\trajpath{\csname trajpath@#1\endcsname}%
  \IfFileExists{\trajpath}%
    {\includegraphics[width=0.78\linewidth]{\trajpath}}%
    {\fbox{\parbox{0.78\linewidth}{\centering\vspace{2.2em}%
       \textit{\stub{trajectory step #1 screenshot}}\vspace{2.2em}}}}%
  \endgroup
}
\makeatother
\title{\cuagymtitleicon\ourwork{}: Scaling Verifiable Training Environments\\
and Tasks for Computer-Use Agents}

\author{%
\begin{tabular}{@{}c@{}}
\fontsize{8.9}{10.6}\selectfont
\textbf{Bowen Wang}$^{1,2}$ \hspace{0.60em}
\textbf{Dunjie Lu}$^{1,2}$ \hspace{0.60em}
\textbf{Junli Wang}$^{3}$ \hspace{0.60em}
\textbf{Tianyi Bai}$^{2}$ \hspace{0.60em}
\textbf{Shixuan Liu}$^{2}$\\[0.15em]
\fontsize{8.9}{10.6}\selectfont
\textbf{Zhipeng Zhang}$^{2}$ \hspace{0.60em}
\textbf{Haiquan Wang}$^{2}$ \hspace{0.60em}
\textbf{Hao Hu}$^{4}$ \hspace{0.60em}
\textbf{Tianbao Xie}$^{1}$ \hspace{0.60em}
\textbf{Shuai Bai}$^{2}$\\[0.15em]
\fontsize{8.9}{10.6}\selectfont
\textbf{Dayiheng Liu}$^{2}$ \hspace{0.60em}
\textbf{Que Shen}$^{2}$ \hspace{0.60em}
\textbf{Junyang Lin}$^{2}$ \hspace{0.60em}
\textbf{Tao Yu}$^{1}$\thanks{Corresponding author.}\\[0.65em]
\fontsize{8.4}{10.0}\selectfont
$^{1}$The University of Hong Kong
\qquad
$^{2}$Qwen Team, Alibaba Inc.\\[0.2em]
\fontsize{8.4}{10.0}\selectfont
$^{3}$University of California, San Diego
\qquad
$^{4}$Tsinghua University
\end{tabular}
}

\renewcommand{\paperlink}{https://cua-gym.xlang.ai}
\renewcommand{\hflink}{https://huggingface.co/datasets/xlangai/CUA-Gym}
\renewcommand{\ghlink}{https://github.com/xlang-ai/CUA-Gym}
\renewcommand{\envlink}{https://github.com/xlang-ai/CUA-Gym-Hub}
\showcuagymresources

\begin{document}

\maketitle

\etocdepthtag.toc{mtmain}

\begin{abstract}
Reinforcement learning with verifiable rewards (RLVR) has driven breakthroughs in domains such as math, tool-use, and software engineering, yet its extension to computer-use agents (CUAs) has been bottlenecked by the scarcity of scalable training data with deterministic rewards. 
Constructing such data for CUAs requires consistent task instruction, executable environment, and verifiable reward. 
However, hand-curated benchmarks achieve high reward fidelity but cover few applications and LLM-as-judge-based datasets scale broadly but lack reliable verification. 
We present \ourwork{}, a scalable pipeline that co-generates task instructions, 
environment states, and reward functions jointly from a shared topic 
specification.
Concretely, a Generator agent constructs the initial and golden 
environment states, and a separate Discriminator agent writes the 
reward function from the task specification alone. An orchestrator agent 
drives the two through iterative rounds until the reward function 
distinguishes the initial and golden environments upon execution. 
Generated tuples then pass a final filter combining LLM majority 
voting and agent rollouts, ensuring quality beyond the per-task 
adversarial loop.
To address the scarcity of training environments, we further apply 
the same agentic pipeline to synthesize \ourhub{}, a broad suite of 
high-fidelity mock web applications grounded in real-world software-use 
distributions, expanding the reachable scale of CUA RLVR data by magnitude.
Using this pipeline, we construct \ourwork{}, a dataset of \textbf{\ourdata{}}
verified RLVR training tuples grounded in \textbf{\ourenv{}} environments.
Trained with GSPO on \ourwork{}, our \ourwork{}-A3B and \ourwork{}-A17B achieve \smallmodelosw{62.1\%} and \largemodelosw{72.6\%} on OSWorld-Verified, outperforming prior open-source CUAs at comparable scales, with performance scaling smoothly in both data
volume and environment diversity. The same checkpoints also improve on the held-out WebArena browser benchmark, indicating transfer beyond the training environments.
We will open-source the full synthesis pipeline, dataset, \ourhub{} environments, and models.
\end{abstract}

\section{Introduction}

Reinforcement learning with verifiable rewards (RLVR) has emerged as
the dominant post-training paradigm for frontier models,
driving breakthroughs across mathematics
~\citep{deepseek-r1, deepseek-math}, software engineering
~\citep{swe-gym, swe-smith, deepswe, swe-universe}, and terminal-use~\citep{endless-terminal,pi2026dataengineeringscalingllm}. The recipe is now well-understood:
procedurally synthesize large-scale training tasks paired with
deterministic reward signals, then optimize the policy against these
signals with algorithms such as GRPO~\citep{deepseek-math}. Across these domains, the supply of verifiable
training data has repeatedly proven to be the rate-limiting factor
for agent capability, and the data-performance scaling curves remain
unsaturated at current volumes. Extending this recipe to computer-use
agents (CUAs) is the natural and urgent next step, given their
potential to automate the vast landscape of digital knowledge work.
Yet despite rapid progress in CUA foundation
models~\citep{claude-computer-use, ui-tars-2, opencua}, RLVR training
data for this domain remains scarce, fragmented, and far below the
scale that has powered breakthroughs elsewhere.

The bottleneck is structural rather than algorithmic. Unlike math 
or code, where a training instance reduces to a problem statement 
and a checkable answer, a CUA RLVR instance is a tuple $(t, s, r)$ 
of task instruction, executable environment state, and reward 
function, with each component a non-trivial engineering artifact 
that must work together with the others. Hand-authoring a single 
such tuple takes hours of expert effort, and the cost compounds 
with application diversity since each new application brings its 
own setup procedures and verification interfaces. As a consequence, CUA RLVR datasets have remained orders of magnitude smaller than RLVR datasets in adjacent text-only domains such as math, software engineering, and terminal operation.
However, no existing approach simultaneously satisfies the three 
properties required for scalable CUA RLVR: deterministic verifiable 
rewards, broad application coverage, and scalable task diversity. 
Supervised datasets~\citep{opencua, aguvis} cover diverse 
applications but provide trajectory-level imitation targets rather 
than outcome rewards, leaving them unusable for RL. VLM-as-a-judge 
frameworks~\citep{zerogui} can score arbitrary tasks but introduce 
reward noise that destabilizes policy optimization. Code-native 
pipelines~\citep{guigenesis, infiniteweb} produce deterministic 
rewards in browser mocking environments but have so far stayed limited in scale and browser environment. As a result, CUA RLVR has been unable 
to follow the data-performance scaling seen in math and code.

We present \ourwork{}, an agentic pipeline that synthesizes verifiable 
RLVR training data for CUAs at scale. Given a topic specification, 
\ourwork{} jointly produces the task instruction, environment setup, 
and reward function as a single verified tuple, using coding agents 
to handle the engineering work that has previously required human 
experts. The pipeline runs three coordinated agents: a Generator 
that constructs the initial and golden environment states, a 
Discriminator that writes the reward function from the task 
description alone, and an Orchestrator that drives the two through 
iterative rounds until the reward function distinguishes the two 
environments under execution. We further apply strict 
information-isolation to prevent reward hacking, and a final 
filtering pass combining LLM majority voting and agent rollouts 
removes any tuples that pass per-task verification but fail under 
realistic agent behavior.
Beyond per-task synthesis, \ourwork{} treats the environment as a scaling axis. 
Real-world CUA training is gated not just 
by the number of tasks, but by the breadth of 
environments, and existing benchmarks 
cover only a narrow slice of the software landscape that knowledge 
workers actually use. To address this, we first select target applications 
through occupational taxonomies from O*NET~\citep{onet} and software 
usage distributions from the Anthropic Economic Index~\citep{aei}. 
For each selected application, we then apply a multi-agent synthesis pipeline to implement self-contained mock web applications with a unified state API, and 
verify interactive functionality through automated browser-based 
testing. The resulting environments are reproducible, programmable, 
and structurally faithful to their real counterparts, and each 
supports thousands of distinct tasks under the per-task 
co-generation pipeline, expanding the reachable scale of CUA RLVR 
data by orders of magnitude. We package these reusable synthetic web
environments as \ourhub{}, the environment substrate that separates
environment scaling from task synthesis and lets new verified tasks
be generated against a shared, resettable application pool.

We instantiate this pipeline to produce \ourwork{}, a dataset of \ourdata{}
verified RLVR training tuples spanning \ourenv{} environments, including \numdesktop{} desktop
applications and \numweb{} synthesized mock web applications from
\ourhub{}. Using
this dataset, we apply RLVR with GSPO~\citep{gspo} on top of two
open-source MoE backbones: \smallmodel{} and \largemodel{}.
On OSWorld-Verified~\citep{osworld_verified}, our trained models reach \smallmodelosw{62.1\%} and
\largemodelosw{72.6\%} respectively, with the smaller A3B model matching the
performance of the untrained A17B base at roughly $10\times$ fewer
total parameters. The same checkpoints also lift their bases on the
held-out WebArena browser benchmark, demonstrating cross-platform transfer from the synthesized
web mocks to a real browser benchmark. Performance scales smoothly
with both data volume and environment pool size, and expanding the
environment pool from 10 to 80 environments yields gains that
trajectory volume alone cannot recover, identifying environment
diversity as a scaling axis complementary to data volume. We additionally observe that RL training 
spontaneously induces multi-action tool calls, compressing 
trajectories by 33--45\% at matched task performance, an emergent 
efficiency behavior that parallels the spontaneous emergence of 
verification and self-reflection observed in reasoning-focused RL.

Our contributions are threefold. \textbf{(1)} We introduce \ourwork{}, 
an agentic pipeline that co-generates verifiable RLVR training 
tuples with reward-hacking mitigated by construction, together
with the largest open CUA RLVR dataset to date. \textbf{(2)} We 
extend the same agentic paradigm to environment synthesis, release
\ourhub{} as a reusable suite of controllable web environments, and
show that environment diversity is an independent scaling axis 
for CUA RL. \textbf{(3)} We present the first  
demonstration that the RLVR data recipe driving breakthroughs in 
math and code transfers to CUAs at scale, with our two trained 
models, \ourwork{}-A3B and \ourwork{}-A17B, setting state-of-the-art 
OSWorld-Verified results among open-source CUAs at their 
respective scales. We will open-source the full synthesis 
pipeline, dataset, \ourhub{} environments, and models to enable reproducible CUA RLVR 
research.

\section{Methods}
\label{sec:methods}

\ourwork{} scalably automates the construction of RLVR training data for computer-use agents through a co-generation pipeline that produces task instructions, environment states, and reward functions jointly. 
The pipeline scales along two complementary axes: data synthesis (\S\ref{sec:data_synthesis}) and environment scaling (\S\ref{sec:env_synthesis}).

\subsection{Data Synthesis}
\label{sec:data_synthesis}

\begin{figure*}[t]
\centering
\includegraphics[width=\linewidth]{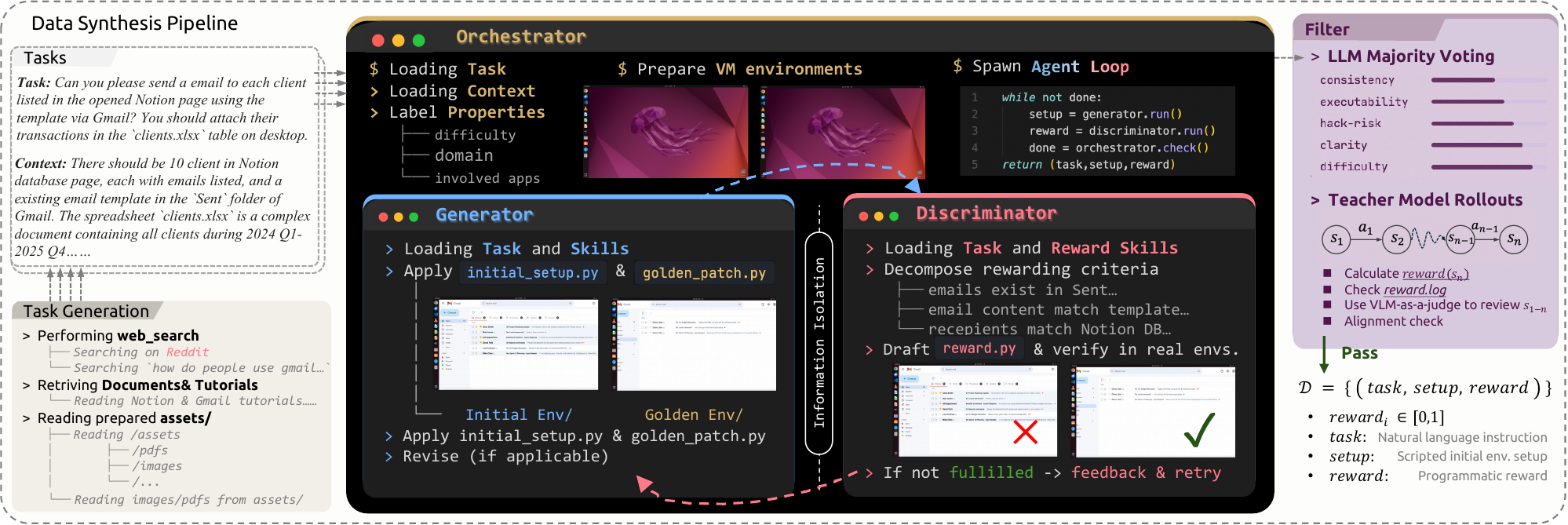}
\caption{\textbf{Overview of the \ourwork{} data synthesis pipeline.}
From a task instruction and its grounded context, the
\textbf{Orchestrator} provisions paired virtual machines and spawns
two adversarially coupled agents. The \textbf{Generator} writes
\texttt{initial\_setup.py} and \texttt{golden\_patch.py} to construct
the initial and golden states. The \textbf{Discriminator}, isolated
from the Generator's scripts by an \textit{information barrier},
independently drafts \texttt{reward.py} from the task alone. The two
agents iterate under the Orchestrator until five agreement conditions
are met. Accepted tuples then pass a \textbf{Filter} combining LLM
majority voting with teacher-model rollout, yielding the final
dataset $\mathcal{D}$.}
% \tao{too many details/words in the figure}  % internal note, hidden for submission
\label{fig:pipeline}
\end{figure*}

We formulate RLVR data for CUA training as a collection of tuples $(t, s, r)$, where $t$ is a natural-language task instruction, $s$ denotes a reproducible initial environment state, and $r: s \rightarrow [0, 1]$ is a verifiable reward function. 
As illustrated in Figure~\ref{fig:pipeline}, \ourwork{} resolves this through a co-generation pipeline in which all three components are derived from a single query specification, ensuring consistency by construction.

\paragraph{Task Generation}
\label{sec:task_gen}
An instruction alone does not determine the environment in which it
must be executed. We therefore generate each task as an
instruction-context pair $(t, c)$, where $c$ enumerates the entities,
files, and application states required of the initial environment.
The context is drawn from three complementary sources of ground
truth: web research on real-world usage patterns, software
documentation, and prepared asset files. Each task is annotated with
difficulty, domain, and applications involved to support the balanced
sampling reported in \S\ref{sec:analysis}. Details are deferred to
Appendix~\ref{app:task_gen}.

\paragraph{Adversarial Setup and Reward Co-Generation}
\label{sec:adversarial}
Writing setup and reward scripts by hand does not scale: each task
demands custom logic across different applications, file formats,
and verification primitives. We delegate this work to coding agents,
which transfer domain expertise across tasks and adapt readily to new
applications. A single agent, however, cannot be trusted with both
sides of the problem. If the same agent constructs the golden state
and writes the reward, the reward tends to re-check the construction
procedure instead of measuring task completion, and the resulting
tuple offers little signal under RL training. We therefore split the
two roles between adversarially coupled subagents separated by a
strict \emph{information barrier}, so that reward synthesis must
draw on the task itself rather than on how the setup was built.

Figure~\ref{fig:pipeline} shows the resulting pipeline. For each
task, the Orchestrator provisions two virtual machines and spawns the
two subagents. The Generator reads the task and the relevant domain
skill, then writes \texttt{initial\_setup.py} and
\texttt{golden\_patch.py}, which it executes on the two VMs to
produce $s_\text{init}$ and $s_\text{gold}$. The Discriminator runs
in a sandboxed process with no access to the Generator's scripts or
working directory; it sees only $t$ and the two resulting
environments. From these it decomposes the task into fine-grained
sub-criteria and writes \texttt{reward.py}, which aggregates them
into a progressive $[0, 1]$ score. The Orchestrator inspects the outputs of both agents after each round and triggers another iteration until the agreement
conditions are jointly satisfied.

Loop-level convergence guarantees that the tuple is internally
consistent, but it cannot surface higher-order issues such as
ambiguous instructions or specifications that no policy can satisfy:
these only manifest under rollout. We therefore pass every accepted
tuple through a dataset-level filter with two independent stages.
The first is an LLM majority vote that scores each tuple along
consistency, executability, hack-risk, clarity, and difficulty
calibration. The second is a teacher-model rollout that verifies the
task is solvable and that the reward signal tracks task success.
Tuples that survive both stages enter the final dataset. This filter
is what closes the gap between loop-level consistency and end-to-end
training utility, removing the ambiguity and infeasibility cases that
the inner loop cannot detect.

\subsection{Environment Scaling}
\label{sec:env_synthesis}

The diversity of \ourwork{} tasks is upper-bounded by the diversity of
available environments. Existing benchmarks cover only a handful of
desktop applications, while real websites are unsuitable as RL
training environments because authentication, rate limits, and
non-reproducible state make them impossible to inject, inspect, or
reset programmatically.
We therefore synthesize \ourhub{}, a suite of self-contained mock web applications that
preserve the interactive fidelity of their real-world counterparts
while exposing full programmatic control over their state. The suite
covers \ourweb{} applications drawn from two sources: widely used
products spanning communication, productivity, e-commerce, and
analytics, and long-tail targets selected to broaden domain coverage.
Each mock is a single-page application backed by a unified HTTP API
that exposes the session state for inspection, injection, and reset.
This design lets the data synthesis pipeline treat each session as
a sandboxed unit and recover state transitions as structured diffs,
giving reward computation an unambiguous substrate. In this sense,
\ourhub{} is not a benchmark by itself, but a reusable environment
layer: the same resettable applications can support many generated
tasks, reward functions, and post-training recipes.

\begin{figure*}[t]
\centering
\includegraphics[width=\linewidth]{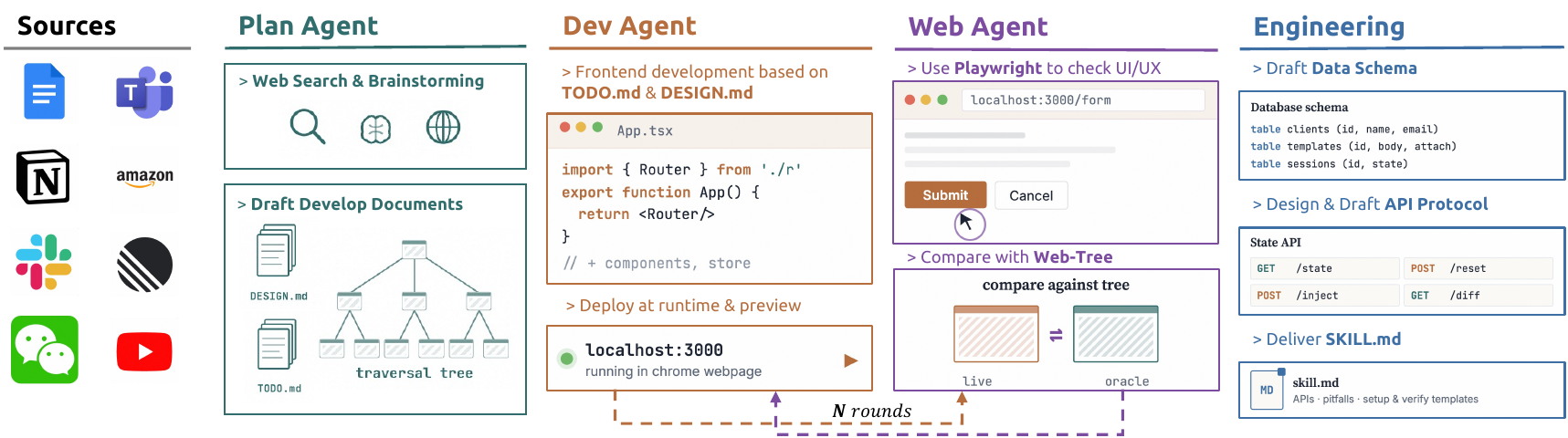}
\caption{\textbf{Multi-agent pipeline for mock environment synthesis.}
From a target application seed, the \textbf{Plan Agent} drafts
\texttt{DESIGN.md} and \texttt{TODO.md} specifying features, data
schema, API protocol, and UI layout. The \textbf{Dev Agent} implements
the single-page application from these specifications. The
\textbf{Web Agent} then exercises every interactive element via
Playwright, comparing the live DOM against the spec and feeding
discrepancies back to the Dev Agent until convergence. The final
artifacts include the state API and a \texttt{SKILL.md} documenting
APIs, pitfalls, and verification templates for downstream task
generation.}
\label{fig:env_pipeline}
\end{figure*}

Synthesizing high-fidelity mocks at scale requires both faithful
modeling of the target application and systematic verification of
interactive correctness, motivating a multi-agent synthesis pipeline
(Figure~\ref{fig:env_pipeline}). A Plan Agent grounds each mock in
its real-world target through web research and reference screenshots,
producing a complete development specification including a UI layout
tree that serves as the verification oracle. A Dev Agent implements
the application from this specification. A Web Agent then exercises
every interactive element and compares the live behavior against the
oracle, returning discrepancies for the Dev Agent to address. The Dev
and Web agents iterate until convergence, removing the need for
manual intervention and ensuring a consistent baseline of fidelity
across all synthesized environments. The resulting mock, together
with a generated \texttt{SKILL.md} documenting its API and
verification templates, plugs directly into the data synthesis
pipeline of \S\ref{sec:adversarial} as a generation source.

\begin{figure*}[t]
\centering
\begin{minipage}[t]{0.32\textwidth}
  \centering
  \includegraphics[width=\linewidth]{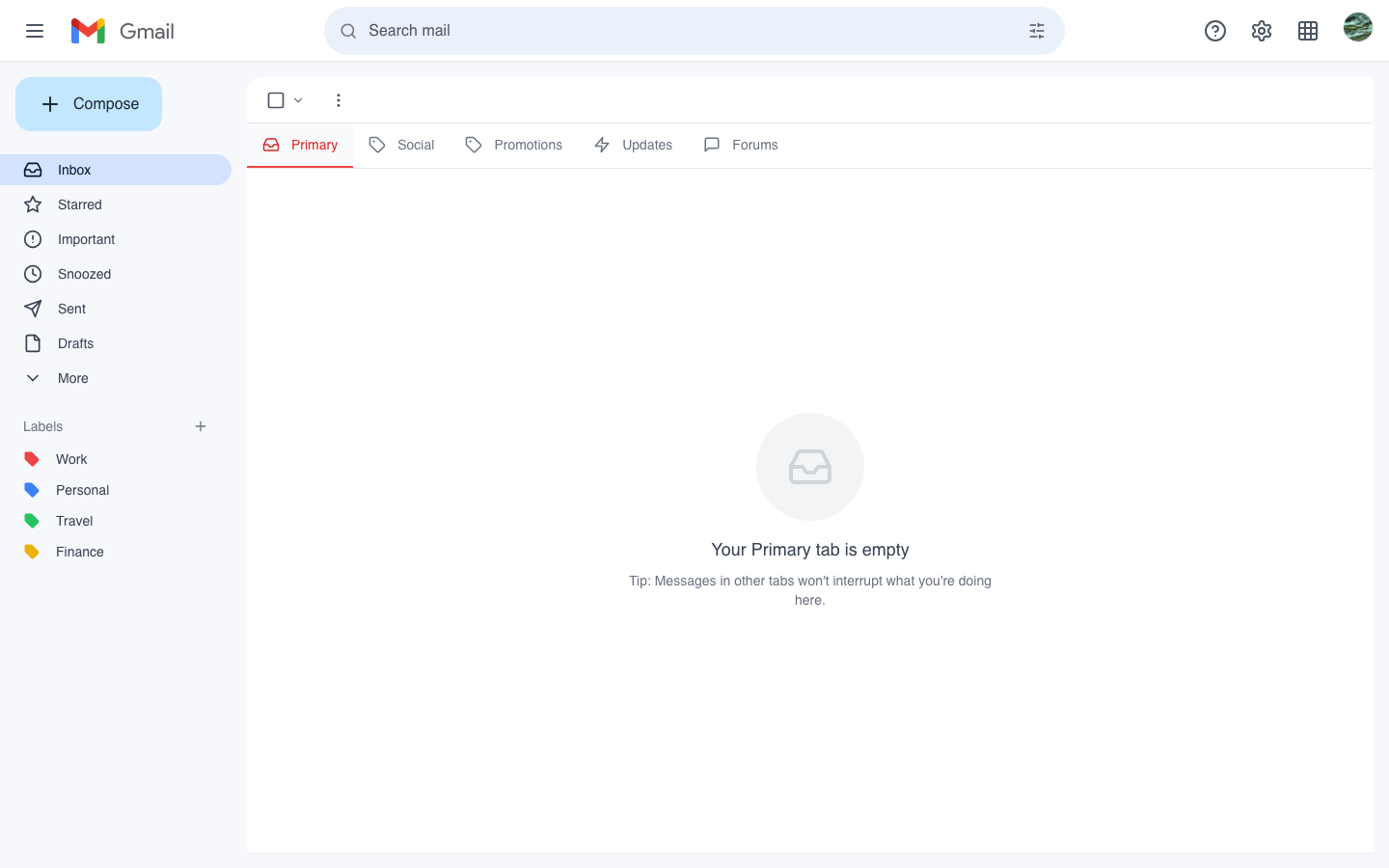}
  \subcaption{Clean inbox.}
  \label{fig:hub_state_clean}
\end{minipage}\hfill
\begin{minipage}[t]{0.32\textwidth}
  \centering
  \includegraphics[width=\linewidth]{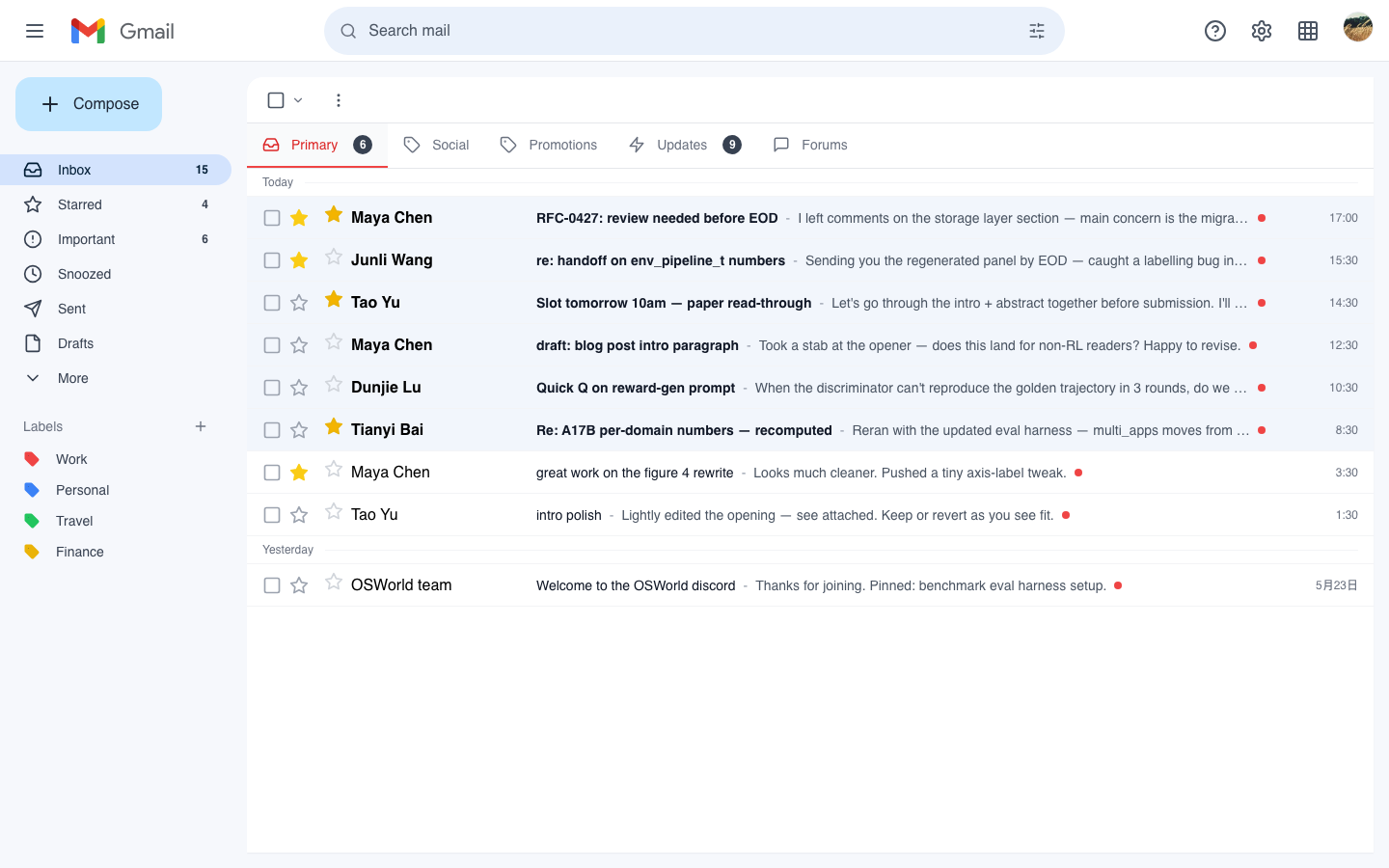}
  \subcaption{Deadline-heavy project state.}
  \label{fig:hub_state_project}
\end{minipage}\hfill
\begin{minipage}[t]{0.32\textwidth}
  \centering
  \includegraphics[width=\linewidth]{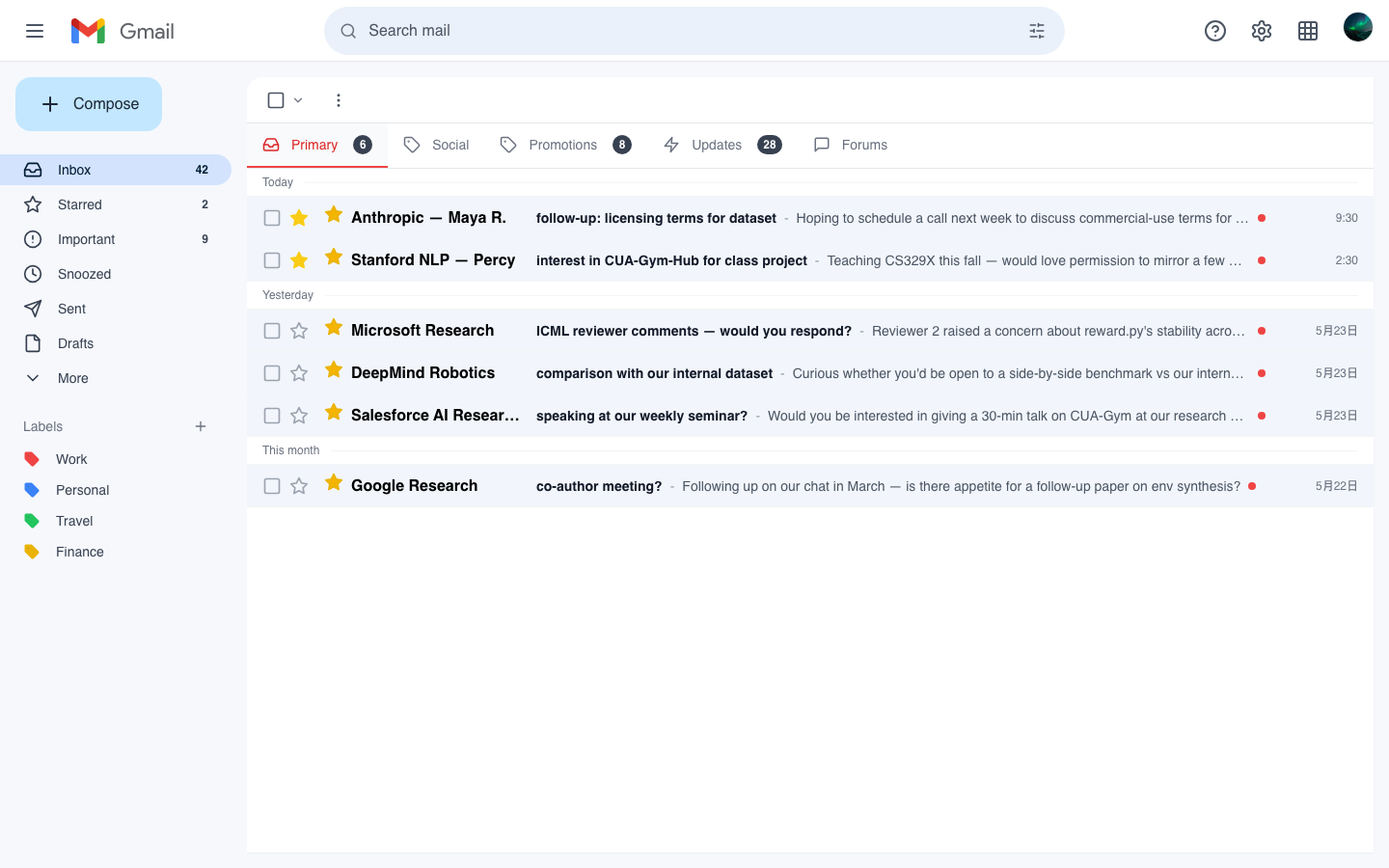}
  \subcaption{Backlog after absence.}
  \label{fig:hub_state_vacation}
\end{minipage}
\caption{\textbf{State-injected environments in \ourhub{}.}
The same mail mock can be instantiated with distinct task-specific
initial states: an empty inbox, a project-deadline state with pending
coordination threads, and a high-volume backlog. Because all mutations
and resets are scoped to a session state, parallel rollout workers can
share the same application implementation while observing isolated
world states.}
\label{fig:hub_state_injection}
\end{figure*}

Two design choices make these mocks dynamic training environments
rather than fixed website replicas. \emph{State injection} materializes
the task-specific world from a structured initial state, so a single
mock implementation can support many distinct tasks without code
changes. \emph{Session isolation} scopes every mutation, upload, and
reset to one session, allowing distributed RL workers to run
concurrently against a shared pool of mock backends without observing
one another's changes. Figure~\ref{fig:hub_state_injection}
illustrates this property in a mail environment: the interface and
application code are unchanged, but the injected state changes the
available messages, labels, unread counts, and therefore the feasible
task distribution.

\section{Experiments}
\label{sec:experiments}
We synthesize \ourwork{} training data over an environment pool
designed to reflect real-world software usage. The pool combines mock
web applications, with targets drawn from occupational task
taxonomies in O*NET~\citep{onet} and software usage distributions in
the Anthropic Economic Index~\citep{aei}, with a
representative set of desktop applications spanning office
productivity, web browsing, image editing, development tools, and
media. Quantitative breakdowns of the resulting dataset are reported
in \S\ref{sec:analysis}.

We evaluate \ourwork{} along two dimensions: the quality of the
generated training data, measured by the performance of agents trained
on it, and the scalability of the pipeline, measured by the effect of
data scale on downstream agent performance.

\subsection{Experimental Setup}
\label{sec:experimental-setup}

\paragraph{Dataset.}
We sample 10{,}858 verified tuples from \ourwork{} as our RLVR training
set, covering tasks across $80+$ environments. We additionally
curate 3{,}578 trajectories for SFT warm-up by rolling out
Claude-Sonnet-4-6~\citep{claude_sonnet} on the same task set and
retaining only successful completions.

\paragraph{Model.}
We conduct experiments at two parameter scales of the Qwen3.5
family~\citep{qwen35}: \smallmodel{} and \largemodel{}.

\begin{center}
\captionsetup{type=figure}
\centering
\includegraphics[width=0.93\textwidth]{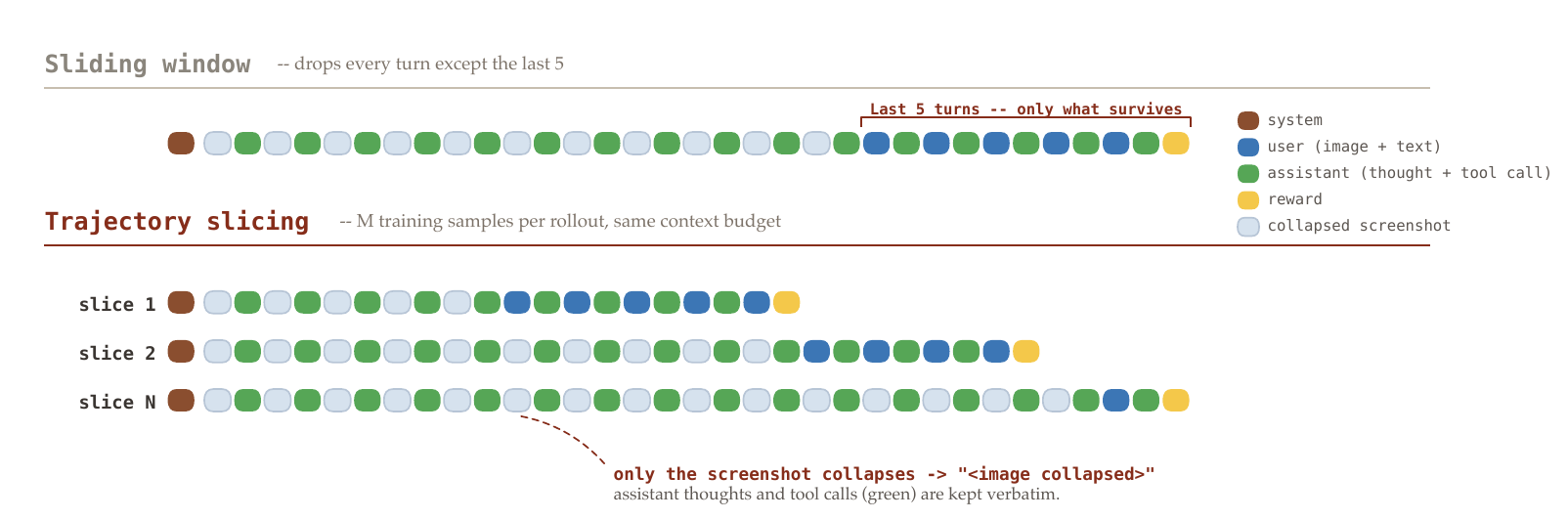}
\caption{\textbf{Training scaffold for long-horizon CUA trajectories.}
Instead of training on a single sliding window that drops older
interaction state, trajectory slicing keeps recent multimodal
observations active while collapsing earlier screenshots into
deterministic placeholders. This preserves late-trajectory supervision
under a fixed context budget and exposes stable prefixes that are
friendly to KV-cache reuse during policy log-probability computation.}
\label{fig:scaffold}
\end{center}

\paragraph{Scaffold.}
The agent observes raw screenshots and emits actions wrapped in
tool-call tags. Long-horizon CUA rollouts create two practical
pressures: context management for image-heavy histories, and efficient
training over many overlapping trajectory prefixes. We therefore use
the scaffold in Figure~\ref{fig:scaffold}: a deterministic
\emph{trajectory slicing} scheme that constructs multiple training
samples per rollout under a fixed context budget. Compared with a
traditional sliding window, which simply shifts over the conversation
and discards the older state that explains later actions, our slices
retain the system/task prefix, collapse stale screenshots into compact
placeholders, and keep recent observations and assistant actions in
full multimodal form. The resulting format preserves supervision for
late turns, avoids lossy learned summaries, and keeps reusable prefixes
explicit so that KV caches can be reused when evaluating policy
log-probabilities over neighboring slices. Full construction details
are given in Appendix~\ref{app:traj_slicing}.

\paragraph{Algorithm.}
We train all models with Group Sequence Policy Optimization
(GSPO)~\citep{gspo}, which provides greater stability for mixture-of-experts RL training. Given task instruction $t$ and initial state $s$,
the policy $\pi_\theta$ generates $G$ rollouts
$\{\tau_1, \ldots, \tau_G\}$, each receiving a verifiable reward
$r_i = r(s, \tau_i) \in [0, 1]$. GSPO maximizes a clipped surrogate objective weighted by the group-normalized
advantage $\hat{A}_i = (r_i - \mu) / \sigma$, where $\mu$ and $\sigma$
are the mean and standard deviation of $\{r_j\}_{j=1}^{G}$, using a
sequence-level importance ratio:
\begin{equation}
\mathcal{L}_\text{GSPO}(\theta)
= \mathbb{E}_{t, s\sim{\mathcal{D}},\, \tau_i \sim \pi_{\theta_\text{old}}}
\Bigg[ \frac{1}{G} \sum_{i=1}^{G}
\min\!\Big( \rho_i\, \hat{A}_i,\, \text{clip}(\rho_i, 1-\varepsilon, 1+\varepsilon)\, \hat{A}_i \Big)
- \beta\, D_\text{KL}[\pi_\theta \| \pi_\text{ref}]
\Bigg],
\end{equation}
where the sequence-level importance ratio is
\begin{equation}
\rho_i = \left( \frac{\pi_\theta(\tau_i \mid t, s)}{\pi_{\theta_\text{old}}(\tau_i \mid t, s)} \right)^{1/|\tau_i|},
\end{equation}
with $|\tau_i|$ the trajectory length, and $\varepsilon$, $\beta$
respectively controlling the clipping threshold and the KL penalty
against the reference policy $\pi_\text{ref}$.

\begin{center}
\captionsetup{type=figure}
\centering
\begin{minipage}[c]{0.53\textwidth}
  \centering
  \includegraphics[width=\linewidth]{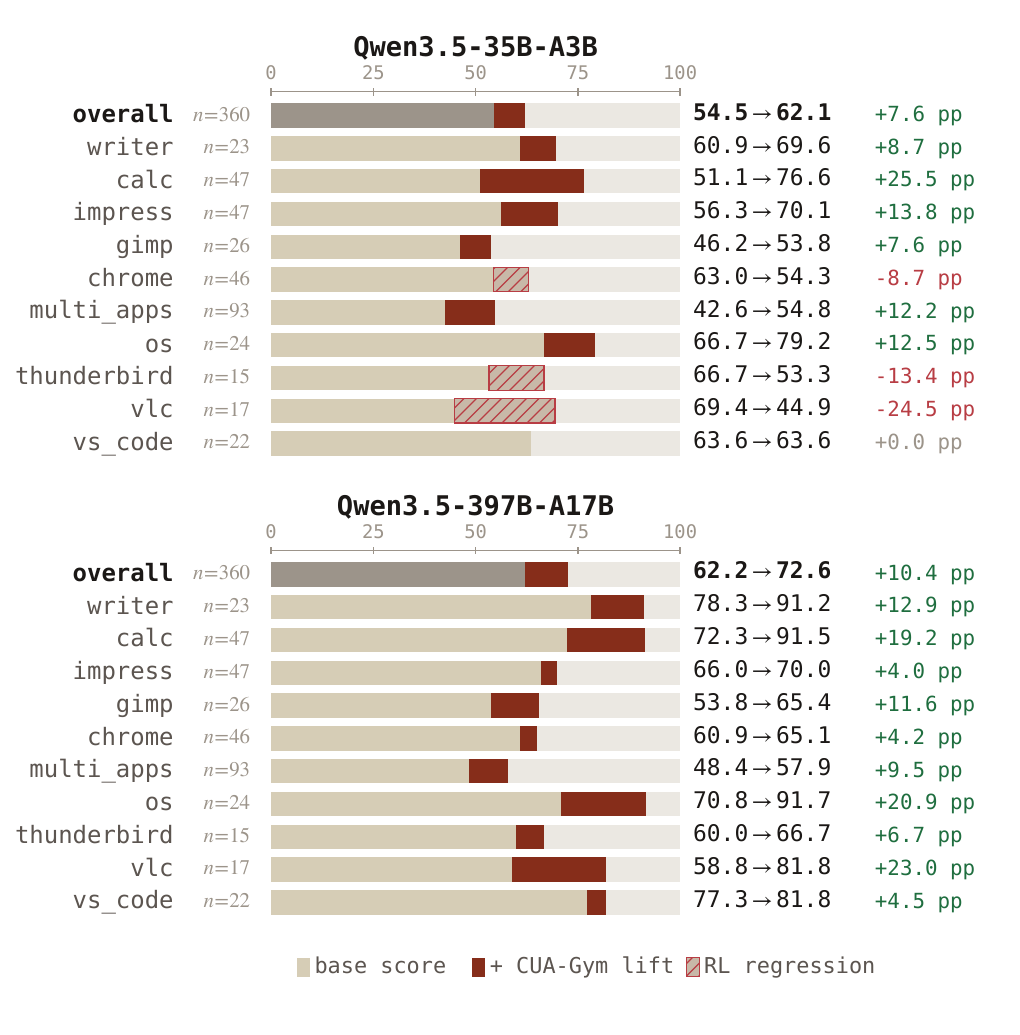}
  \subcaption{Per-domain success rates on OSWorld-Verified.}
  \label{fig:per_domain}
\end{minipage}\hfill
\begin{minipage}[c]{0.45\textwidth}
  \centering
  \footnotesize
  \setlength{\tabcolsep}{6pt}
  \renewcommand{\arraystretch}{0.95}
  \begin{tabular}{@{}lcc@{}}
    \toprule
    \textbf{Model} & \textbf{OSWorld-V.} & \textbf{WebArena} \\
    \midrule
    \multicolumn{3}{@{}l}{\textit{Proprietary Models}} \\
    \cmidrule(r){1-3}
    Claude Sonnet 4.6      & 72.9 & 65.6 \\
    Claude Opus 4.7        & 78.0 & --   \\
    GPT-5.5                & 78.7 & --   \\
    \midrule
    \multicolumn{3}{@{}l}{\textit{Open-source Models}} \\
    \cmidrule(r){1-3}
    EvoCUA-8B              & 46.1 & --   \\
    EvoCUA-32B             & 56.7 & --   \\
    OpenCUA-32B            & 34.8 & --   \\
    OpenCUA-72B            & 45.0 & --   \\
    Step-GUI-8B            & 40.2 & --   \\
    Kimi-K2.6              & 73.1 & --   \\
    \midrule
    \multicolumn{3}{@{}l}{\textit{Our Models}} \\
    \cmidrule(r){1-3}
    \smallmodel{}        & 54.5 & 40.8 \\
    \largemodel{}      & 62.2 & 54.0 \\
    \textbf{\ourwork{}-A3B}   & \smallmodelosw{62.1} & \smallmodelweb{44.5} \\
    \textbf{\ourwork{}-A17B}  & \largemodelosw{72.6} & \largemodelweb{56.0} \\
    \bottomrule
  \end{tabular}
  \subcaption{Main results on OSWorld-Verified and WebArena.}
  \label{tab:main_results}
\end{minipage}
\caption{\textbf{Main results.}
(\subref{fig:per_domain}) Per-domain success rates on OSWorld-Verified
for \ourwork{}-A3B (red) versus its SFT-initialized base (gray,
dashed); $n$ denotes task count per domain, and trained per-domain
scores use the run-4 checkpoint breakdown. Largest gains appear on
\texttt{libreoffice\_calc}, \texttt{multi\_apps}, and \texttt{vs\_code}.
(\subref{tab:main_results}) Success rates on OSWorld-Verified and
WebArena. RLVR on \ourwork{} data improves both \smallmodel{} and
\largemodel{} bases, with \ourwork{}-A3B matching the
\largemodel{} base at roughly $10\times$ fewer total parameters.}
\label{fig:main_results}
\end{center}

\paragraph{Main Results}
\label{sec:main-results}
Table~\ref{tab:main_results} reports performance on
OSWorld-Verified~\citep{osworld_verified}, a verified variant of
OSWorld~\citep{xie2024osworldbenchmarkingmultimodalagents}, and
WebArena. RLVR on \ourwork{} data improves the base model at
both parameter scales: \ourwork{}-A3B lifts the \smallmodel{} base
from 54.5 to 62.1 ($+7.6$\,pp), and \ourwork{}-A17B lifts the
\largemodel{} base from 62.2 to 72.6 ($+10.4$\,pp). The gain
persists at the larger scale rather than diminishing, which is the
regime in which RL gains for GUI agents have most often been observed
to plateau.
% Meanwhile, \ourwork{}-A3B reaches 62.1 on OSWorld-Verified,
% matching the unmodified \largemodel{} base (62.2) at roughly
% $10\times$ fewer total parameters.
The same checkpoints also improve on WebArena, a held-out browser
benchmark whose site clones are disjoint from the \numweb{} mock web
applications used during training. \ourwork{}-A3B lifts the
\smallmodel{} base from 40.8 to 44.5 ($+3.7$\,pp), and
\ourwork{}-A17B lifts the \largemodel{} base from 54.0 to 56.0
($+2.0$\,pp). The
transfer indicates that the web-mock pipeline of
\S\ref{sec:env_synthesis} produces skill that generalizes across
browser environments rather than overfitting to the synthesized
mocks, supporting the joint desktop and web positioning of
\ourwork{}.

\section{Analysis}
\label{sec:analysis}

% \subsection{Data and Environment Analysis}
% \label{sec:data_analysis}

\begin{figure*}[t]
\centering

% --- Top row: two image subfigures ---
\begin{subfigure}[t]{0.48\textwidth}
    \includegraphics[width=\linewidth]{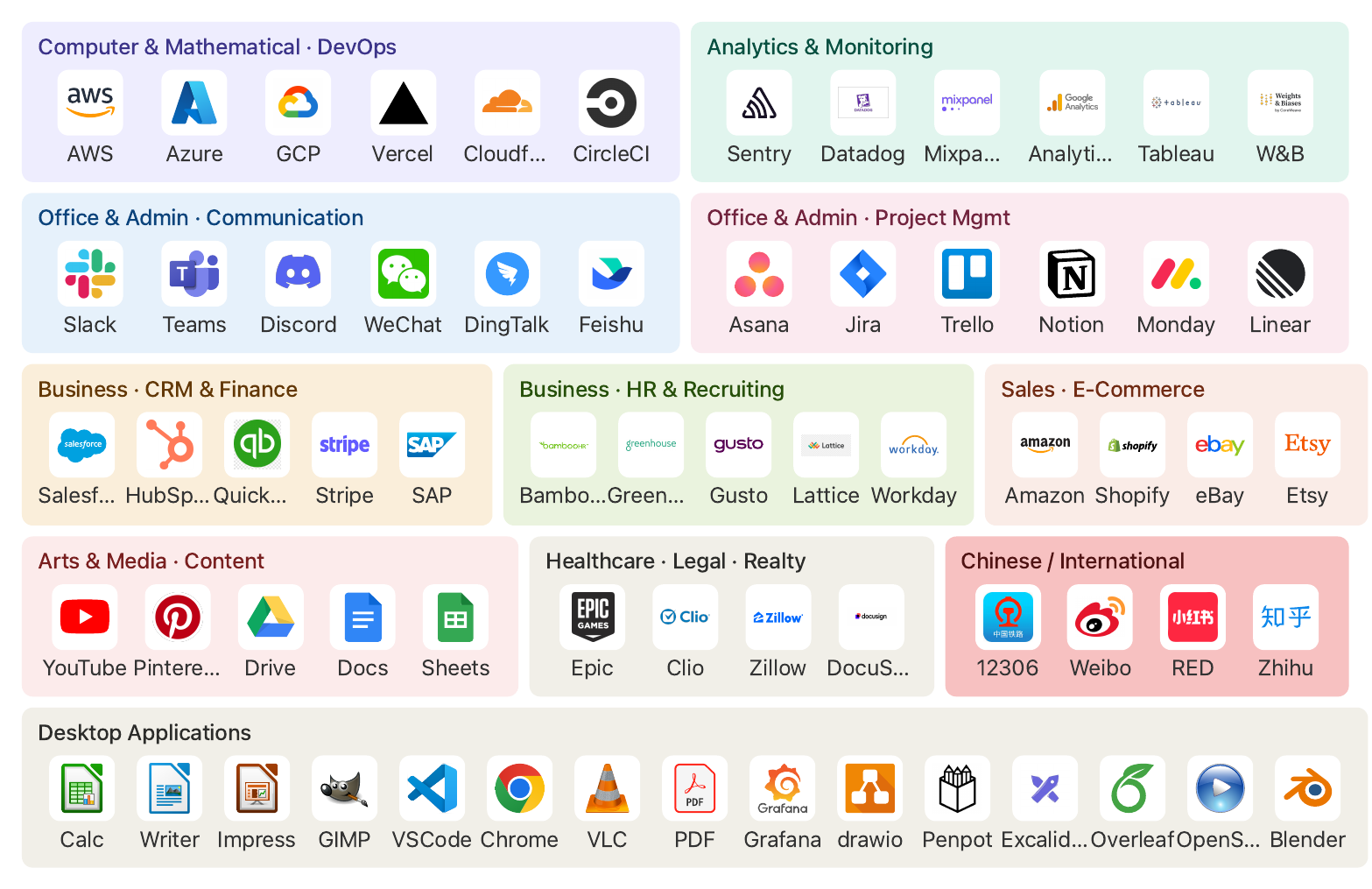}
    \caption{Environment coverage across occupational categories.}
    \label{fig:overview-env}
\end{subfigure}
\hfill
\begin{subfigure}[t]{0.48\textwidth}
    \includegraphics[width=\linewidth]{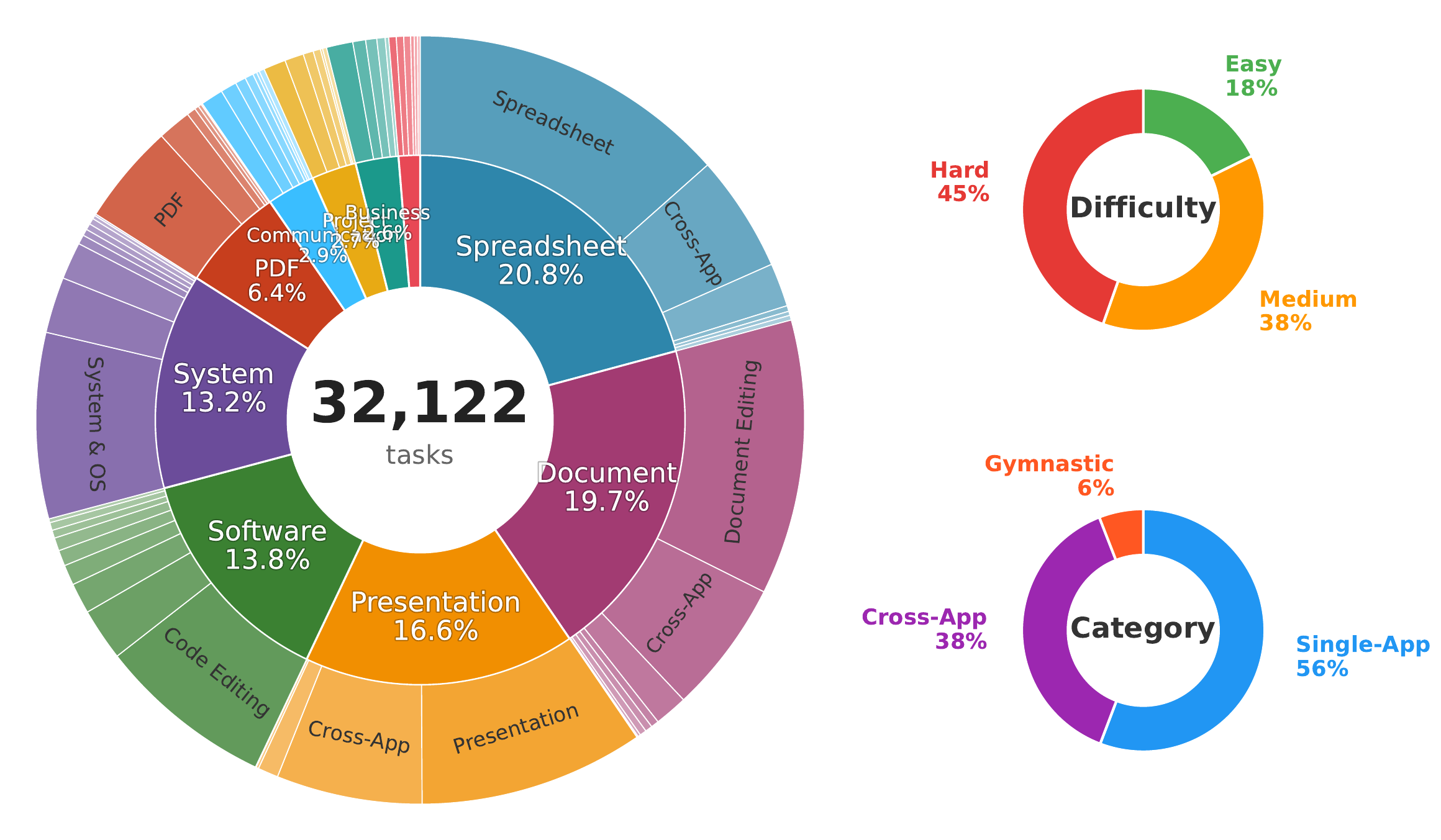}
    \caption{Task distribution across category, difficulty, and cross-app complexity.}
    \label{fig:overview-data}
\end{subfigure}

\vspace{6pt}

% --- Bottom: dataset comparison table as a subfigure ---
\begin{subfigure}[t]{\textwidth}
\centering
\footnotesize
\setlength{\tabcolsep}{8pt}
\renewcommand{\arraystretch}{1.0}
\begin{tabular}{@{}llrrll@{}}
\toprule
\textbf{Dataset} & \textbf{Platform} & \textbf{Data Size} & \textbf{Env. Size} & \textbf{Reward} & \textbf{Open} \\
\midrule
GUI-Genesis~\citep{guigenesis}            & Mobile               & 969              & 1             & Programmatic           & No            \\
WebArena-Infinity~\citep{webarenainfty}   & Web                  & 1{,}260          & 10            & Programmatic           & Yes           \\
InfiniteWeb~\citep{infiniteweb}           & Web                  & 600              & --            & Programmatic           & No$^{\star}$  \\
UltraCUA~\citep{ultracua}                 & Desktop              & 17{,}000         & 9             & Programmatic           & No$^{\star}$  \\
Gym-Anything~\citep{gym-anything}          & Desktop              & 7{,}277          & 193           & VLM                    & Yes           \\
\midrule
\textbf{\ourwork{} (Ours)} & \textbf{Desktop+Web} & \textbf{\ourdata{}} & \textbf{\ourenv{}} & \textbf{Programmatic} & \textbf{Yes} \\
\bottomrule
\end{tabular}
\caption{Comparison with existing GUI agent training datasets.
\textit{Programmatic} = code-native assertions; \textit{VLM} = VLM-as-a-judge;
$^{\star}$ promised but not yet released.}
\label{tab:dataset_comparison}
\end{subfigure}

\caption{\textbf{\ourwork{} data and environments.}
\textbf{(a)} \ourwork{} spans \ourenv{} environments, including \ourhub{} web environments aligned with O*NET occupational categories~\citep{onet}.
\textbf{(b)} The \ourdata{} verified tuples cover application categories evenly (no category exceeds 21\%) and are skewed toward harder (45\%) and cross-application (38\%) tasks.
\textbf{(c)} Compared to existing RLVR datasets, \ourwork{} is the largest open-released collection and uniquely covers both desktop and web platforms with programmatic verification.}
\label{fig:overview}
\end{figure*}

The \ourwork{} dataset contains \ourdata{} verified RLVR training
tuples spanning \ourenv{} environments. The environment pool
(Figure~\ref{fig:overview-env}) is grounded in two external sources:
occupational task taxonomies from O*NET, which determine the
\ourweb{} web application categories aligned with SOC major
groups, and software usage distributions from the Anthropic Economic
Index, which shape the within-category selection. A complementary
desktop suite covers OS-level workflows that web environments cannot
express. We release the synthesized web portion as \ourhub{}, a
resettable environment suite intended to make environment coverage
reusable beyond the particular \ourwork{} task corpus. Task
composition (Figure~\ref{fig:overview-data}) is
balanced across categories, with no single category exceeding 21\%,
and is weighted toward the harder regime: 45\% of tasks are labeled
hard, and 38\% involve cross-application workflows.
Table~\ref{tab:dataset_comparison} compares \ourwork{} with existing
GUI agent training datasets; it is, to our knowledge, the largest
open-source corpus combining programmatic verification with joint
desktop and web coverage.

\subsection{Data Scaling}
\label{sec:data_scaling}

\begin{figure}[t]
\centering
\includegraphics[width=\linewidth]{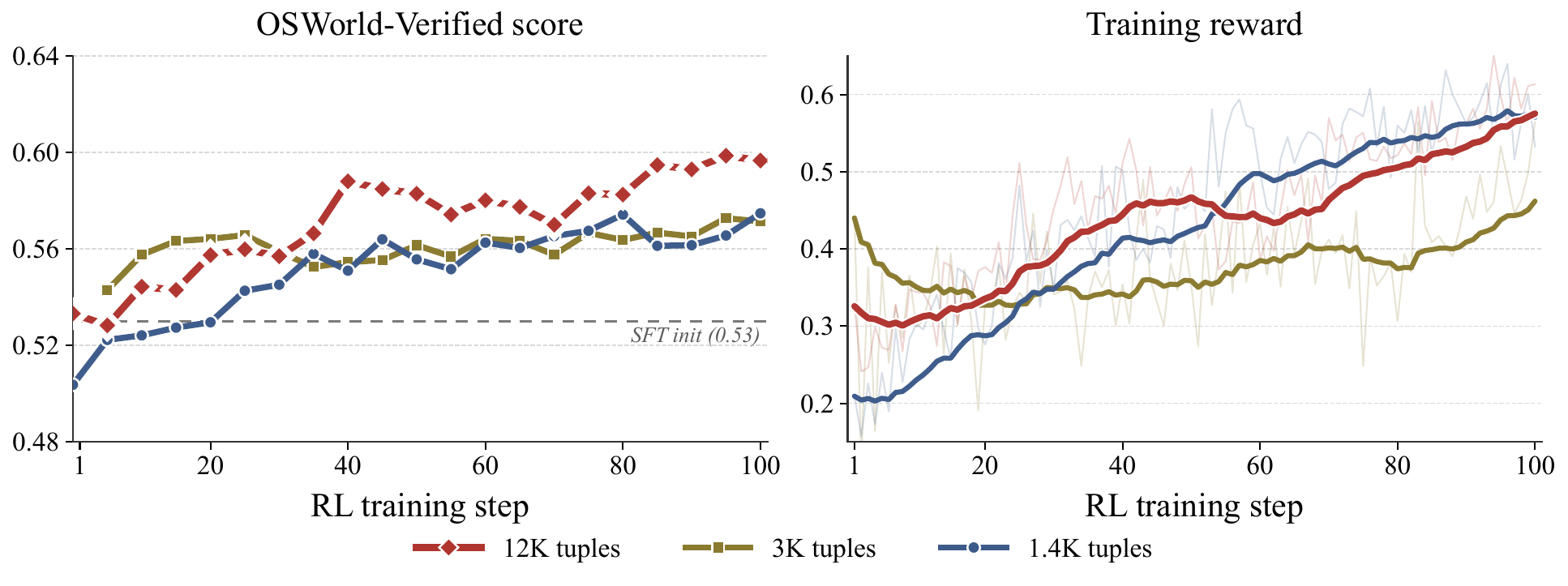}
\caption{\textbf{Data scaling of RL training on \ourwork{}.}
OSWorld-Verified score (left) and training reward (right) vs.\ RL training
step on \smallmodel{} for three \ourwork{} training subsets (1.4K, 3K, 12K
verified tuples), all initialized from the same SFT checkpoint
(gray dashed line at 0.53). Faint lines show raw per-step values; bold
lines are exponentially smoothed.}
\label{fig:data_scaling}
\end{figure}

A central premise of \ourwork{} is that scaling verified RLVR data
yields measurable gains in downstream RL training. We test this
premise by training \smallmodel{} with GSPO on three verified
subsets of \ourwork{} at 1.4K, 3K, and 12K tuples, holding the SFT
initialization and all training hyperparameters fixed across runs.
The resulting training curves additionally serve as a diagnostic on
the information-barrier design of \S\ref{sec:adversarial}, since
exploitable reward functions are known to induce characteristic
instabilities in RL optimization.

\paragraph{Scaling RL data raises the ceiling of RL training.}
The three runs in Figure~\ref{fig:data_scaling} separate along data
scale, with their relative ordering preserved throughout training.
The 12K run attains the highest peak, departs from the SFT baseline
earlier than either smaller run, and sustains a visibly higher band
across the explored window; the 3K and 1.4K runs both plateau
closer to the baseline. The marginal value of additional \ourwork{}
data therefore manifests along two axes simultaneously, the
asymptotic ceiling and the overall shape of the training curve,
indicating that scale produces real RL value rather than diminishing
returns within the regime examined here. The 12K curve further
exhibits no inflection toward saturation, indicating that the
data-scaling regime accessible through this pipeline has not been
exhausted. We also note that all three runs improve monotonically
from the SFT baseline without the oscillation, collapse, or
reward-success decoupling commonly seen in RLVR training, suggesting
that the synthesized rewards are stable enough to support training
across the data scales we examine.

\subsection{Environment Scaling}
\label{sec:env_scaling}

\begin{wrapfigure}{r}{0.48\textwidth}
\vspace{-8pt}
\centering
\includegraphics[width=\linewidth]{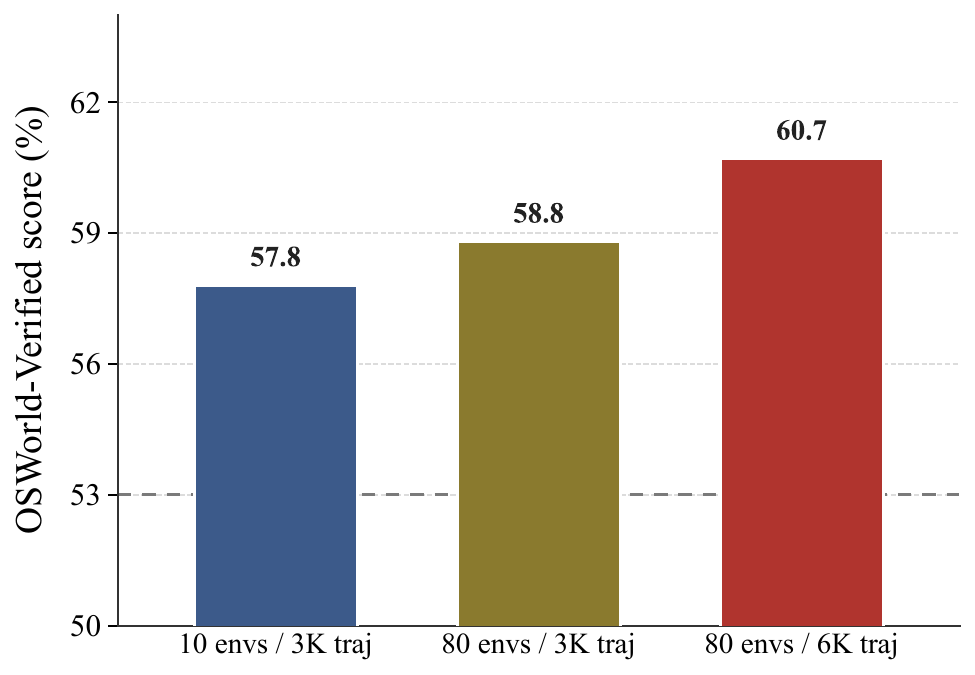}
\caption{\textbf{Environment scaling under teacher distillation.}
OSWorld-Verified score versus number of training environments at 100-step
evaluation budgets. The broad setting (80 envs,
75 trajectories each) outperforms the narrow setting (10 envs, 300
each) despite using $4\times$ fewer trajectories per environment.}
\label{fig:env_scaling}
\vspace{-8pt}
\end{wrapfigure}

The data-scaling results of \S\ref{sec:data_scaling} establish
trajectory volume as one axis along which \ourwork{} data drives RL
performance. A second axis is implicit in the dataset's construction:
the \ourhub{} mock-website synthesis pipeline of \S\ref{sec:env_synthesis}
invests substantial engineering in broadening environment coverage,
on the premise that diversity contributes to downstream performance
beyond what trajectory volume alone can deliver. Whether this premise
holds is an empirical question we address here. The granularity
required for an environment-level ablation makes RL-based evaluation
prohibitively expensive, so we conduct the experiment in a teacher
distillation setup, which has the additional benefit of demonstrating
that \ourwork{} data is consumable under post-training recipes that do
not require RL infrastructure.

\paragraph{Distillation pipeline.}
We roll out Claude-Sonnet-4-6~\citep{claude_sonnet} on \ourwork{} tasks
and retain only trajectories satisfying $r(s, \tau) = 1$, yielding a
filtered pool of teacher demonstrations. The student is the same
\smallmodel{} base used in \S\ref{sec:data_scaling}, fine-tuned on
filtered trajectories under standard SFT. We construct three
environment-scale settings drawn from the \ourhub{} pool: \emph{narrow}
(10 environments, 3K trajectories, 300 per env), \emph{mid}
(80 environments, 3K trajectories, 38 per env), and \emph{broad}
(80 environments, 6K trajectories, 75 per env). Narrow versus mid
isolates environment diversity at fixed total data; mid versus broad
isolates trajectory volume at fixed environment coverage. All three
students share identical SFT hyperparameters and differ only in the
composition of their training set.

\paragraph{Environment diversity and data volume contribute as
complementary axes.}
Figure~\ref{fig:env_scaling} decomposes the scaling effect along the
two ablation axes. Adding environments at fixed data
produces a modest improvement; doubling trajectory volume on the
broadened pool (mid to broad) produces a substantially larger one.
This reading provides
direct empirical support for the synthesis pipeline of
\S\ref{sec:env_synthesis}, whose engineering investment in broadening
environment coverage is justified precisely because trajectory volume
alone cannot substitute for diverse exposure. Put differently,
\ourhub{} contributes reusable environment diversity, while
\ourwork{} instantiates that substrate into verified RLVR data.

\subsection{Emergent Multi-Action Tool Calls}
\label{sec:multi_action}

\begin{wrapfigure}{r}{0.46\textwidth}
\vspace{-8pt}
\centering
\includegraphics[width=\linewidth]{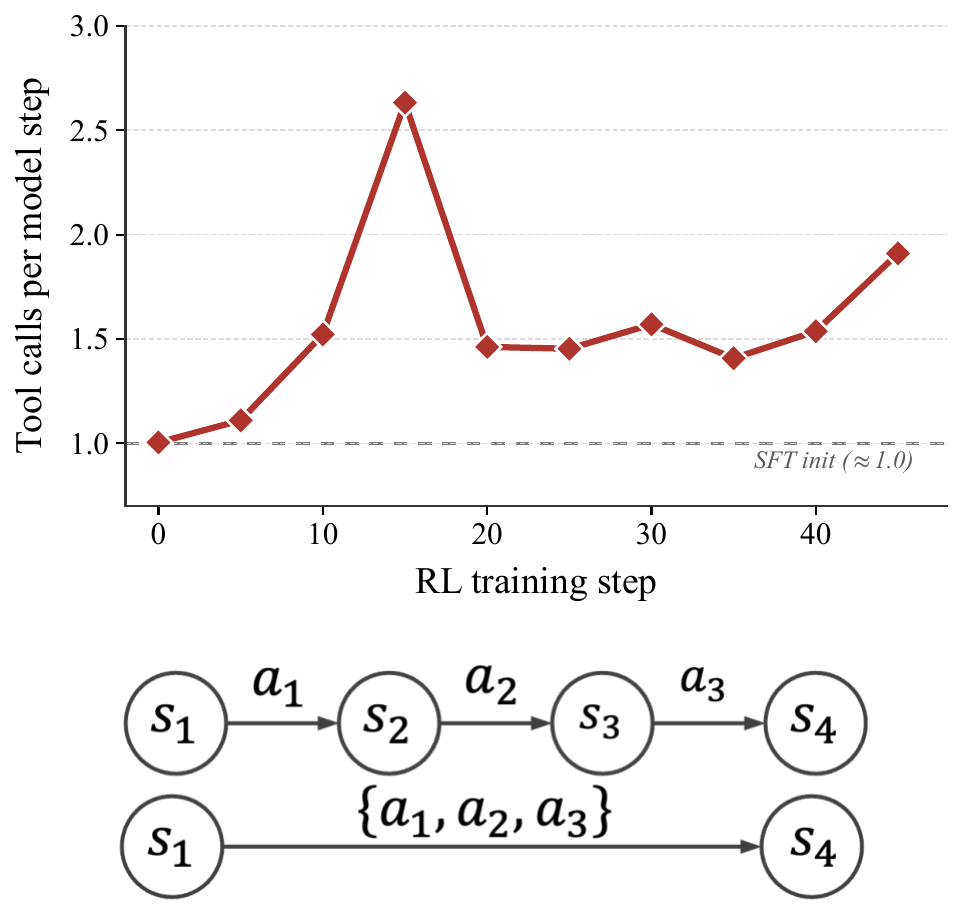}
\caption{\textbf{Emergent action batching during RL.} Average tool
calls per model step on \ourwork{}-A3B. The SFT-initialized policy emits
approximately one call per step; RL drives this to a stable
$1.4$--$1.9$ band.}
\label{fig:tc_per_step}
\vspace{-8pt}
\end{wrapfigure}

A behavior we did not optimize for emerges in \ourwork{} RL training:
the policy spontaneously learns to issue multiple tool calls per
turn, compressing what begin as one-action-per-turn rollouts into
denser multi-action turns. Figure~\ref{fig:tc_per_step} reports a
sustained increase from approximately one call per step at the SFT
initialization to a stable $1.4$--$1.9$ band during RL training, with
trajectory length shortening by 33--45\% at matched task performance.
We attribute this to step-budget pressure under group-normalized
advantage estimation: trajectories that finish within the fixed
per-task budget attain higher relative rewards than those that time
out, so GSPO implicitly selects for policies that pack deterministic
sub-sequences such as menu traversals, form filling, and
keyboard-clipboard chains into single turns. Inspection of
post-training rollouts confirms this pattern. Deterministic action
sequences such as
\texttt{click(File)$\to$click(Export)$\to$click(PDF)} are emitted as
one turn, while actions whose outcomes depend on non-deterministic
state, including network responses and dialogs requiring visual
verification, are conspicuously absent from batched groups. The
policy has internalized a coarse model of which actions safely admit
batching.

The implication of this emergence is twofold. Trajectory shortening
on the order of $40\%$ directly reduces rollout cost during training
and inference latency at deployment, an efficiency gain attributable
purely to RL with no architectural or reward-shaping intervention.
More broadly, the finding demonstrates that structural efficiency
behaviors, beyond task-level competence, can emerge from RL with
verifiable rewards alone. This parallels the spontaneous emergence of
verification and self-reflection in reasoning-focused RL, now observed
in the action-execution domain.

\section{Related Work}
\label{sec:related-work}

\paragraph{Data Synthesis and Post-Training for Digital Agents}
\label{sec:rw-digital-agents}

Synthesizing tasks by perturbing or reverse-engineering artifacts in
executable environments, then verifying outcomes through programmatic
checks, has become the dominant recipe for scalable RLVR data
generation across digital agent
domains~\citep{endless-terminal,swe-smith,swe-universe,pi2026dataengineeringscalingllm,r2e-gym,swe-next}.
RLVE~\citep{zeng2025rlvescalingreinforcementlearning} studies the
complementary direction of scaling RL with adaptive verifiable
environments, further underscoring the importance of environment
generation as a substrate for RLVR.
In software engineering, SWE-smith~\citep{swe-smith} synthesizes bugs
through LM-guided function rewriting and AST-level mutation within
real codebases, R2E-Gym~\citep{r2e-gym} derives tasks from commits
via execution-assisted back-translation, and
SWE-Gym~\citep{swe-gym} scales the paradigm to thousands of
repositories with pre-built execution environments, all using unit
tests as binary reward signals. In terminal operation, Endless
Terminals~\citep{endless-terminal} and
Terminal-Task-Gen~\citep{pi2026dataengineeringscalingllm} procedurally
generate containerized CLI tasks under a dual-validation protocol in
which prerequisite tests must fail initially and pass only upon
correct execution.
The empirical payoff of these pipelines is what justifies their
engineering cost. SWE-smith and SWE-Gym both report log-linear
performance scaling with synthesized task volume that remains
unsaturated at tens of thousands of
tasks~\citep{swe-smith,swe-gym}, and Endless
Terminals~\citep{endless-terminal} observes the same trend on
terminal operation. The consistency of this scaling behavior across
otherwise heterogeneous domains suggests that RLVR data synthesis is
not yet a saturating regime, and motivates analogous infrastructure
for GUI-based computer-use agents, where the data bottleneck is more
acute and no comparable scaling study has been reported.

\paragraph{Task and Environment Synthesis for GUI Agents}
\label{sec:rw-gui-agents}

Extending RLVR data synthesis to GUI agents faces a fundamental
tension between reward verifiability and environment scope. One line
of work uses VLM-based reward estimation to generalize across
applications without environment-specific
instrumentation~\citep{zerogui,autoplay};
ZeroGUI~\citep{zerogui} itself confirms through ablation that
the resulting false-positive rewards destabilize RL training. A second
line synthesizes controllable web environments with code-native reward
oracles, GUI-Genesis~\citep{guigenesis} via per-task Flask
applications, InfiniteWeb~\citep{infiniteweb} via task-centric
test-driven development, and AutoWebWorld~\citep{autowebworld} via
finite-state-machine state-transition models. Yet all remain confined
to browser-based interactions and cannot cover OS-level desktop tasks
or cross-application workflows. Recent attempts to bridge desktop and
verifiability either fall back to checklist-based VLM
evaluation~\citep{gym-anything} or remain narrow in application
coverage~\citep{sea,evocua,agenttrek,videoagenttrek,jedi,computeragentarena}.
OSWorld~\citep{xie2024osworldbenchmarkingmultimodalagents} provides a
complementary evaluation benchmark for real desktop environments.
No existing work simultaneously
achieves deterministic verifiable rewards, broad OS-level application
coverage, and scalable task diversity for GUI agent RLVR training.

% \subsection{Reinforcement Learning for Agent Post-Training}
% \label{sec:rw-rl}

% Applying RL to GUI agents amplifies two challenges absent from
% text-based domains. First, GUI tasks involve long-horizon, multi-turn
% interactions where sparse episode-level rewards provide weak signals
% across tens to hundreds of steps; UI-TARS-2~\citep{ui-tars-2} and
% SEA~\citep{sea} mitigate this through value-model pretraining and
% step-level reward decomposition respectively, but both require
% carefully instrumented environments that resist scaling. Second, GUI
% rollouts incur per-step rendering and visual-processing costs that
% text-based RL avoids. These
% efforts together underscore that effective RL for GUI agents depends
% critically on the availability of scalable environments with
% efficient, verifiable reward signals.

\section{Limitations}
\label{sec:limitations}
\ourwork{} rewards verify terminal environment states rather than the
full process by which an agent reaches them, which keeps reward
authoring scalable but cannot distinguish a clean edit from a
destructive sequence that recreates the same final state; similarly,
the information barrier and forbidden-pattern scan reduce but do not
formally eliminate residual reward-hacking modes such as overly loose
semantic checks or state proxies that correlate with task success.
\ourhub{} mock environments broaden training coverage, yet remain
approximations of real applications: authentication flows, third-party
integrations, network latency, rate limits, and rare server-side
failure states are outside the current mock surface. Finally, our
largest RL runs are reported with single seeds due to their compute
cost, so we interpret the results as evidence that verified task and
environment scale are valuable in the studied regime, not as a final
characterization of the pipeline's ceiling or deployment robustness.

\section{Conclusion}
\label{sec:conclusion}
We present \ourwork{}, an automated pipeline that co-generates task
instructions, environment states, and reward functions for RLVR
training of computer-use agents. The pipeline couples a Generator and
a Discriminator under a process-level information barrier, an
iterative loop that enforces tuple consistency at both endpoints, and
a dataset-level filter that catches failures only visible under
rollout. To broaden the environments accessible to this pipeline, we
synthesize \ourhub{}, a suite of \ourweb{} mock web
applications grounded in O*NET occupational taxonomies and the
Anthropic Economic Index.
On OSWorld-Verified, models trained on \ourwork{} data scale with
data volume without saturating in the regime we examine. The
same checkpoints also improve on the held-out WebArena browser
benchmark.
We read these results as evidence that task and reward diversity
were a substantive bottleneck for current GUI agents rather than as
claims about the pipeline's ultimate ceiling. We release the full
synthesis pipeline, \ourhub{} environments, and generated data so the
community can study both task scaling and environment scaling.

\clearpage

\bibliographystyle{plainnat}
\bibliography{main}

\appendix
\etocdepthtag.toc{mtapp}

% =====================================================
% APPENDIX -- \ourwork{}
% =====================================================

% Local table of contents: hide main paper, expose appendix to subsubsection.
\etocsettagdepth{mtmain}{none}
\etocsettagdepth{mtapp}{subsubsection}

\setcounter{tocdepth}{3}
\setcounter{secnumdepth}{3}
\etocsetnexttocdepth{subsubsection}
\etocsettocstyle{\section*{Appendix Contents}\medskip}{\bigskip}
{\hypersetup{linkcolor=black}\tableofcontents}
\clearpage

% =====================================================
\section{Data Synthesis Pipeline Details}
\label{app:data_synthesis}

\subsection{Pipeline Pseudocode}
\label{app:pseudocode}
Algorithm~\ref{alg:pipeline} summarizes the end-to-end co-generation pipeline. The Orchestrator provisions paired virtual machines and dispatches the Generator and Discriminator under the information barrier of \S\ref{sec:adversarial}. The loop iterates until all five agreement conditions (\S\ref{app:five_conditions}) hold or a max-round budget is exhausted; surviving tuples pass through the dataset-level filter of \S\ref{app:filter}.

\begin{algorithm}[h]
\caption{\ourwork{} co-generation pipeline.}
\label{alg:pipeline}
\begin{algorithmic}[1]
\State \textbf{Input:} task instruction $t$, context $c$, domain skill $\mathcal{S}_\text{dom}$, max rounds $K$
\State \textbf{Output:} verified tuple $(t, s_\text{init}, s_\text{gold}, r)$ or $\bot$
\State Provision isolated VMs $V_\text{init}, V_\text{gold}$
\For{$k = 1, \ldots, K$}
  \State \emph{Generator} reads $(t, c, \mathcal{S}_\text{dom})$ and writes \texttt{initial\_setup.py}, \texttt{golden\_patch.py}
  \State Execute \texttt{initial\_setup.py} on $V_\text{init}$; execute \texttt{golden\_patch.py} on $V_\text{gold}$
  \State \emph{Discriminator} reads $(t, c, \mathcal{S}_\text{dom})$ \textbf{only} (\textit{information barrier}) and writes \texttt{reward.py}
  \State Static scan \texttt{reward.py} against forbidden-pattern list (\S\ref{app:forbidden_patterns})
  \State Evaluate $r_\text{init} \gets r(V_\text{init})$, $r_\text{gold} \gets r(V_\text{gold})$
  \If{all five agreement conditions of \S\ref{app:five_conditions} hold}
    \State \textbf{break}
  \Else
    \State Discriminator emits structured feedback $\mathcal{F}_k \to$ Generator
  \EndIf
\EndFor
\If{convergence not reached} \Return $\bot$ \EndIf
\State Submit $(t, s_\text{init}, s_\text{gold}, r)$ to the dataset-level filter (\S\ref{app:filter})
\State \Return verified tuple if filter accepts, else $\bot$
\end{algorithmic}
\end{algorithm}

\subsection{Task Generation}
\label{app:task_gen}

\subsubsection{Feature Taxonomy Tree Construction}
\label{app:taxonomy_tree}
For each application, we construct a feature taxonomy tree whose internal nodes group related capabilities (e.g., \texttt{libreoffice\_calc} $\to$ \texttt{formatting} $\to$ \texttt{conditional\_formatting}) and whose leaves correspond to atomic UI capabilities. The tree is bootstrapped from official application documentation and an LLM-driven web research pass that grounds the inventory in real-world usage patterns. Trees are constrained to depth $\le 4$ with branching factor $\le 12$ at any node, balancing coverage against sampler tractability. Each leaf is annotated with a one-line capability description and a list of UI elements involved, both of which are surfaced to the Generator at task-synthesis time.

\subsubsection{Scenario Matrix and Sampling}
\label{app:scenario_matrix}
On top of the taxonomy, tasks are sampled across a five-dimensional matrix:
\emph{platform} (desktop / web / cross),
\emph{domain} (10 O*NET-aligned categories),
\emph{difficulty} (easy / medium / hard, with operational definitions: easy $\le 3$ atomic actions, medium $3$--$10$ steps, hard requires long-horizon planning),
\emph{scenario} (commercial / educational / scientific / personal),
and \emph{task type} (single-application / cross-application).
The sampler maintains running per-cell counts and applies inverse-frequency weighting to bias subsequent draws toward under-covered cells. Per-application minima are enforced as hard constraints to prevent the long tail of niche applications from being starved.

\subsubsection{Three-Pass Generation: Breadth, Gap-fill, Edge Cases}
\label{app:three_pass}
Generation proceeds in three sequential passes over the matrix.
\textbf{Pass~1 (breadth)} samples uniformly across the taxonomy to maximize feature coverage; output volume is calibrated so that every leaf receives at least $\tasksperleaf$ tasks.
\textbf{Pass~2 (gap-fill)} re-samples under-represented matrix cells identified by post-hoc coverage analysis on the Pass-1 output. This is where the cross-application and hard-difficulty distributions are deliberately upweighted.
\textbf{Pass~3 (edge cases)} targets boundary conditions identified during the loop-failure analysis of \S\ref{app:filter}: tasks at the limit of single-shot solvability, tasks that combine multiple applications in unusual orderings, and tasks that stress-test rare action primitives (drag, hotkey chords, multi-window navigation).

\subsubsection{Anti-Repetition Rules}
\label{app:anti_repetition}
Within and across passes, generated instructions are de-duplicated via three checks. First, a sentence-embedding cosine similarity threshold ($\ge 0.85$) flags near-duplicates against the running corpus. Second, a token-level $4$-gram overlap test rejects task instructions sharing more than $50\%$ $4$-grams with any prior task. Third, a slot-template diversity rule limits any single instruction template (e.g., ``Apply \texttt{X} formatting to column \texttt{Y}'') to at most $\templatecap$ instantiations per application, forcing the Generator to vary surface form. Tasks failing any check are returned to the Generator for re-synthesis under a "diversify" instruction.

\subsubsection{Coverage Guarantees and Quantitative Targets}
\label{app:coverage}
The sampler imposes the following quantitative targets, enforced as hard constraints:
\begin{itemize}
\item No domain category exceeds $21\%$ of the final dataset (matched empirically in Figure~\ref{fig:overview-data}).
\item Hard tasks comprise at least $40\%$ of the corpus.
\item Cross-application tasks comprise at least $35\%$ of the corpus.
\item Each application receives at least $\tuplesperapp$ verified tuples; no application exceeds $15\%$ of the corpus.
\end{itemize}
Sampler termination occurs when all matrix cells exceed their per-cell minima and total verified tuple count exceeds the project budget.

\subsubsection{Worked Example: Taxonomy Tree for LibreOffice Calc Formatting}
\label{app:taxonomy_example}
The \texttt{libreoffice\_calc} feature taxonomy is constructed by the
Generator from the application's documentation index and ribbon
inventory. Listing~\ref{lst:calc_taxonomy} shows a trimmed version
focused on the \texttt{formatting} subtree --- the rest of the
taxonomy (data import/export, formulas, charts, pivot tables,
sheet structure, protection, references, scripting) is elided for
space. Leaf nodes are scoped tightly enough that a single task
instruction targets one or two adjacent primitives; we list two
illustrative leaf-level tasks per leaf to demonstrate the
within-leaf diversity rule of \S\ref{app:anti_repetition}.

\begin{plainlisting}{Trimmed feature tree: \texttt{libreoffice\_calc} > formatting}
libreoffice_calc/
|-- formatting/
|   |-- cell_appearance/
|   |   |-- number_format/
|   |   |   - "Format column D as currency with the EUR symbol,
|   |   |      two decimals, and a thousands separator."
|   |   |   - "Apply a custom number format to the 'Days Open'
|   |   |      column so values >30 render in red without negative
|   |   |      signs."
|   |   |-- date_time_format/
|   |   |   - "Convert the date column from ISO 'YYYY-MM-DD' to
|   |   |      'DD MMM YYYY' (e.g., 03 Mar 2026)."
|   |   |   - "On the 'Logs' sheet, format column A so timestamps
|   |   |      display the day name plus 24-hour clock, e.g.,
|   |   |      'Mon 14:32'."
|   |   |-- conditional_format/
|   |   |   - "Highlight cells in 'Q4 Revenue' that fall below the
|   |   |      column average using a light-red fill."
|   |   |   - "Add a 3-color gradient (red-yellow-green) across the
|   |   |      'Score' column scaled to its min/max."
|   |   |-- borders_and_fills/
|   |       - "Apply a thin black outer border and dotted inner
|   |          gridlines to the table B2:F18."
|   |       - "Fill alternating rows of the 'Forecast' table with a
|   |          5\% gray to improve readability."
|   |-- alignment_and_wrapping/
|   |   |-- horizontal_vertical/
|   |   |   - "Center-align all header cells in row 1 and right-align
|   |   |      every numeric column."
|   |   |   - "Set vertical alignment to 'middle' for the merged
|   |   |      title cell A1:F1."
|   |   |-- text_wrap_indent/
|   |       - "Enable text wrapping on column C and indent the
|   |          subcategory rows by one level."
|   |       - "Configure column B to shrink-to-fit so long product
|   |          names fit without wrapping."
|   |-- typography/
|   |   |-- font_face_size/
|   |   |   - "Change the 'Summary' sheet to Liberation Sans 11pt
|   |   |      throughout, except keep the title at 16pt bold."
|   |   |   - "Apply a 9pt monospaced font to the 'SQL' column so the
|   |   |      query text aligns visually."
|   |   |-- emphasis/
|   |       - "Bold the totals row, italicize the variance row, and
|   |          underline column headers."
|   |       - "Strike through every line item in column A whose
|   |          status cell equals 'cancelled'."
|   |-- styles_and_themes/
|       |-- named_styles/
|       |   - "Define a custom cell style 'KPI-positive' (green text,
|       |      bold) and apply it to all positive deltas in column G."
|       |   - "Replace every 'Default' style usage on the 'Q3' sheet
|       |      with the 'Heading-2' built-in style."
|       |-- table_styles/
|           - "Convert the range A1:F50 to a styled table with the
|              built-in 'Tabular Light Blue' style."
|           - "Re-skin the 'Forecast' table with a striped style and
|              freeze the first row so the header stays visible."
\end{plainlisting}
\label{lst:calc_taxonomy}

The full \texttt{libreoffice\_calc} tree contains $147$ leaves spread
across $9$ top-level subtrees; \texttt{formatting} alone accounts for
$\sim\!21$ leaves. Pass-1 sampling targets two task instructions per
leaf to satisfy the per-leaf coverage minimum (\S\ref{app:coverage}),
and Pass-2/Pass-3 then upweight underrepresented leaves and difficult
boundary conditions.

\subsection{Adversarial Generator-Discriminator Loop}
\label{app:adv_loop}

\subsubsection{Five Agreement Conditions (Formal Definition)}
\label{app:five_conditions}
A tuple $(t, s_\text{init}, s_\text{gold}, r)$ is accepted by the loop iff all five of the following conditions hold simultaneously:

\begin{enumerate}
\item[\textbf{C1.}] \textbf{Initial-state executability}: \texttt{initial\_setup.py} runs to completion on $V_\text{init}$ without exception.
\item[\textbf{C2.}] \textbf{Golden-state executability}: \texttt{golden\_patch.py} runs to completion on $V_\text{gold}$ without exception.
\item[\textbf{C3.}] \textbf{Golden-state reward}: $r(V_\text{gold}) = 1.0$.
\item[\textbf{C4.}] \textbf{Initial-state reward}: $r(V_\text{init}) = 0.0$.
\item[\textbf{C5.}] \textbf{Reward integrity}: \texttt{reward.py} contains no member of the forbidden-pattern list (\S\ref{app:forbidden_patterns}).
\end{enumerate}

C1--C2 ensure the environment artifacts are well-formed; C3--C4 ensure the reward function is sensitive in the correct direction at the two endpoint states; C5 ensures the reward verifies task semantics rather than artifact identity. The conjunction of all five is necessary and sufficient for loop-level acceptance, and is checked by the Orchestrator from the Discriminator's structured \texttt{REVIEW.md} verdict.

\subsubsection{Information Barrier: Sandbox and Access Matrix}
\label{app:info_barrier}
The information barrier is implemented as process-level isolation between the Generator and Discriminator processes. Each process executes in a separate working directory with disjoint file-system views; inter-process communication is mediated solely by the Orchestrator. Table~\ref{tab:access_matrix} summarizes the access rights enforced.

\begin{table}[h]
\centering
\footnotesize
\setlength{\tabcolsep}{6pt}
\renewcommand{\arraystretch}{1.05}
\begin{tabular}{@{}lcc@{}}
\toprule
\textbf{Resource} & \textbf{Generator} & \textbf{Discriminator} \\
\midrule
Task instruction $t$, context $c$              & \checkmark & \checkmark \\
Domain skill file $\mathcal{S}_\text{dom}$     & \checkmark & \checkmark \\
\texttt{initial\_setup.py} (Generator output)  & write      & \textbf{denied} \\
\texttt{golden\_patch.py} (Generator output)   & write      & \textbf{denied} \\
Generator working directory                    & full       & \textbf{denied} \\
$V_\text{init}$ post-setup, via state-only API & read       & read \\
$V_\text{gold}$ post-patch, via state-only API & read       & read \\
File-system contents of $V_\text{init}$, $V_\text{gold}$ & write & \textbf{denied} (read via API only) \\
\texttt{reward.py} (Discriminator output)      & \textbf{denied} & write \\
\texttt{REVIEW.md} (verdict + feedback)        & read       & write \\
\bottomrule
\end{tabular}
\caption{Access matrix enforcing the information barrier between Generator and Discriminator. The Discriminator's only task-specific signal is the natural-language pair $(t, c)$ and a state-only view of the two virtual machines, eliminating the reverse-engineering shortcut described in \S\ref{sec:adversarial}.}
\label{tab:access_matrix}
\end{table}

\subsubsection{Forbidden Pattern List for Anti-Hacking Static Scan}
\label{app:forbidden_patterns}
The Discriminator-emitted \texttt{reward.py} is scanned at write-time against the following pattern list. Any match aborts the round and triggers a structured re-prompt to the Discriminator with the matched pattern as feedback.

\begin{enumerate}
\item \textbf{Direct Boolean assignment} to a verification flag without computation, e.g., \texttt{chart\_verified = True}.
\item \textbf{Placeholder verification}: a flag is assigned a constant before being conditionally added to the score, with no intervening evaluation.
\item \textbf{Hardcoded success}: a function returns a constant in $\{0.5, 1.0\}$ along the success path with no inspection of the environment.
\item \textbf{Bare existence scoring}: a positive score is awarded purely on \texttt{os.path.exists(...)} without any property check on the file's contents.
\item \textbf{\texttt{subprocess} usage}: the reward shells out to external processes, which are non-reproducible and easily spoofed.
\item \textbf{Comment-only verification}: a score increment is preceded by a comment asserting a check (\texttt{\# assume X is correct}) without code performing the check.
\end{enumerate}

The list is enforced by a combination of regex matching and Python AST traversal; full source patterns are released with the code.

\subsubsection{Loop Termination, Max Rounds, and Feedback Protocol}
\label{app:loop_protocol}
The loop is capped at $K = 5$ rounds. Each unsuccessful round produces a structured \texttt{REVIEW.md} from the Discriminator listing the failing condition(s), the observed reward values $(r_\text{init}, r_\text{gold})$, the matched forbidden pattern (if any), and a free-text recommendation. The Generator consumes \texttt{REVIEW.md} as its sole feedback signal and revises its scripts. If round $K$ also fails, the tuple is rejected and logged for offline analysis. We deliberately avoid increasing $K$: empirically, tuples that fail to converge by round $5$ are dominated by ambiguous task instructions and unsolvable specifications that no amount of script revision will fix; surfacing these to the gap-fill pass (\S\ref{app:three_pass}) is more cost-effective than additional inner-loop iterations.

\iffalse  % Empirical convergence breakdown deferred: pending production-run log audit.
\subsubsection{Empirical Convergence Behavior}
\label{app:convergence}
\stub{Histogram of rounds-to-convergence over the released dataset, plus a per-failure-mode breakdown for the rejected tail. To be filled with measurements from the production run logs.}
\fi

\subsection{Filter Module}
\label{app:filter}

\subsubsection{LLM Majority Voting: Voters, Rubric, and Aggregation}
\label{app:majority_vote}
Loop-accepted tuples enter a dataset-level filter implemented as an LLM majority vote across an ensemble of $V$ heterogeneous critics. Each critic receives the four-tuple $(t, s_\text{init}, s_\text{gold}, r)$, emits a structured JSON verdict over the schema $\langle$\texttt{verdict}, \texttt{severity}, \texttt{can\_fix\_with\_query\_only}, \texttt{query\_issues}, \texttt{setup\_reward\_risks}, \texttt{training\_pool\_fit}, \texttt{confidence}, \texttt{reasoning\_summary}, \texttt{revised\_query}$\rangle$, and operates under the conservative-by-default decision policy of the prompt in \S\ref{app:prompt_filter}. The ensemble comprises voters drawn from independent foundation-model families to reduce family-specific bias correlations.

\paragraph{Severity rubric (per voter).}
\begin{itemize}
\item \textbf{P0} (reject): fatal setup / reward / environment problem; mismatch between $(t, s, r)$ that cannot be repaired by query revision.
\item \textbf{P1} (modify-query): the query is missing essential context, contains harmful ambiguity, or leaks process in a way that biases evaluation.
\item \textbf{P2} (keep-or-modify): non-fatal quality issues; the tuple is usable but not ideal.
\item \textbf{P3} (keep): no meaningful issue beyond surface style.
\end{itemize}

\paragraph{Aggregation rule.}
A tuple is admitted to the next filter stage iff a strict majority ($\lceil V/2 \rceil + 1$) of voters return \texttt{verdict}\,$\in$\,\{\texttt{keep}, \texttt{modify\_query}\}. Among admitted tuples, the canonical instruction is selected by majority vote over voter \texttt{verdict}: if the majority is \texttt{modify\_query}, the released instruction is the highest-confidence \texttt{revised\_query} among modifying voters; if the majority is \texttt{keep}, the original instruction is retained. Ties default to rejection. The ensemble's per-tuple severity is reported as the maximum severity assigned by any voter, providing a worst-case-aware quality grade in the released metadata.

\subsubsection{Teacher-Model Rollout Verification with VLM-as-a-Judge}
\label{app:teacher_rollout}
Critic-admitted tuples additionally pass through a teacher-rollout verification stage. We roll out a strong teacher policy (Claude-Sonnet-4-6~\citep{claude_sonnet}) on each tuple for $N_\text{teach}$ trials, scoring each rollout twice: the programmatic reward $r(s, \tau) \in [0, 1]$ implemented in \texttt{reward.py}, and a VLM-as-a-judge score $\hat{r}_\text{vlm}(s, \tau) \in \{0, 1\}$ produced by a separate VLM judge that consumes the final-state screenshot, the task instruction, and a checklist of acceptance criteria distilled from $t$.

\paragraph{Why two scores.}
The two scores measure different aspects of correctness. $r$ is grounded in environment state (file contents, structured-state diffs, library introspection) and exhibits low variance but only sees what the reward author chose to instrument. $\hat{r}_\text{vlm}$ is grounded in the final visual state and catches surface-level failures the programmatic reward misses (e.g., a UI that loaded with the right data but in the wrong color theme, a chart whose values are correct but axis labels are missing). Concordance between the two is a strong signal that the reward function is faithful to the task's intent; persistent disagreement is a sign that either the reward is too narrow or the task itself is under-specified.

\paragraph{Inclusion criteria.}
Tuples are categorized by $(\bar{r}, \bar{\hat{r}}_\text{vlm})$ averaged across the $N_\text{teach}$ trials:
\begin{itemize}
\item Both means in $(0, 1)$ with $|\bar{r} - \bar{\hat{r}}_\text{vlm}| < \delta$: \emph{accepted}.
\item Both means equal $0$: \emph{plausibly unsolvable}; removed from the corpus.
\item Both means equal $1$ on the first trial: \emph{plausibly trivial}; down-sampled to control the easy-tail.
\item Means disagree by $\ge \delta$: \emph{flagged}; routed back to the Discriminator with the disagreement region highlighted as feedback for reward refinement.
\end{itemize}

The accepted set forms the released dataset. Disagreement-flagged tuples that survive Discriminator refinement re-enter teacher rollout; tuples that fail to converge across two refinement passes are dropped to the offline-analysis bucket.

\subsubsection{Per-Stage Filter Yield}
\label{app:filter_yield}
The two filter stages reject distinct failure modes and contribute roughly 70/30 to total filter loss. Starting from the loop-accepted pool, the LLM majority-voting stage (\S\ref{app:majority_vote}) rejects approximately $3{,}100$ tuples on P0 setup/reward defects, P1 query issues, or persistent voter disagreement; this stage primarily catches static specification errors that survive the loop's agreement conditions but fail end-to-end critique. The subsequent teacher-rollout stage (\S\ref{app:teacher_rollout}) rejects an additional $1{,}278$ tuples that the teacher cannot solve in $N_\text{teach}$ trials or that score trivially on the first trial; this stage primarily catches infeasibility and triviality that only manifest under realistic agent behavior. The final corpus is the \ourdata{} verified tuples reported in the main text.

\subsection{Domain Skill Files}
\label{app:skill_files}

\subsubsection{SKILL.md Standard Structure}
\label{app:skill_structure}
Each application domain ships with a \texttt{SKILL.md} document loaded by both Generator and Discriminator at synthesis time. The standard structure has six sections: (1) domain-specific concepts and Python libraries (e.g., \texttt{python-docx} for \texttt{libreoffice\_writer}, \texttt{openpyxl} for \texttt{libreoffice\_calc}); (2) state and file-system layout, including canonical paths for documents and configuration; (3) \texttt{initial\_setup.py} templates parameterized by the most common task contexts; (4) \texttt{golden\_patch.py} templates with corresponding parameter slots; (5) \texttt{reward.py} scoring patterns illustrating progressive partial-credit decomposition; (6) ``bitter lessons'', a curated list of pitfalls observed during development (e.g., set-vs-set\_current confusion in mock APIs, style-inheritance gotchas in document libraries).

\subsubsection{Auto-Generation Pipeline for SKILL.md}
\label{app:skill_autogen}
Initial SKILL.md drafts are produced from a one-shot research pass over the application's documentation, then refined by inspecting the rejection logs of the first generation pass and surfacing recurring failure patterns as new ``bitter lesson'' entries. SKILL.md files are versioned alongside the synthesis pipeline; all empirical results in this paper use the version frozen at release.

\subsubsection{Example: Excerpt of a Domain SKILL.md}
\label{app:skill_example}
We reproduce excerpts from the \texttt{libreoffice\_calc} SKILL.md to ground the format. The frontmatter, GUI-startup template (Section~0), and the \emph{Bitter Lessons} list (Section~6) are reproduced in full; the file-creation, file-reading, and reward-pattern sections (1, 4, 5) are abbreviated for space.

\paragraph{Frontmatter and role.}
\begin{plainlisting}{libreoffice\_calc SKILL.md frontmatter}
---
name: libreoffice-calc
description: How to programmatically create, modify, and verify
             LibreOffice Calc (.xlsx) files using Python openpyxl,
             for setup-gen and reward-gen.
---

# LibreOffice Calc -- Python Manipulation Guide

This skill teaches setup-gen (create/modify xlsx) and reward-gen
(read/verify xlsx) how to work with spreadsheet files using pure
Python code.

  - Library: openpyxl  (+ pandas for bulk data)
  - Install: pip3 install openpyxl pandas
\end{plainlisting}

\paragraph{Section 0 -- GUI startup template.}
After generating the initial workbook, \texttt{initial\_setup.py} must leave the task in a GUI-ready state. Because GUI applications launched without an explicit display fail silently on the headless VM, every launch sets \texttt{DISPLAY=:0} and uses non-blocking \texttt{Popen} so that the script can exit cleanly:

\begin{pythonlisting}{Section 0 -- GUI startup helper}
import os, shlex, subprocess, time

def launch_gui(command: str, delay_sec: float = 1.0):
    env = os.environ.copy()
    env["DISPLAY"] = ":0"
    subprocess.Popen(
        shlex.split(command),
        stdout=subprocess.DEVNULL,
        stderr=subprocess.DEVNULL,
        env=env,
    )
    time.sleep(delay_sec)

# Open the initial workbook for the GUI agent.
launch_gui(f'libreoffice --calc "{output_path}"', delay_sec=2.0)

# Optional multi-app startup, e.g., for cross-app tasks.
# launch_gui('nautilus "/home/user"', delay_sec=1.0)
\end{pythonlisting}

\paragraph{Sections 1--5 (abbreviated).}
Section 1 catalogues the openpyxl API surface used by \texttt{initial\_setup.py} and \texttt{golden\_patch.py}: workbook construction, cell-level writes, bulk row writes, sheet creation and reordering, font / fill / alignment / border styling, number formats, merged cells, formulas, charts (bar, line, pie, scatter), data validation, autofilter ranges, and conditional formatting. Section 2 covers the reading API used by \texttt{reward.py}: loading workbooks with and without \texttt{data\_only=True}, iterating cells, verifying styles, reading conditional-formatting rules, and detecting merged-cell regions. Section 4 contains progressive-scoring reward templates parameterized by task type (data-edit, formatting, formula, chart, layout); Section 5 contains targeted helper functions for common verification idioms (column-extraction with type coercion, partial-equality checks, numeric tolerance comparisons). Full text accompanies the released pipeline.

\paragraph{Section 6 -- Bitter Lessons.}
\begin{plainlisting}{Section 6 -- Bitter lessons learned during deployment}
1.  Formula values are NOT computed by openpyxl. cell.value returns
    the formula string "=SUM(A1:A10)", not the result. Use
    data_only=True to get the last-cached value (requires the file
    to have been opened in Calc/Excel at least once).

2.  Always use 8-char ARGB for colors. PatternFill(start_color=
    "4472C4") silently becomes "004472C4" (alpha=00, transparent).
    Write "FF4472C4". On read, compare against the 8-char form.

3.  fgColor is the visible background, not bgColor. cell.fill
    .fgColor.rgb gives you the background color you see; bgColor is
    rarely what you want.

4.  Merged cells: only the top-left has data. After
    merge_cells("A1:D1"), B1/C1/D1 become MergedCell with value=None.
    Style the top-left cell only.

5.  Copy-then-modify for golden files. Never recreate from scratch
    if an initial file exists. shutil.copy(initial, golden) then
    open and modify. This preserves metadata, print settings, and
    other invisible properties.

6.  Pivot tables cannot be created by openpyxl; only read/preserved.
    Use a template file with the pivot already built.

7.  Filters are definition-only. ws.auto_filter.ref = "A1:D20" sets
    the range, but rows are NOT actually hidden. Filtering only takes
    effect when the file is opened in LibreOffice.

8.  showDropDown=False means SHOW the dropdown. In DataValidation,
    this boolean is inverted. Set False to display the dropdown arrow.

9.  Chart type naming. BarChart(type="col") is a vertical column
    chart; BarChart(type="bar") is horizontal. Does not match
    LibreOffice's menu names.

10. Styles are immutable after assignment. You cannot do
    cell.font.bold = True. Create a new Font object:
    cell.font = Font(bold=True, ...). Same for Fill, Alignment, Border.

11. Theme colors may return None for .rgb. If a cell uses a theme
    color instead of explicit RGB, cell.font.color.rgb can be None
    or a theme index. Wrap color reads in try/except.

12. data_only=True loses formulas. Loading with data_only=True
    replaces formulas with their cached values. If you need both,
    load the file twice -- once normally, once with data_only=True.
\end{plainlisting}

The bitter-lessons section is the most distinctive feature of the SKILL.md format. Each entry is born from a specific debugging episode in early development, and each is the kind of pitfall that an LLM Generator or Discriminator would otherwise rediscover at the cost of a wasted adversarial round. Bitter lessons are versioned alongside the pipeline; new entries are added whenever a recurring failure mode surfaces in the rejection logs.

\subsection{Output Bundle and OSWorld Compatibility}
\label{app:output_bundle}

\subsubsection{File Layout per Verified Task}
\label{app:file_layout}
Each verified tuple is emitted as a self-contained directory under \texttt{output/final/<task\_id>/} with the following layout:

\begin{plainlisting}{Per-task output bundle layout}
output/final/<task_id>/
  config.json          # OSWorld evaluator contract
  meta.json            # task taxonomy fields, difficulty, source pass
  initial_setup.py     # produces s_init on a fresh VM
  golden_patch.py      # produces s_gold on a fresh VM
  reward.py            # implements r: state -> [0, 1]
  REVIEW.md            # Discriminator verdict from the final accepted round
  task_config.json     # natural-language task instruction + context
\end{plainlisting}

This layout is consumed without modification by the OSWorld
evaluator~\citep{xie2024osworldbenchmarkingmultimodalagents}, by our
SFT teacher rollout pipeline, and by the RL trainer of
\S\ref{app:training}.

\subsubsection{config.json Schema and Evaluator Contract}
\label{app:config_schema}
\texttt{config.json} follows the OSWorld evaluator contract verbatim, with fields for VM image, snapshot identifier, action observation type (\texttt{a11y\_tree} or \texttt{screenshot}), step budget, and a list of post-execution evaluator entry points pointing into our \texttt{reward.py}. We add no schema extensions; full compatibility with the upstream OSWorld evaluator~\citep{xie2024osworldbenchmarkingmultimodalagents} was an explicit design goal so that \ourwork{} tuples are usable as drop-in OSWorld benchmark items.

% =====================================================
\section{Mock Web Environment Synthesis}
\label{app:mock_synthesis}

\subsection{Environment Selection Methodology}
\label{app:env_selection}

\subsubsection{Mapping O*NET SOC Major Groups to Web Application Categories}
\label{app:onet_mapping}
The mock-application suite is grounded in the Standard Occupational Classification (SOC) major groups maintained by O*NET~\citep{onet}. Each SOC major group corresponds to a distinct cluster of digital workflows; we map these clusters onto application categories that span the modal software encountered in each occupation. Table~\ref{tab:onet_mapping} summarizes the mapping. The mapping is many-to-many: an SOC major group may touch multiple application categories (e.g., Management uses both CRM and project-management tools), and an application category often serves multiple SOC groups (e.g., communication apps appear across nearly all knowledge-work occupations).

\begin{table}[h]
\centering
\footnotesize
\setlength{\tabcolsep}{6pt}
\renewcommand{\arraystretch}{1.05}
\begin{tabular}{@{}p{4.6cm}p{8cm}@{}}
\toprule
\textbf{O*NET SOC major group} & \textbf{Mock application categories represented} \\
\midrule
Management Occupations                       & CRM, project management, communication, analytics \\
Business and Financial Operations            & Accounting, payments, marketing analytics, CRM \\
Computer and Mathematical                    & Code hosting, CI/CD, monitoring, cloud consoles, infra \\
Architecture and Engineering                 & Cloud infrastructure, CAD/diagram, project management \\
Life, Physical, and Social Science           & Document collaboration, paper review, analytics \\
Community and Social Service                 & Communication, scheduling, document collaboration \\
Legal Occupations                            & Document signing, case management, legal research \\
Educational Instruction and Library          & Learning management, document collaboration, scheduling \\
Arts, Design, Entertainment, Sports          & Design tools, social media, content publishing \\
Healthcare Practitioners and Technical       & EHR systems, medical imaging, scheduling \\
Sales and Related                            & CRM, e-commerce admin, marketing analytics \\
Office and Administrative Support            & Email, scheduling, document collaboration, HRIS \\
Transportation and Material Moving           & Booking, travel planning, e-commerce \\
Government and Civil Service                 & Government portals, tax, identity verification \\
\bottomrule
\end{tabular}
\caption{Mapping from O*NET SOC major groups to mock-application categories represented in \ourwork{}. Knowledge-work-heavy occupations (Management, Business, Computer, Sales) receive proportionally more environments, reflecting their relative software intensity in the AEI weighting (\S\ref{app:aei_weighting}).}
\label{tab:onet_mapping}
\end{table}

\subsubsection{Anthropic Economic Index Weighting}
\label{app:aei_weighting}
Within each SOC group, application priorities are weighted by the Anthropic Economic Index~\citep{aei}, which provides empirical software-usage frequencies estimated from large-scale agent-traffic logs. We extract a per-category usage frequency $f_c$, normalize across the categories selected for inclusion, and use the resulting weights $w_c = f_c / \sum_{c'} f_{c'}$ as the per-category synthesis budget. The result is a mock-application allocation that approximates the relative time a knowledge worker spends across application categories rather than the application diversity of any single occupation.

\subsubsection{Coverage Cutoff Criteria}
\label{app:cutoff}
We synthesize a mock for category $c$ only if its weight $w_c$ exceeds an inclusion threshold $w_\text{min}$, calibrated such that the long tail of low-weight categories ($<1\%$ of weighted usage) does not consume engineering budget that could be better spent on category depth. Within an included category, individual applications are selected in decreasing order of usage frequency until the per-category budget is exhausted. This produces a long-tail truncation that is qualitatively conservative: the released suite covers the head of the distribution at high fidelity rather than the full tail at low fidelity.

\subsection{Complete List of Synthesized Mocks}
\label{app:mock_list}

\subsubsection{Application-by-Application Inventory}
\label{app:mock_inventory}
\begin{figure}[p]
\centering
\setlength{\fboxsep}{0pt}\setlength{\fboxrule}{0.3pt}
\makeatletter
\expandafter\def\csname mockpath@gmail\endcsname{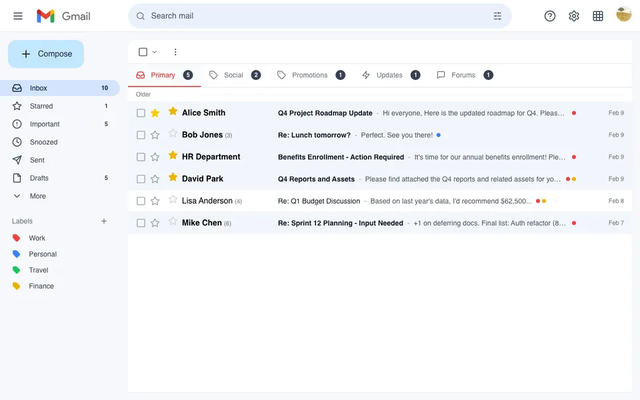}
\expandafter\def\csname mockpath@slack\endcsname{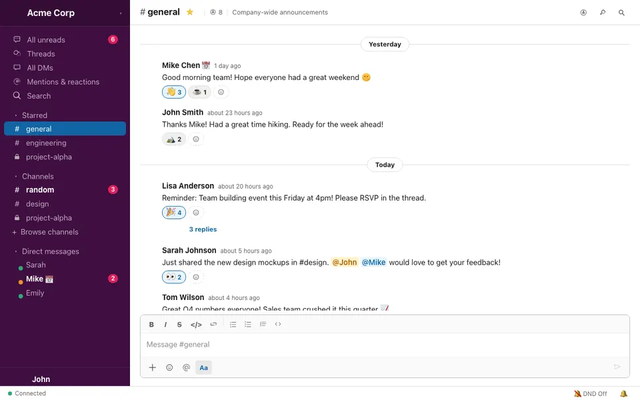}
\expandafter\def\csname mockpath@discord\endcsname{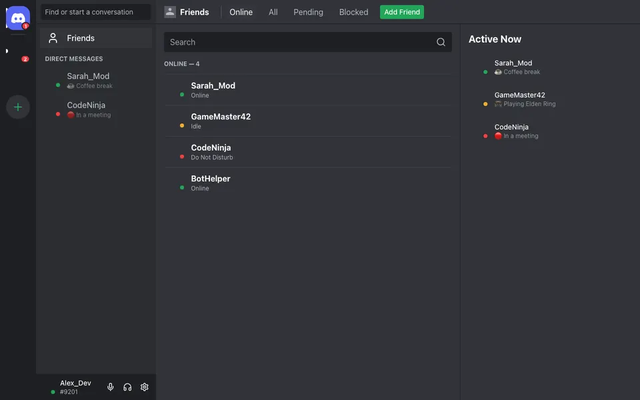}
\expandafter\def\csname mockpath@zoom-web\endcsname{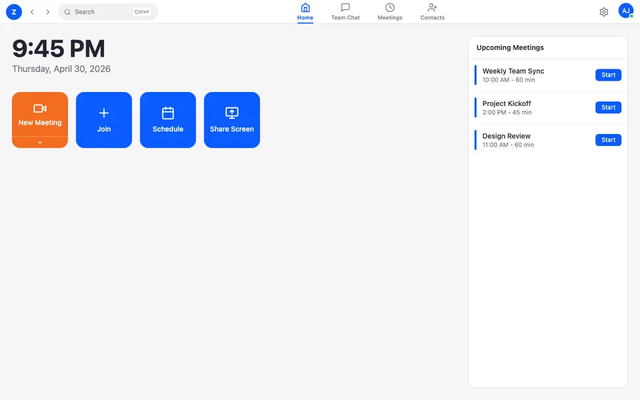}
\expandafter\def\csname mockpath@microsoft-teams\endcsname{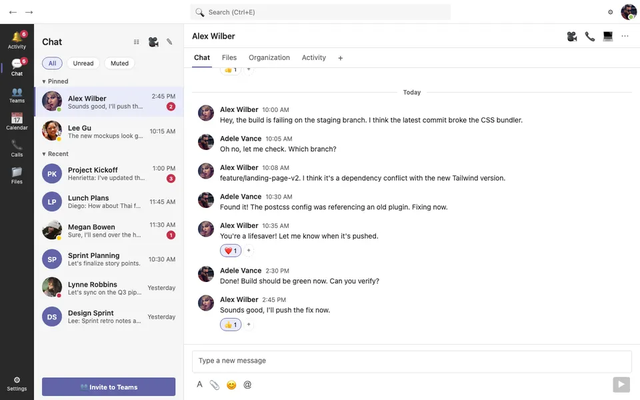}
\expandafter\def\csname mockpath@outlook-web\endcsname{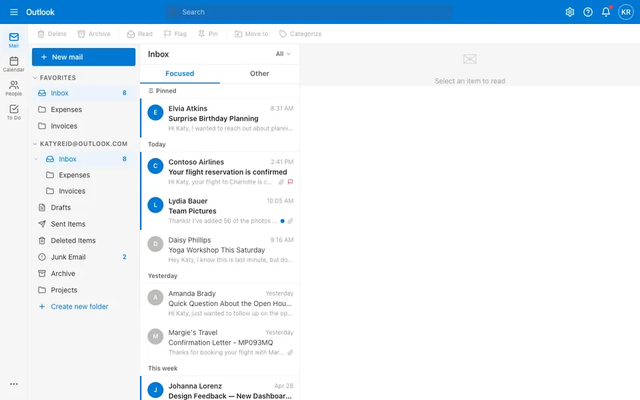}
\expandafter\def\csname mockpath@dingtalk\endcsname{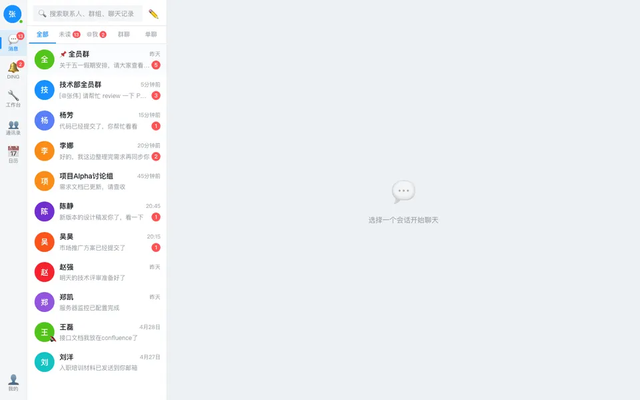}
\expandafter\def\csname mockpath@feishu\endcsname{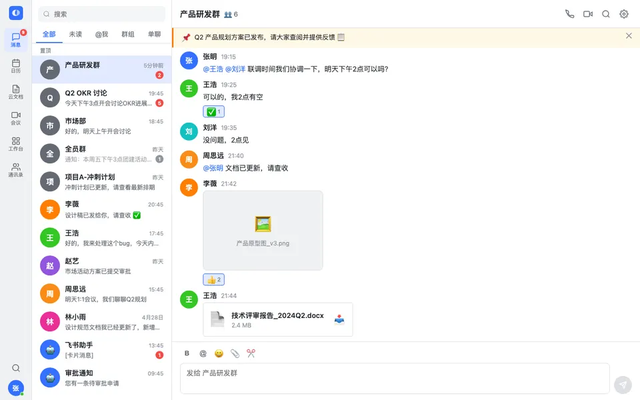}
\expandafter\def\csname mockpath@notion\endcsname{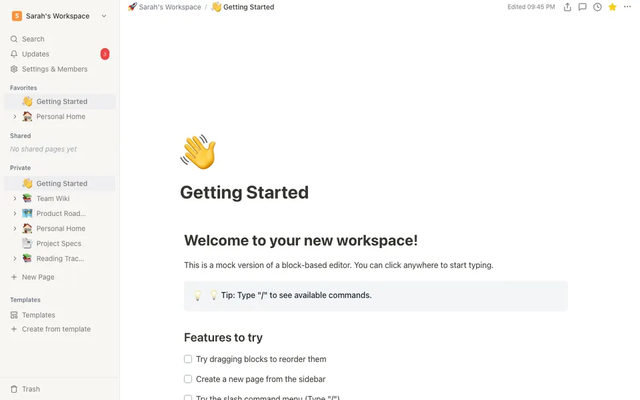}
\expandafter\def\csname mockpath@google-docs\endcsname{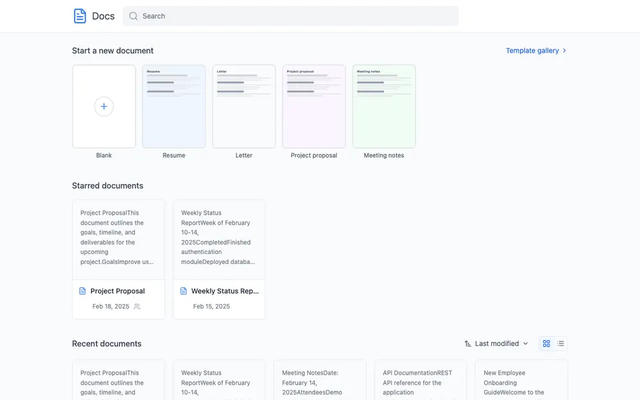}
\expandafter\def\csname mockpath@google-sheets\endcsname{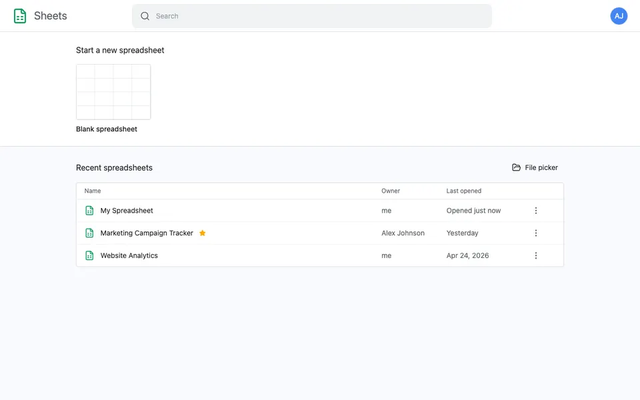}
\expandafter\def\csname mockpath@miro\endcsname{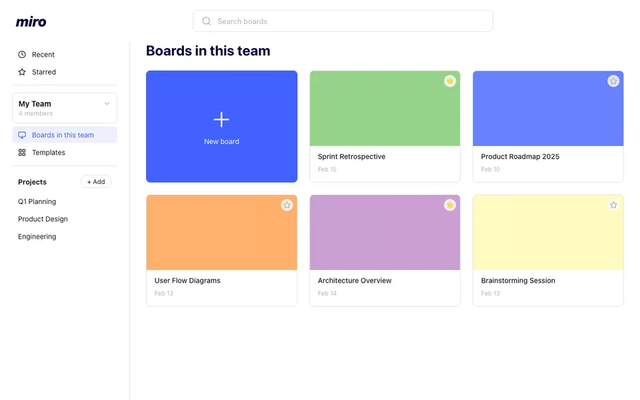}
\expandafter\def\csname mockpath@figma\endcsname{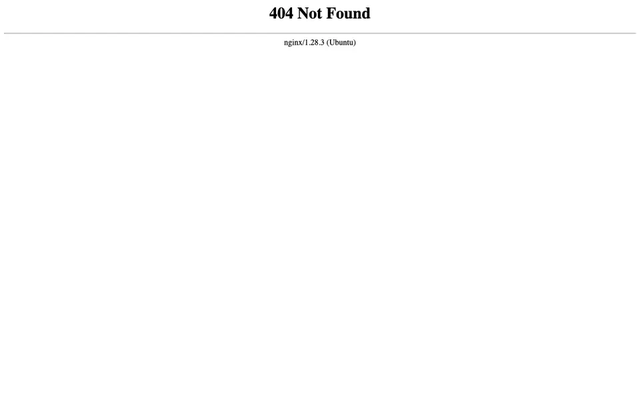}
\expandafter\def\csname mockpath@lucidchart\endcsname{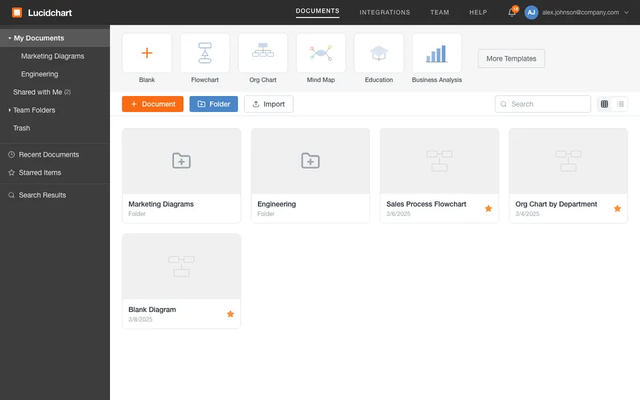}
\expandafter\def\csname mockpath@monday\endcsname{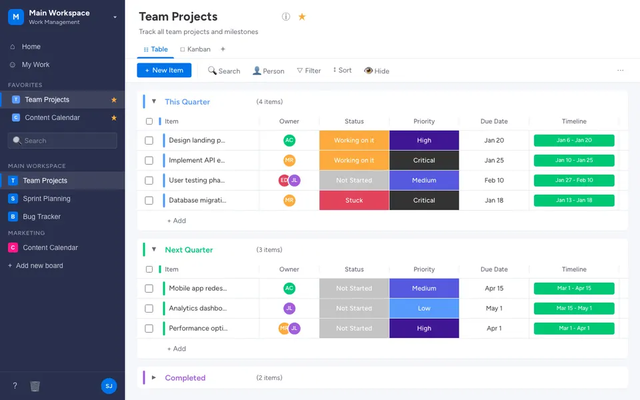}
\expandafter\def\csname mockpath@trello\endcsname{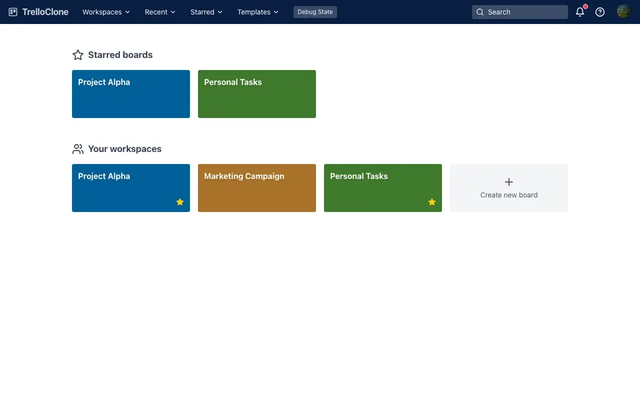}
\expandafter\def\csname mockpath@github\endcsname{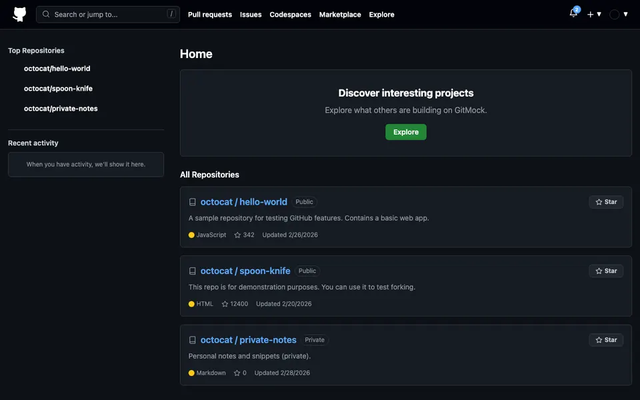}
\expandafter\def\csname mockpath@gitlab\endcsname{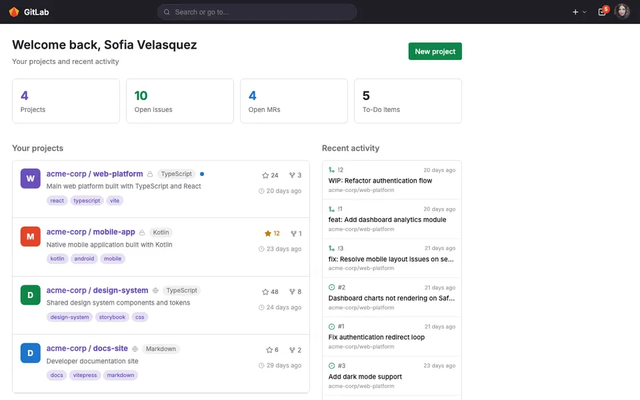}
\expandafter\def\csname mockpath@jira\endcsname{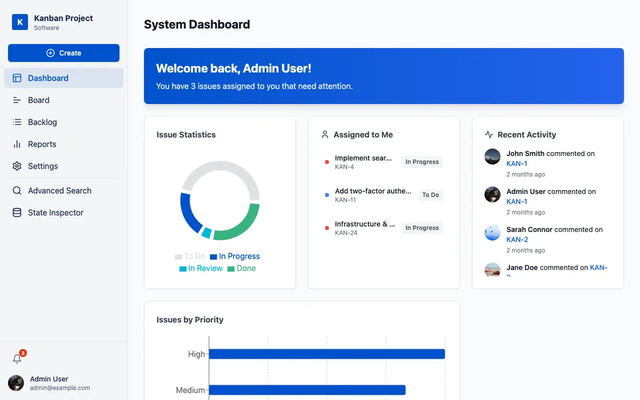}
\expandafter\def\csname mockpath@linear\endcsname{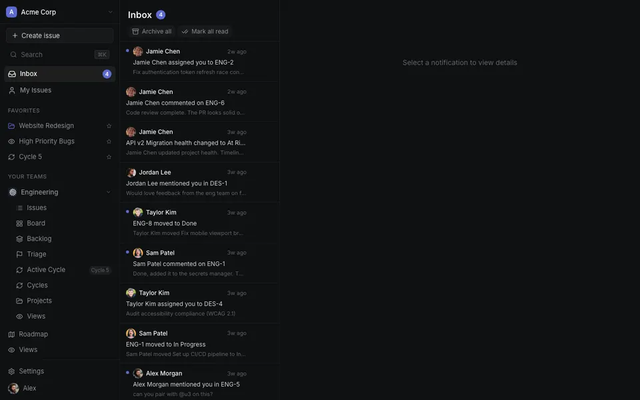}
\expandafter\def\csname mockpath@sentry\endcsname{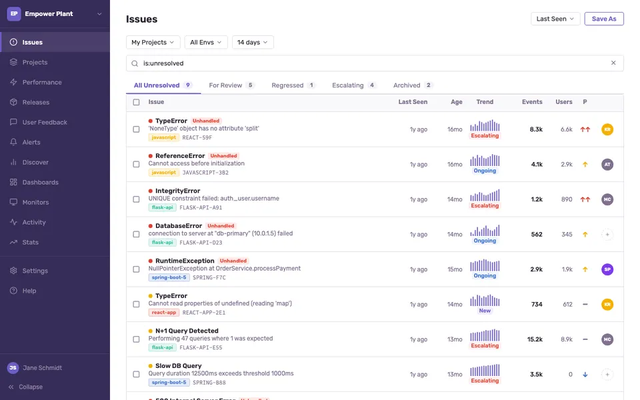}
\expandafter\def\csname mockpath@datadog\endcsname{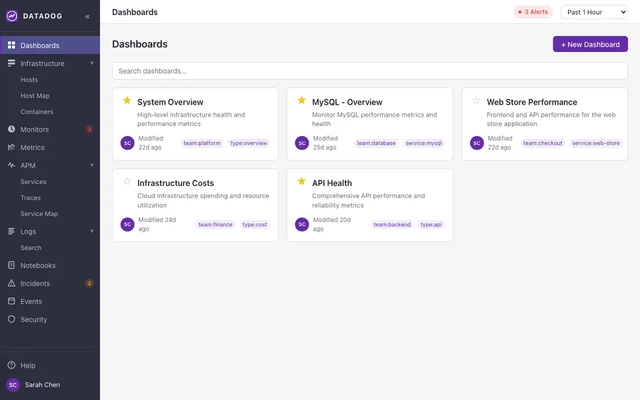}
\expandafter\def\csname mockpath@postman\endcsname{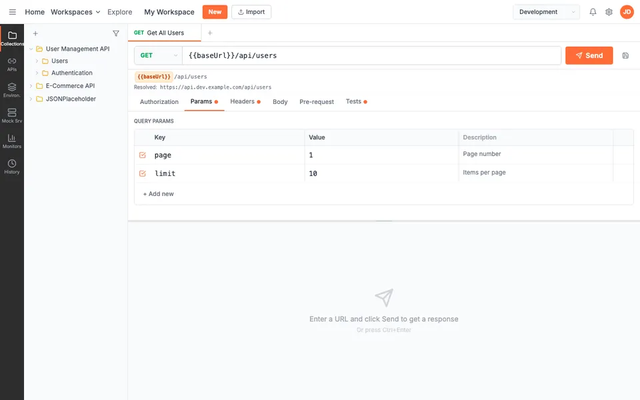}
\expandafter\def\csname mockpath@vercel\endcsname{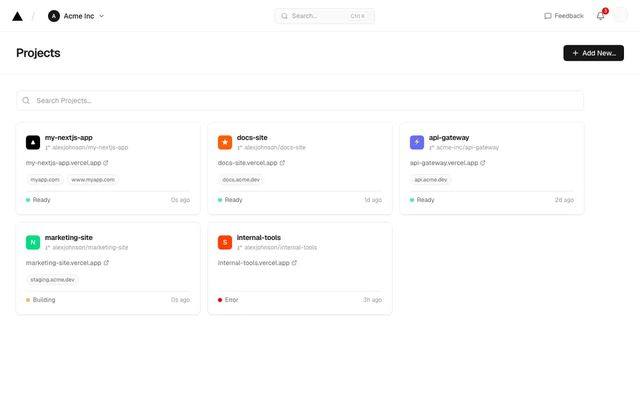}
\expandafter\def\csname mockpath@salesforce\endcsname{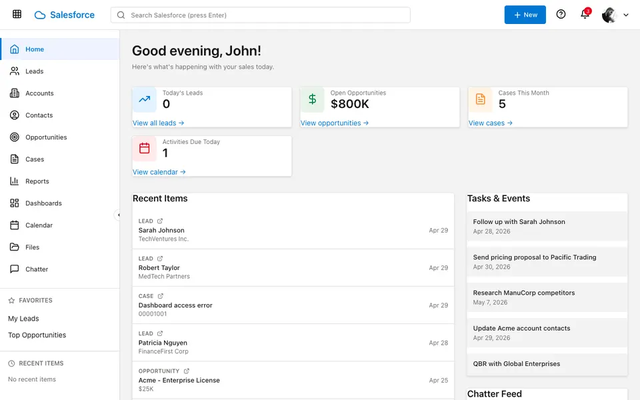}
\expandafter\def\csname mockpath@shopify-admin\endcsname{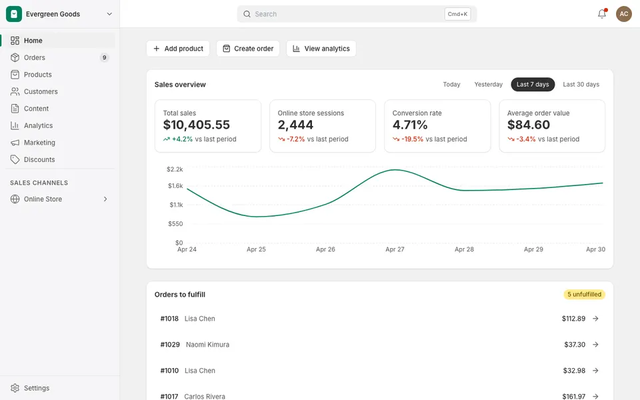}
\expandafter\def\csname mockpath@stripe-dashboard\endcsname{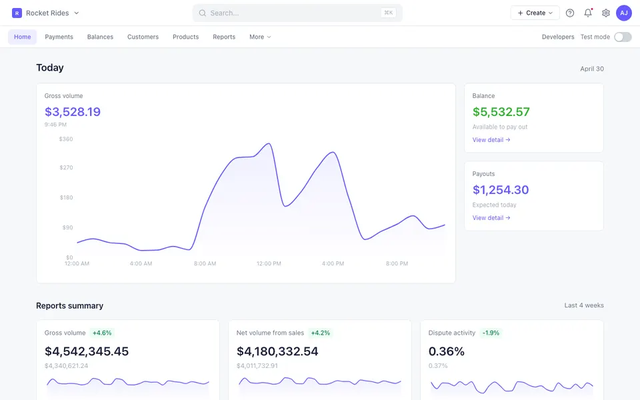}
\expandafter\def\csname mockpath@amazon\endcsname{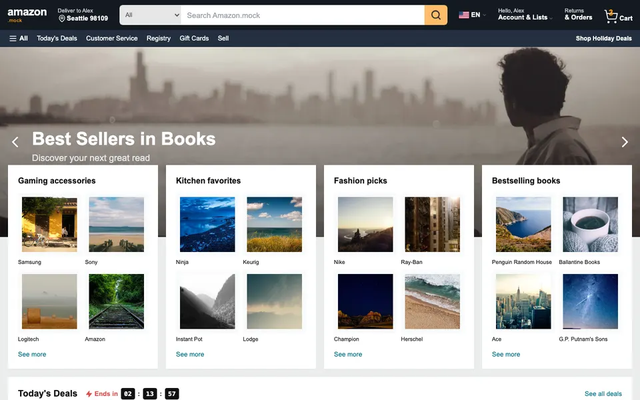}
\expandafter\def\csname mockpath@aws-console\endcsname{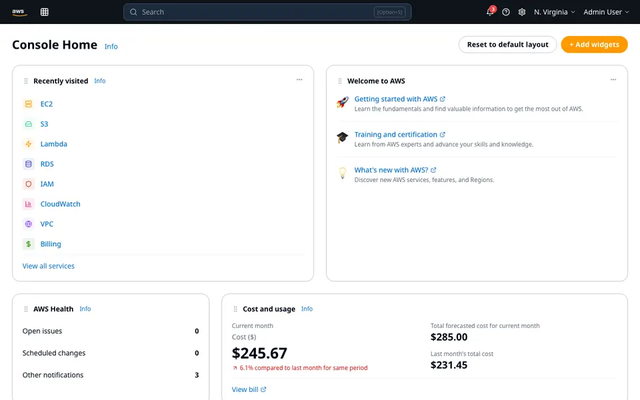}
\expandafter\def\csname mockpath@azure\endcsname{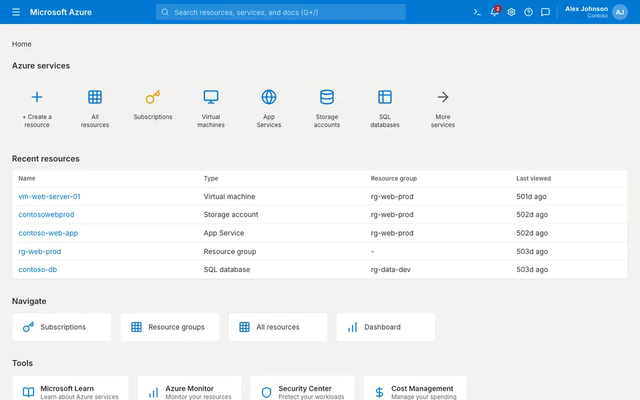}
\expandafter\def\csname mockpath@linkedin\endcsname{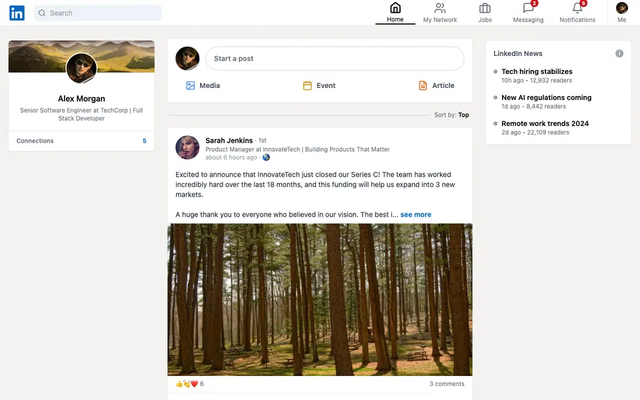}
\expandafter\def\csname mockpath@youtube\endcsname{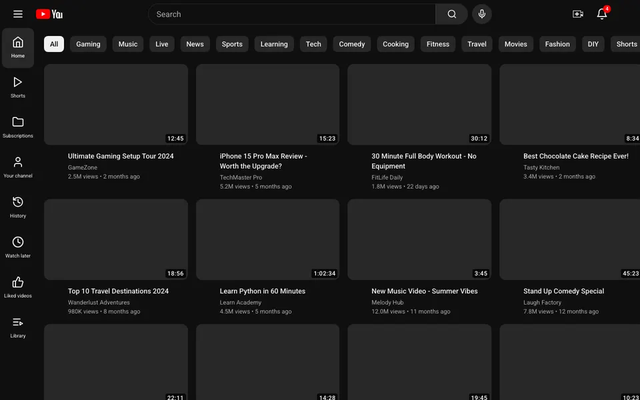}
\newcommand{\mockcell}[1]{%
  \begin{subfigure}[t]{0.235\linewidth}
    \centering
    \edef\mockpath{\csname mockpath@#1\endcsname}%
    \fbox{\includegraphics[width=\linewidth]{\mockpath}}
    \subcaption*{\footnotesize\texttt{#1}}
  \end{subfigure}}
\makeatother
\mockcell{gmail}\hfill
\mockcell{slack}\hfill
\mockcell{discord}\hfill
\mockcell{zoom-web}\\[6pt]
\mockcell{microsoft-teams}\hfill
\mockcell{outlook-web}\hfill
\mockcell{dingtalk}\hfill
\mockcell{feishu}\\[6pt]
\mockcell{notion}\hfill
\mockcell{google-docs}\hfill
\mockcell{google-sheets}\hfill
\mockcell{miro}\\[6pt]
\mockcell{figma}\hfill
\mockcell{lucidchart}\hfill
\mockcell{monday}\hfill
\mockcell{trello}\\[6pt]
\mockcell{github}\hfill
\mockcell{gitlab}\hfill
\mockcell{jira}\hfill
\mockcell{linear}\\[6pt]
\mockcell{sentry}\hfill
\mockcell{datadog}\hfill
\mockcell{postman}\hfill
\mockcell{vercel}\\[6pt]
\mockcell{salesforce}\hfill
\mockcell{shopify-admin}\hfill
\mockcell{stripe-dashboard}\hfill
\mockcell{amazon}\\[6pt]
\mockcell{aws-console}\hfill
\mockcell{azure}\hfill
\mockcell{linkedin}\hfill
\mockcell{youtube}
\caption{Landing-page screenshots of $32$ representative released mock
web applications, organized by thematic pair-of-rows:
communication and meetings (rows~1--2), productivity and
collaboration (rows~3--4), engineering and developer tooling
(rows~5--6), and commerce / cloud / social (rows~7--8). Each mock
reproduces the surface visual layout, navigation tree, and
primary-feature inventory of its named reference application
without authentication, third-party integrations, or outbound
network calls (\S\ref{app:fidelity}). The full inventory of
\numweb{} released mocks is enumerated in
Table~\ref{tab:mock_inventory}.}
\label{fig:mock_grid}
\end{figure}

Table~\ref{tab:mock_inventory} lists the \numweb{} synthesized mock applications grouped by category. Each mock retains the canonical real-world application name as its identifier; the implementation is a self-contained single-page application with no dependency on the corresponding real service. Figure~\ref{fig:mock_grid} shows representative landing-page screenshots of $32$ mocks spanning the major scenario categories, illustrating the visual fidelity range of the released suite.

\begin{longtable}{@{}p{3.4cm}p{10cm}@{}}
\caption{Full inventory of synthesized mock web applications, grouped by category.}
\label{tab:mock_inventory} \\
\toprule
\textbf{Category} & \textbf{Mock applications} \\
\midrule
\endfirsthead
\toprule
\textbf{Category} & \textbf{Mock applications} \\
\midrule
\endhead
\bottomrule
\endfoot
Communication / messaging & \texttt{slack}, \texttt{microsoft\_teams}, \texttt{discord}, \texttt{dingtalk}, \texttt{feishu}, \texttt{wechat}, \texttt{zoom\_web}, \texttt{gmail}, \texttt{outlook\_web} \\
Social / publishing       & \texttt{twitter}, \texttt{instagram}, \texttt{facebook}, \texttt{linkedin}, \texttt{reddit}, \texttt{weibo}, \texttt{xiaohongshu}, \texttt{zhihu}, \texttt{pinterest}, \texttt{youtube} \\
Project management        & \texttt{jira}, \texttt{asana}, \texttt{trello}, \texttt{monday}, \texttt{linear}, \texttt{notion} \\
Document collaboration    & \texttt{google\_docs}, \texttt{google\_sheets}, \texttt{google\_drive}, \texttt{google\_calendar}, \texttt{airtable}, \texttt{miro}, \texttt{lucidchart}, \texttt{openreview} \\
Code hosting / DevOps     & \texttt{github}, \texttt{gitlab}, \texttt{vercel}, \texttt{circleci}, \texttt{sentry}, \texttt{postman}, \texttt{wandb} \\
Cloud / infrastructure    & \texttt{aws\_console}, \texttt{azure}, \texttt{aliyun}, \texttt{cloudflare}, \texttt{datadog} \\
E-commerce / marketplace  & \texttt{amazon}, \texttt{ebay}, \texttt{taobao\_seller}, \texttt{amazon\_seller}, \texttt{shopify\_admin}, \texttt{woocommerce}, \texttt{instacart}, \texttt{uber\_eats} \\
Travel                    & \texttt{booking\_com}, \texttt{expedia}, \texttt{tripadvisor}, \texttt{google\_flights}, \texttt{12306} \\
Finance / payments        & \texttt{paypal}, \texttt{robinhood}, \texttt{coinbase}, \texttt{stripe\_dashboard}, \texttt{quickbooks}, \texttt{TradingView} \\
HR / hiring               & \texttt{workday}, \texttt{bamboohr}, \texttt{greenhouse}, \texttt{gusto}, \texttt{lattice}, \texttt{adp} \\
CRM / customer service    & \texttt{salesforce}, \texttt{hubspot}, \texttt{Zendesk}, \texttt{ServiceNow}, \texttt{SAP} \\
Marketing / ads           & \texttt{hubspot\_marketing}, \texttt{klaviyo}, \texttt{mailchimp}, \texttt{meta\_ads}, \texttt{google\_ads} \\
Analytics                 & \texttt{google\_analytics}, \texttt{mixpanel}, \texttt{amplitude}, \texttt{hotjar}, \texttt{looker\_studio}, \texttt{tableau} \\
Healthcare / legal        & \texttt{epic-health}, \texttt{PACS-viewer}, \texttt{clio}, \texttt{westlaw}, \texttt{contractbook}, \texttt{docusign}, \texttt{Expensify} \\
Government / education    & \texttt{california\_tax}, \texttt{usa-gov}, \texttt{visa\_portal\_ds160}, \texttt{Canvas-LMS} \\
Real estate / utilities   & \texttt{zillow} \\
\end{longtable}

\subsubsection{Real-World Reference Mapping and Fidelity Notes}
\label{app:fidelity}
Each mock retains the surface-level visual layout, navigation tree, and primary-feature inventory of its named reference application. Mocks differ from references in three deliberate ways: (i) authentication is removed (no sign-in flow, no OAuth, no SSO); (ii) all data is local to the session and seeded synthetically; (iii) outbound network calls are replaced by in-process state mutations. Where the reference application exposes capabilities that depend on these stripped layers (e.g., real-time push notifications, cross-account sharing, payment processing), the mock instead exposes a deterministic stub that supports task verification without network dependence.

\subsubsection{Trademark, Trade Dress, and Asset Compliance}
\label{app:compliance}
The mock identifiers used in Table~\ref{tab:mock_inventory} and Figure~\ref{fig:mock_grid} (e.g., \texttt{slack}, \texttt{gmail}, \texttt{shopify\_admin}) are \emph{internal} development labels that pin each synthesis target to its real-world reference for grounding the Plan and Web agents. The publicly released artifact applies the following compliance steps to avoid trademark, trade-dress, and asset-licensing concerns:
\begin{itemize}
\item \textbf{Renaming to generic identifiers.} At release, every mock is renamed to a generic, non-confusing label that describes its functional category rather than the reference product (e.g., \texttt{slack} $\to$ \texttt{team-chat-mock}, \texttt{gmail} $\to$ \texttt{mail-mock}, \texttt{shopify\_admin} $\to$ \texttt{ecommerce-admin-mock}). The release package ships a mapping table between internal names and public identifiers, and all task instructions, file paths, and \texttt{config.json} entries reference only the public identifiers.
\item \textbf{No third-party logos or proprietary brand assets.} No real logos, brand-specific icon sets, or proprietary illustration assets are embedded in the released mocks. Reference logos viewed during the Plan Agent's research pass are used only as ephemeral inputs to the LLM and are not stored in the mock source tree. Brand-distinctive color palettes (e.g., a specific shade of purple, orange, or blue identifying a real product) are replaced with neutral palettes drawn from a curated 12-color set.
\item \textbf{Independently authored UI copy and illustrations.} Marketing copy, microcopy, placeholder illustrations, and marketing imagery are written from scratch by the Dev Agent rather than being copied from reference applications. The navigation tree and feature inventory are functionally inspired by references but do not reproduce verbatim text strings or visual elements that would constitute trade dress.
\item \textbf{Reference screenshots are not redistributed.} Reference screenshots collected by the Plan Agent during web research are kept locally for the duration of synthesis and are not included in any released artifact. Figure~\ref{fig:mock_grid} renders the released \emph{mock} screenshots, all of which were produced by our Dev Agent.
\item \textbf{TOS-sensitive mocks are flagged for additional review.} Mocks corresponding to applications with restrictive terms of service (financial services, healthcare, government portals) are flagged in the release manifest for additional human review prior to public release; in cases where the reference application's TOS forbid even functional reproduction, the mock will be withheld from the public release and replaced by a generic substitute in the same category.
\item \textbf{Compliance manifest at release.} We will publish a per-mock disclosure document alongside the dataset that lists, for each released mock, (a)~the public identifier, (b)~the functional category, (c)~the compliance steps applied, and (d)~the result of TOS-sensitive review. This manifest will be included in the release package and tracked in the dataset card.
\end{itemize}
The internal-name convention is retained in this paper to preserve readability of the inventory and to keep the cross-references in \S\ref{app:env_pipeline}, \S\ref{app:mock_cases}, and the released task identifiers traceable; readers should treat all real-product names as internal labels for the corresponding generic mock category.

\subsubsection{Per-Mock LOC and Endpoint Statistics}
\label{app:mock_stats}
We measure each released mock along three structural axes:
source-code size (LOC, summed over JavaScript/TypeScript/CSS/HTML
files under \texttt{src/}, excluding \texttt{node\_modules} and
build artifacts), route count (\texttt{<Route>} declarations in the
SPA's React Router config), and entity count (top-level entities
in the in-memory data model declared in \texttt{SCHEMA.md} or
\texttt{src/utils/dataManager.js}). The released production suite
contains \ourweb{} mocks; the source tree additionally retains
$5$ templated scaffolds (used as starting points for new mocks)
that are excluded from the released task corpus. Aggregate
statistics over the full $99$-directory measurement set are
reported in Table~\ref{tab:mock_size_stats}; the LOC bucket
histogram is in Table~\ref{tab:mock_loc_buckets}. Restricting to
the \ourweb{} released mocks shifts the means by less than
$3\%$ on every axis (the scaffolds sit in the very low-LOC bin),
so we report the inclusive numbers for transparency.

\begin{table}[h]
\centering
\footnotesize
\setlength{\tabcolsep}{8pt}
\renewcommand{\arraystretch}{1.0}
\begin{tabular}{@{}lrrrrrrrr@{}}
\toprule
\textbf{Metric} & \textbf{mean} & \textbf{p10} & \textbf{p25} & \textbf{p50} & \textbf{p75} & \textbf{p90} & \textbf{p99} & \textbf{max} \\
\midrule
Source LOC          & 6{,}127 & 4{,}237 & 4{,}828 & 5{,}663 & 7{,}196 & 9{,}180 & 13{,}095 & 13{,}095 \\
Route components    & 15.0 & 4 & 8 & 14 & 20 & 27 & 60 & 60 \\
Data-model entities & 5.9 & 3 & 3 & 5 & 7 & 10 & 23 & 23 \\
\bottomrule
\end{tabular}
\caption{Structural statistics across the $99$ measured mock applications.}
\label{tab:mock_size_stats}
\end{table}

\begin{table}[h]
\centering
\footnotesize
\setlength{\tabcolsep}{10pt}
\begin{tabular}{@{}lrr@{}}
\toprule
\textbf{Source LOC} & \textbf{Mocks} & \textbf{\%} \\
\midrule
$<\!1$K           &  1 &  1.0 \\
$1$K--$2.5$K      &  1 &  1.0 \\
$2.5$K--$5$K      & 31 & 31.3 \\
$5$K--$10$K       & 59 & 59.6 \\
$10$K--$20$K      &  7 &  7.1 \\
$\geq\!20$K       &  0 &  0.0 \\
\bottomrule
\end{tabular}
\caption{Distribution of mock-application source-LOC. The bulk
of mocks ($91\%$) sit in the $2.5$K--$10$K LOC band, consistent
with the design constraint of a self-contained SPA implementing
roughly $10$--$20$ navigable views with a small typed in-memory
schema. Outliers above $10$K LOC (e.g., \texttt{aws\_console\_mock},
\texttt{Canvas-LMS\_mock}, \texttt{expedia\_mock}) cover broader
feature surfaces; the two smallest mocks are templated scaffolds
retained for ancestry rather than as task targets.}
\label{tab:mock_loc_buckets}
\end{table}

\subsection{Unified State API Specification}
\label{app:state_api}

\subsubsection{Endpoint Contracts}
\label{app:endpoints}
Every mock implements a four-endpoint state API that supports state injection, state inspection, and reset. The API is implemented as a Vite middleware plugin co-resident with the SPA, requires no separate process, and is reachable via the same origin as the SPA itself.

\paragraph{\texttt{POST /post?sid=<sid>}} -- state injection and lifecycle control. Body: \texttt{\{ "action": <"set" | "set\_current" | "reset" | "merge">, "state": <object>, "merge": <bool> \}}. Action semantics:
\begin{itemize}
\item \texttt{set}: writes \texttt{state} to the session's \emph{initial} snapshot. Used by \texttt{initial\_setup.py}.
\item \texttt{set\_current}: writes \texttt{state} to the session's \emph{current} snapshot, leaving the initial snapshot untouched. Used by \texttt{golden\_patch.py} and rarely by \texttt{initial\_setup.py} when an in-progress state is needed.
\item \texttt{merge}: same as \texttt{set\_current}, but performs a deep merge into the existing current snapshot rather than replacing it.
\item \texttt{reset}: clears both initial and current snapshots and reloads the default seed data.
\end{itemize}
Response: \texttt{\{ "success": true, "sid": <sid>, "state\_id": <hash> \}}.

\paragraph{\texttt{GET /go?sid=<sid>}} -- structural state inspection. Response: \texttt{\{ "initial\_state": <object>, "current\_state": <object>, "state\_diff": <object> \}}. The \texttt{state\_diff} is a flat key-path map from changed fields to \texttt{\{ "old": <v>, "new": <v> \}} pairs, computed as detailed in \S\ref{app:state_diff}. This is the primary interface used by \texttt{reward.py}.

\paragraph{\texttt{GET /state?sid=<sid>}} -- raw current state. Response: \texttt{\{ "stored\_state": <object>, "has\_custom\_state": <bool>, "sid": <sid> \}}. Provided as a lower-level inspection endpoint when the diff abstraction is unnecessary.

\paragraph{\texttt{POST /upload?sid=<sid>}} -- file upload, used by tasks that require user-supplied attachments. Body: \texttt{multipart/form-data}; response includes a per-file URL that the SPA can resolve. Files are scoped to the \texttt{sid} and discarded on session reset.

\subsubsection{Session Isolation via sid}
\label{app:sid}
Every API call carries a session identifier \texttt{sid} as a query parameter, scoping all state reads and writes to a per-session storage namespace. The trainer generates a fresh \texttt{sid} (UUIDv4) for each rollout and persists it to a known file path inside the VM, so that all subsequent \texttt{initial\_setup.py}, agent-issued, and \texttt{reward.py} calls share the same session. Sessions are independent: concurrent rollouts on the same mock instance do not interfere. Session state has a TTL of one hour past the last access; expired sessions are garbage-collected without intervention.

\subsubsection{State Diff Computation Rules}
\label{app:state_diff}
The \texttt{state\_diff} returned by \texttt{/go} is computed as a key-path comparison between the initial and current snapshots, with three rules tuned for reward authoring. Object fields are compared structurally (recursive descent on nested objects). Arrays are treated as ordered sequences and a difference at any index marks the entire array as changed; this avoids false partial-equality matches when the agent reorders rather than edits. Whitelisted volatile fields (e.g., \texttt{lastViewedAt}, computed UI-state caches) are masked before diff to prevent passive viewing actions from polluting the diff. The diff representation is intentionally flat (key-path strings) rather than nested, because it is consumed by reward authors who write linear assertions over individual fields.

\subsubsection{Integration with reward.py}
\label{app:reward_integration}
The canonical \texttt{reward.py} pattern for web-mock tasks fetches \texttt{/go} once, then evaluates a sequence of partial-credit assertions against \texttt{state\_diff}, accumulating a $[0, 1]$ score:

\begin{pythonlisting}{Canonical reward.py pattern for web-mock tasks}
import requests

def verify_task():
    sid = open('/tmp/task_web_sid').read().strip()
    response = requests.get(f'http://localhost:8080/go?sid={sid}').json()
    diff = response['state_diff']
    score = 0.0
    if 'channels[0].name' in diff and diff['channels[0].name']['new'] == 'engineering':
        score += 0.25
    if 'messages.engineering[0].content' in diff:
        score += 0.25
    # ... additional assertions ...
    print(f'REWARD: {score}')
    return score
\end{pythonlisting}

\subsection{Multi-Agent Synthesis Pipeline}
\label{app:env_pipeline}

\subsubsection{Plan Agent: Web Research and DESIGN.md}
\label{app:plan_agent}
The Plan Agent is responsible for understanding the target application before any code is written. Given an application name, it performs structured web research (documentation crawl, screenshot collection, feature inventory across user-role personas) and emits four artifacts: \texttt{DESIGN.md} (color palette, typography, spacing tokens, component styles), \texttt{assets/README.md} (UI layout descriptions and primary user workflows), \texttt{assets/data\_model.md} (entity definitions for the in-memory state), and \texttt{TODO.md} (a prioritized P0/P1/P2 work queue for the Dev Agent). The role specification explicitly excludes authentication flows, real network communication, and persistence beyond \texttt{localStorage}; the mock's contract is to be a faithful interactive sandbox, not a functional clone of the reference service.

\subsubsection{Dev Agent: Implementation Protocol}
\label{app:dev_agent}
The Dev Agent consumes the Plan Agent's artifacts and implements the mock as a Vite + React single-page application following a fixed project layout (\texttt{src/App.jsx} for routing, \texttt{src/context/AppContext.jsx} for global state, \texttt{src/utils/dataManager.js} for state initialization and \texttt{localStorage} persistence, \texttt{src/utils/stateTracker.js} for diff computation, and \texttt{vite.config.js} for the state-API plugin). Coordination with the Plan and Web Agents is purely file-based via the artifacts of \S\ref{app:plan_agent} and the test reports of \S\ref{app:web_agent}; no direct messaging is permitted. Priority for any given work cycle follows the order: AUDIT P0 issues $>$ TEST P0 issues $>$ AUDIT P1 issues $>$ TODO P0 items $>$ TODO P1 items $>$ TODO P2 items, ensuring code-correctness regressions are addressed before new feature work.

\subsubsection{Web Agent: Playwright Verification against UI Tree}
\label{app:web_agent}
The Web Agent operates a headless Playwright browser against the deployed mock and verifies the UI tree against \texttt{DESIGN.md} and \texttt{assets/README.md}. It exercises every interactive element catalogued in \texttt{TODO.md}, performs visual diff against reference screenshots when available, and produces \texttt{TEST.md} (functional and visual bug reports) and \texttt{AUDIT.md} (code-level issues such as dead handlers, untracked state mutations, and missing entries in the diff API). Reports use a P0/P1/P2 severity that mirrors the Dev Agent's priority order, ensuring that fixes propagate in a deterministic sequence.

\subsubsection{Iterative Convergence Loop}
\label{app:iter_loop}
Plan, Dev, and Web Agents iterate until \texttt{TEST.md} and \texttt{AUDIT.md} both report zero P0 and zero P1 issues, or until a per-mock round budget is exhausted. Each round consists of: Web Agent test pass $\to$ Dev Agent fix pass $\to$ Web Agent re-test. The Plan Agent re-engages only when fundamental specification gaps are surfaced (e.g., a feature appears in real screenshots but is absent from \texttt{TODO.md}). Mocks failing to converge within the round budget are flagged for manual review and excluded from the released suite until resolved.

\subsection{Representative Mock Case Studies}
\label{app:mock_cases}

\subsubsection{Slack Mock: State Schema and Feature Inventory}
\label{app:slack_case}
The Slack mock illustrates the typical state-schema shape for a communication application: a tree of channels, each owning an ordered message list with metadata (sender, timestamp, thread parent, reactions, attachments, edit flag). The state object carries a top-level \texttt{currentUser} reference plus a flat \texttt{users} list, exposing user-switching tasks (e.g., ``send the message as user X'') without the authentication scaffolding. Verifiable interactions include channel creation and archival, message posting and editing, thread replies, reaction add/remove, and direct-message creation. Notable mock-vs-real divergence: the mock has no notion of presence, push notifications, or workspace-level admin policies; tasks requiring these are deferred to other categories.

\subsubsection{Jira Mock: Cross-Entity State Transitions}
\label{app:jira_case}
The Jira mock illustrates state-transition fidelity across linked entities. The schema spans projects, sprints, boards, issues, and the issue-status workflow (\texttt{To Do} $\to$ \texttt{In Progress} $\to$ \texttt{In Review} $\to$ \texttt{Done}). Key cross-entity invariants are enforced in the data model: an issue's sprint membership must be consistent with the sprint's project; status transitions must respect the configured workflow graph; board column membership is a derived view of issue status rather than an independent field. Verification of multi-step tasks (e.g., ``move all unassigned bugs from the current sprint into the backlog'') proceeds by inspecting the diff over the issues collection rather than chasing individual UI events.

\subsubsection{Salesforce Mock: Form-Heavy Workflow}
\label{app:salesforce_case}
The Salesforce mock illustrates the form-heavy CRM workflow pattern. The state schema centres on Leads, Contacts, Accounts, and Opportunities, with the canonical lead-to-opportunity conversion flow involving multi-step forms over $\sim$25 fields (qualification stage, stakeholder identification, opportunity sizing, close-date forecasting). Reward verification for these tasks is value-level rather than path-level: the diff is checked for the correct set of populated fields with the correct types and value constraints, but the agent is free to use any UI route to populate them. This decoupling is why even visually unfamiliar agents can solve form-heavy tasks once the fields are correctly identified.

% =====================================================
\section{Training Details}
\label{app:training}

\subsection{Agent Scaffold}
\label{app:scaffold}

\subsubsection{Observation and Action Space}
\label{app:obs_action}
The agent observes raw screenshots of the desktop or browser viewport and emits actions as XML-wrapped tool calls. Each rollout step pairs an image observation with a single \texttt{<tool\_call>} block; the policy may emit multiple primitive actions inside one block to compress deterministic action chains into a single turn (\S\ref{sec:multi_action}).

\paragraph{Screenshot preprocessing.}
Screenshots are captured at the native resolution of the underlying virtual machine ($1000 \times 1000$ for OSWorld desktop tasks; native viewport for web tasks). The vision encoder operates on a token budget of \texttt{min\_pixels} $= 64 \cdot 32^2 = 65{,}536$ and \texttt{max\_pixels} $= 2048 \cdot 32^2 = 2{,}097{,}152$ pixels, with the image processor selecting the closest aspect-ratio-preserving resize within these bounds. No additional cropping or saliency masking is applied.

\paragraph{Tool-call schema.}
A single \texttt{computer\_use} function is exposed to the policy. Each action primitive is invoked as a \texttt{<parameter=action>} child of the \texttt{<function>} block, with positional arguments (\texttt{coordinate}, \texttt{text}, \texttt{key}, \texttt{seconds}) supplied as additional parameters:

\begin{plainlisting}{Tool-call XML format}
<tool_call>
<function=computer_use>
<parameter=action>left_click</parameter>
<parameter=coordinate>[500, 250]</parameter>
</function>
</tool_call>
\end{plainlisting}

\paragraph{Action primitives.}
The action set spans pointer, keyboard, navigation, and control primitives:
\begin{itemize}
\item \emph{Pointer}: \texttt{left\_click}, \texttt{right\_click}, \texttt{middle\_click}, \texttt{double\_click}, \texttt{triple\_click}, \texttt{left\_click\_drag}, \texttt{mouse\_move}, \texttt{left\_mouse\_down}, \texttt{left\_mouse\_up}.
\item \emph{Keyboard}: \texttt{type}, \texttt{key}, \texttt{key\_down}, \texttt{key\_up}.
\item \emph{Navigation}: \texttt{scroll}, \texttt{hscroll}, \texttt{screenshot}, \texttt{wait}.
\item \emph{Control}: \texttt{terminate} (declares episode success or failure), \texttt{call\_user} (requests clarification, used only at evaluation, never during RL).
\end{itemize}

\paragraph{Multi-action format.}
The XML format admits multiple consecutive \texttt{<function>} blocks within a single \texttt{<tool\_call>}, executed in order without intervening screenshots. The policy is not constrained to use this; emergent batching during RL training (\S\ref{sec:multi_action}) is purely policy-driven.

\subsubsection{Trajectory Slicing Scheme}
\label{app:traj_slicing}
Long-horizon CUA rollouts routinely exceed the model's context budget. Under our settings, an episode of $\texttt{max\_turns} = 100$ with up to $\texttt{max\_response} = 2048$ tokens per turn plus per-turn screenshots can grow to roughly $200$K tokens, against a hard context cap of $144$K. Naively truncating on overflow either discards the supervision signal from later turns or silently produces shape-mismatched batches. Trajectory slicing addresses this by emitting \emph{multiple training samples per rollout} under a fixed context budget, each viewing the same trajectory from a different starting point in the conversation history.

\paragraph{Slice construction.}
At every \texttt{traj\_slice\_interval} $= 10$ turn-pairs, the manager emits a slice indexed by an integer \texttt{collapsed\_length}. The slice's \emph{prompt} portion comprises the system message followed by the first \texttt{collapsed\_length} turn-pairs, with the screenshots in those turn-pairs replaced by short placeholder text (\texttt{"<image collapsed>"}); the slice's \emph{response} portion is all subsequent turns up to the current point in the rollout, with full multimodal observations preserved. The first slice has \texttt{collapsed\_length}\,$= 0$ and contains the trajectory in its raw form. Subsequent slices view progressively older turns through their text placeholders, freeing context budget for fresh observations.

\paragraph{Loss masking.}
For each message in a slice, we set:
\begin{itemize}
\item \texttt{loss\_mask} $= \texttt{False}$ on every prompt-portion token (no gradient).
\item \texttt{loss\_mask} $= \texttt{False}$ on every \texttt{user} or \texttt{tool} message in the response portion (these are environment observations, not policy outputs).
\item \texttt{loss\_mask} $= \texttt{True}$ on every \texttt{assistant} message in the response portion, except where the cached log-probability is missing (indicating the message was rolled out under a stale policy version) or where the message contains more than \texttt{MAX\_TOOL\_CALLS\_PER\_TURN}\,$= 10$ tool calls (a malformed-output guard).
\end{itemize}

\paragraph{Reward attribution.}
Each slice receives the \emph{full episode reward} of the parent trajectory, replicated identically. We do not split or discount the reward across slices. The justification is twofold: the reward is a property of the final environment state, not of any individual turn; and group-normalized advantage estimation (\S\ref{sec:experimental-setup}) is applied per-prompt, so identical rewards across slices simply provide more views of the same advantage signal.

\paragraph{Context overflow.}
If a slice's accumulated tokens still exceed the context budget after collapsing, the slice is replaced by a \emph{dummy slice}: pad tokens with \texttt{loss\_mask} entirely \texttt{False}. The dummy contributes no gradient but preserves the batch shape required by the dataloader. This case is rare in practice ($<1\%$ of slices in our runs) because the slicing schedule keeps the active context well below cap.

\paragraph{Comparison with alternatives.}
Slicing strictly dominates two natural baselines for our setting. \emph{Truncation} (drop the trajectory at context overflow) discards the supervision signal from precisely the late-trajectory turns where success or failure is determined, biasing the policy toward early-turn behavior. \emph{Summarization} (replace prefix with an LM-generated summary) preserves more signal than truncation but injects an extra LM call per training sample, induces summary-quality variance into the gradient, and loses the deterministic mapping from screenshot pixels to grounded coordinates. Slicing achieves the same context relief as summarization with neither the extra inference cost nor the lossy compression: the textual placeholder is fully deterministic, and the gradient is computed on the full trajectory across the union of all slices.

\subsubsection{Inference-Time Configuration}
\label{app:inference_config}

\paragraph{Step and history budgets.}
Each rollout terminates at $\texttt{max\_turns} = 100$ model steps or upon a \texttt{terminate} action, whichever comes first. The agent retains the last $\texttt{history\_n} = 100$ turns in active context (effectively all of them under the step cap); older turns, when present, are subject to slicing as above.

\paragraph{Decoding.}
Training rollouts and validation rollouts share the same decoding stack (\texttt{sglang} server with multi-token prediction, \texttt{enable\_mtp} $= $ \texttt{True}, \texttt{speculative\_num\_steps} $= 3$, \texttt{speculative\_num\_draft\_tokens} $= 4$, attention backend FA3). Validation decoding uses temperature $0.6$, top-$p$ $0.95$, top-$k$ unconstrained, repetition penalty $1.0$. Training rollouts use the GSPO sampling temperature inherited from the rollout config (default $1.0$). Each turn produces at most $\texttt{max\_new\_tokens} = 2048$ output tokens.

\paragraph{Failure recovery.}
Three failure modes are handled by the orchestrator without aborting the rollout: (i) malformed tool-call XML triggers a re-prompt with a parser error message in the next \texttt{user} turn; (ii) VM-side action failure (e.g., target element not found) returns an error string that the model observes as a regular environment response; (iii) sampler-side exceptions (timeout, server disconnect) trigger up to $\texttt{retry\_times} = 5$ retries with exponential backoff before the rollout is marked as failed and discarded by the over-sampling filter.

\subsubsection{System Prompt and Tool Definitions}
\label{app:agent_prompt}
The agent's system prompt is constant across all rollouts and tasks. It establishes the tool-call grammar, the response shape, and the termination semantics, but contains no task-specific information; the task instruction is supplied as the first \texttt{user} message.

\begin{promptlisting}{Inference-time agent system prompt}
You are a multi-purpose intelligent assistant. Based on my requests,
you can use tools to help me complete various tasks.

# Tools

You have access to the following functions:

{
  "name": "computer_use",
  "description": "Use a mouse and keyboard to interact with a computer,
                  and take screenshots.",
  "parameters": { ... action / coordinate / text / key / seconds ... }
}

If you choose to call a function ONLY reply in the following format
with NO suffix:

<tool_call>
<function=example_function_name>
<parameter=example_parameter_1>value_1</parameter>
<parameter=example_parameter_2>value_2</parameter>
</function>
</tool_call>

# Response format

For normal UI interaction steps:
1) Action: a short imperative describing what to do in the UI.
2) A single <tool_call>...</tool_call> block.

Rules:
- For non-terminal UI steps, output exactly: Action then <tool_call>.
- Be brief: one sentence for Action.
- Do not output anything after a tool call.
- Use call_user when you need user information or confirmation.
- Use terminate when you want to explicitly end the task with a
  success or failure status.
\end{promptlisting}

The full \texttt{computer\_use} tool schema (omitted here for brevity) enumerates each primitive of \S\ref{app:obs_action} along with its accepted argument types and is included verbatim in the released code.

\subsection{SFT Warm-up}
\label{app:sft}

\subsubsection{Teacher Rollout Curation}
\label{app:teacher_curation}
The SFT initialization is curated by rolling out a strong teacher policy on the same task pool used for RL. We sample \rolloutspertask{} rollouts per task at temperature $1.0$ and retain only trajectories with terminal reward $r(s, \tau) = 1$. This success-only filter discards approximately \teacherfilterrate{}\% of teacher rollouts, yielding a final SFT corpus of \sftcorpussize{} trajectories spanning the full task distribution. No additional re-weighting or domain rebalancing is applied at the SFT stage.

\subsubsection{SFT Hyperparameters}
\label{app:sft_hp}
SFT runs the standard cross-entropy objective on assistant tokens, masking user, system, and tool messages. We use AdamW with peak learning rate $\sftlr$ and a cosine decay schedule to $7\!\times\!10^{-7}$ over the run, weight decay $0.01$, batch size \sftbatch{} prompts, and \sftepochs{} epoch over the curated corpus. Sequence length is $256$K tokens; trajectory slicing (\S\ref{app:traj_slicing}) is applied identically to keep the SFT and RL data shapes compatible. SFT runs in \texttt{bf16} mixed-precision with FP8 weights for the matmul path, on $512$ GPUs.

\subsection{GSPO Training}
\label{app:gspo}

\subsubsection{Hyperparameters}
\label{app:gspo_hp}
Table~\ref{tab:gspo_hp} reports the hyperparameter values used for all RL runs in this paper. Defaults follow the GSPO formulation in \S\ref{sec:experimental-setup} except where noted.

\begin{table}[h]
\centering
\footnotesize
\setlength{\tabcolsep}{8pt}
\renewcommand{\arraystretch}{1.05}
\begin{tabular}{@{}llp{6.5cm}@{}}
\toprule
\textbf{Symbol} & \textbf{Value} & \textbf{Description} \\
\midrule
$G$                  & $16$                & Group size: rollouts per prompt for GSPO advantage estimation. \\
$N_\text{oversample}$ & $20$               & Over-sampling target; the first $G=16$ valid rollouts per prompt are retained, invalid ones (timeouts, malformed outputs) discarded. \\
$\varepsilon$        & $0.2$               & Clipping range for the importance ratio (positive and negative clip equal). \\
$\beta$              & $0$                 & KL penalty against the reference policy. We disable the KL term and the reference model entirely in our reported runs; ablations with $\beta>0$ did not yield improvements at this data scale. \\
LR                   & $1\!\times\!10^{-6}$ & AdamW learning rate, constant schedule, no warmup. \\
($\beta_1, \beta_2$, $\epsilon_\text{Adam}$) & $(0.9, 0.999, 10^{-8})$ & AdamW optimizer parameters. \\
Weight decay         & $0.01$              & Applied to all non-bias non-norm parameters. \\
Batch size           & $128$ prompts       & Outer GSPO batch (collected via the over-sampled rollouts). \\
Mini-batch size      & $32$ prompts        & Inner gradient-accumulation step. \\
PPO epochs           & $1$                 & Single optimizer pass per outer batch. \\
Total steps          & $1000$              & Outer-batch updates over the full training run. \\
Rollout temperature  & $1.0$               & Sampling temperature during training rollouts. \\
Eval temperature     & $0.6$               & Sampling temperature during validation rollouts. \\
Top-$p$ / top-$k$    & $0.95$ / unconstrained & Nucleus sampling parameters (eval). \\
Max new tokens / turn & $2048$             & Per-turn response cap. \\
Max context length   & $144$K              & Hard context cap; triggers trajectory slicing (\S\ref{app:traj_slicing}). \\
Max prompt length    & $8$K                & Initial prompt budget per slice. \\
\bottomrule
\end{tabular}
\caption{GSPO training hyperparameters for the runs reported in \S\ref{sec:main-results} and \S\ref{sec:data_scaling}.}
\label{tab:gspo_hp}
\end{table}

\paragraph{Auxiliary losses.}
We additionally apply a Chinese-token penalty (coefficient $0.5$) that down-weights low-quality bilingual outputs, disable the format / language / endless penalties, and disable group filtering. The entropy bonus is kept at zero throughout.

\subsubsection{Reference Policy Update Schedule}
\label{app:ref_update}
With $\beta = 0$, no reference policy is loaded into memory or evaluated during training. This frees per-node GPU memory for larger rollout batches and allows the rollout instances (\S\ref{app:hardware}) to operate at the full $24$-instance scale without contention. We re-evaluated this choice at $\beta \in \{10^{-3}, 10^{-2}\}$ in early prototyping and observed neither training-stability gains nor task-success improvements at the data scales reported here.

\subsubsection{Advantage Normalization Details}
\label{app:adv_norm}
Advantages are computed group-wise per prompt: $\hat{A}_i = r_i - \mu_g$, where $\mu_g$ is the mean reward over the $G = 16$ rollouts for the prompt. We deliberately disable standard-deviation normalization (\texttt{divide\_by\_std}\,$=$\,\texttt{False}); the progressive $[0,1]$-valued reward (\S\ref{sec:adversarial}) is already on a fixed scale, and dividing by $\sigma_g$ amplifies noise on prompts where the group is near-uniform (all rollouts succeed or all fail). For zero-variance groups we keep the mean-centered advantage, which evaluates to zero and contributes no gradient.

\subsection{Infrastructure}
\label{app:infra}

\subsubsection{Hardware and Cluster Setup}
\label{app:hardware}
RL runs use NVIDIA~H200 SXM GPUs ($141$\,GB HBM3e each), $8$~GPUs per node, on a high-bandwidth NVLink/NVSwitch interconnect. We report two scale-matched configurations: a \emph{turbo} cluster of $192$~GPUs ($24$~nodes) for \smallmodel{}-scale RL, and a \emph{plus} cluster of $512$~GPUs ($64$~nodes) for \largemodel{}-scale RL. Each cluster is split evenly between training and rollout subclusters via the disaggregated trainer infrastructure of \texttt{verl}~\citep{verl}: plus uses $32$~training nodes paired with $32$~rollout nodes, and turbo uses an analogous balanced split scaled to its $24$-node footprint. The two subclusters communicate via a custom dispatch layer with dispatch ratio $2.35$ and a maximum policy-version skew of $4$ steps between rollout and training. Training parallelism shares $\text{TP}=2$, $\text{EP}=8$, $\text{CP}=4$ across the two configurations; pipeline depth is $\text{PP}=8$ for plus (deeper splits absorb the larger backbone) and $\text{PP}=2$ for turbo. Rollout serving uses $\text{TP}=8$ with data-parallel replication factor $4$ on the inference cluster, with the \texttt{fa3} attention backend. Speculative decoding (\texttt{enable\_mtp}, $3$ speculative steps, $4$ draft tokens) is active throughout. Rollout serving uses \texttt{sglang}~\citep{sglang}.

\subsubsection{VM Provisioning for Rollouts}
\label{app:vm_provisioning}
Each rollout connects to one of \vmpoolsize{} parallel OSWorld virtual-machine instances served behind an internal router. A rollout request submits the task instruction and an upper bound on agent steps; the router selects an idle instance, restores the snapshot specified in the task's \texttt{config.json}, and returns the per-step screenshot and reward stream to the trainer. After episode termination (\texttt{terminate} action or step-budget exhaustion), the VM is reset to a clean snapshot before being released back to the pool. With the rollout subcluster feeding a pool of size $\vmpoolsize$, the average VM-utilization per training step exceeds $\vmutil\%$ and rollout latency is not the training bottleneck.

\subsubsection{Wall-Clock Time and Cost per Run}
\label{app:wallclock}
A full $1000$-step RL run on the \smallmodel{} backbone (turbo, $192$~GPUs) completes in approximately $5$~days of wall-clock time, for an aggregate cost of $\sim\!23{,}040$~GPU-hours. The corresponding run on the \largemodel{} backbone (plus, $512$~GPUs) also completes in approximately $5$~days, for an aggregate cost of $\sim\!61{,}440$~GPU-hours. Cost is dominated by rollout, with the trainer comparatively idle while waiting for episode completions. Smaller-scale data ablations (\S\ref{sec:data_scaling}) at $1.4$K and $3$K finish proportionally faster but use the same per-step compute, since trajectory volume rather than step count differs across runs.

\subsection{Reproducibility Checklist}
\label{app:repro_checklist}
The NeurIPS Paper Checklist is reproduced in full following this appendix, with our \texttt{Yes}/\texttt{No}/\texttt{NA} responses and justifications cross-referenced back into the relevant sections (synthesis pipeline in \S\ref{app:env_pipeline}, hyperparameters in \S\ref{app:gspo_hp}, infrastructure in \S\ref{app:hardware}, and limitations in \S\ref{sec:limitations}).

% =====================================================
\section{Extended Experiments}
\label{app:extended_exp}

\iffalse  % per-domain / per-site tables: deferred (results pending).
\subsection{Per-Domain Results Tables}
\label{app:per_domain_tables}

\subsubsection{OSWorld-Verified Full Breakdown}
\label{app:osworld_table}
\stub{Per-domain table: base / SFT / RL / delta with task counts.}

\subsubsection{WebArena Per-Site Breakdown}
\label{app:webarena_table}
\stub{Per-site (gitlab, reddit, shopping, \ldots) success rates.}
\fi

\iffalse  % data-scaling per-domain heatmap and env-leave-one-out: out of scope for this submission.
\subsection{Data Scaling Extended}
\label{app:data_scaling_ext}

\subsubsection{Per-Domain Final Scores at 1.4K / 3K / 12K}
\label{app:scaling_per_domain}
\stub{Heatmap or table: domain $\times$ data scale at the final OSWorld score.}

\subsection{Environment Scaling Extended}
\label{app:env_scaling_ext}

\subsubsection{Per-Environment Marginal Contribution}
\label{app:per_env_marginal}
\stub{For each environment, leave-one-out drop in OSWorld score (estimated post-hoc).}
\fi

\subsection{Emergent Multi-Action Behavior}
\label{app:multi_action_ext}

We characterize the multi-action emergence reported in
\S\ref{sec:multi_action} by parsing the verbatim
\texttt{<tool\_call>} stream from $1{,}105$ rollouts of the
\largemodel{}-scale RL checkpoint at training step $30$ on the
OSWorld-Verified test split, totalling $22{,}361$ model steps. The
checkpoint averages $1.41$ tool calls per model step on this
distribution, with the per-step batching distribution as follows:
$69.6\%$ single-call, $26.7\%$ two-call, $2.4\%$ three-call,
$0.8\%$ four-call, and $0.5\%$ five-or-more-call steps. The bulk of
the trajectory-shortening effect therefore comes from two-call
batching; longer batches are deployed selectively for chains the
policy has internalized as locally deterministic.

\subsubsection{Top-K Batched Action Patterns}
\label{app:top_k_patterns}
Table~\ref{tab:top_batched_seqs} reports the most frequent
length-$\geq\!2$ tool-call sequences emitted within a single model
step. The two dominant patterns are deterministic input chains
(\texttt{type\,$\to$\,key}, $3{,}942$ steps) and continuous-scroll
preludes (\texttt{mouse\_move\,$\to$\,scroll}, $1{,}052$ steps);
together they account for $\sim\!75\%$ of all batched steps. The
next tier consists of action-and-observe pairs
(\texttt{wait\,$\to$\,screenshot}, \texttt{left\_click\,$\to$\,screenshot})
where the policy interleaves a deterministic environment poke with a
fresh observation in the same turn, saving a round trip. Three-call
form-filling patterns
(\texttt{type\,$\to$\,key\,$\to$\,screenshot},
\texttt{key\_down\,$\to$\,left\_click\,$\to$\,key\_up}) and
multi-step macros
(\texttt{type\,$\to$\,key\,$\to$\,wait\,$\to$\,screenshot},
\texttt{left\_click\,$\to$\,type\,$\to$\,key\,$\to$\,screenshot})
appear in the long tail.

\begin{table}[h]
\centering
\footnotesize
\setlength{\tabcolsep}{6pt}
\begin{tabular}{@{}rrl@{}}
\toprule
\textbf{$|$seq$|$} & \textbf{Count} & \textbf{Sequence} \\
\midrule
2 & 3{,}942 & \texttt{type $\to$ key} \\
2 & 1{,}052 & \texttt{mouse\_move $\to$ scroll} \\
2 &    315 & \texttt{wait $\to$ screenshot} \\
2 &    303 & \texttt{left\_click $\to$ screenshot} \\
3 &    158 & \texttt{type $\to$ key $\to$ screenshot} \\
3 &    119 & \texttt{key\_down $\to$ left\_click $\to$ key\_up} \\
2 &     99 & \texttt{mouse\_move $\to$ left\_click\_drag} \\
2 &     96 & \texttt{key $\to$ screenshot} \\
3 &     54 & \texttt{mouse\_move $\to$ scroll $\to$ screenshot} \\
4 &     51 & \texttt{type $\to$ key $\to$ wait $\to$ screenshot} \\
4 &     35 & \texttt{left\_click $\to$ type $\to$ key $\to$ screenshot} \\
3 &     33 & \texttt{left\_click $\to$ wait $\to$ screenshot} \\
\bottomrule
\end{tabular}
\caption{Top batched (length-$\geq\!2$) tool-call sequences per model step on the
\largemodel{}-scale RL checkpoint, aggregated over $1{,}105$ test rollouts.}
\label{tab:top_batched_seqs}
\end{table}

\subsubsection{Avoided Non-Deterministic Action Categories}
\label{app:avoided_actions}
A complementary view is the \emph{solo rate} of each action type ---
the fraction of its occurrences that appear alone in a model step
(Table~\ref{tab:solo_rate}). Pointer click variants
(\texttt{right\_click}, \texttt{double\_click},
\texttt{triple\_click}) are emitted alone $94$--$98\%$ of the time,
consistent with the hypothesis that these actions reveal context
menus or selection state whose post-action UI is hard to predict.
\texttt{left\_click} is also high-solo ($90.6\%$): individual clicks
typically trigger view changes the policy needs to re-observe before
deciding the next move. Conversely, \texttt{scroll}, \texttt{key\_down},
\texttt{key\_up}, and \texttt{left\_click\_drag} are \emph{never}
emitted alone --- they are mechanical sub-components of larger
gestures (e.g., \texttt{mouse\_move\,$\to$\,scroll}, the
chord-modified-click pattern, drag selection) that have no semantics
in isolation. The policy's batching behavior therefore aligns with a
crisp principle: batch sub-components of mechanical gestures and
deterministic chains; defer batching across actions whose outcomes
depend on stochastic UI state.

\begin{table}[h]
\centering
\footnotesize
\setlength{\tabcolsep}{8pt}
\begin{tabular}{@{}lrrr@{}}
\toprule
\textbf{Action} & \textbf{Solo} & \textbf{Total} & \textbf{Solo \%} \\
\midrule
right\_click       &   144 &   147 & 98.0 \\
double\_click      &   328 &   346 & 94.8 \\
triple\_click      &   630 &   666 & 94.6 \\
left\_click        & 10{,}132 & 11{,}187 & 90.6 \\
screenshot         & 1{,}023 & 2{,}319 & 44.1 \\
key                & 2{,}351 & 7{,}356 & 32.0 \\
wait               &   160 &   615 & 26.0 \\
type               &   738 & 5{,}199 & 14.2 \\
mouse\_move        &    55 & 1{,}279 &  4.3 \\
scroll             &     0 & 1{,}112 &  0.0 \\
key\_down          &     0 &   153 &  0.0 \\
key\_up            &     0 &   153 &  0.0 \\
left\_click\_drag  &     0 &   104 &  0.0 \\
\bottomrule
\end{tabular}
\caption{Solo rate of each tool action: counts of times the action
appears alone in a model step versus its total occurrences across
all batched and unbatched steps. Actions toward the bottom never
appear alone, indicating they are exclusively batched as
sub-components of larger gestures.}
\label{tab:solo_rate}
\end{table}

% =====================================================
\section{Qualitative Examples}
\label{app:qualitative}

\subsection{End-to-End Task Walk-throughs}
\label{app:walkthroughs}

We present four representative tuples drawn from the released dataset, spanning desktop / web / cross-application domains. Each walkthrough quotes the task instruction verbatim, summarizes the reward decomposition, and reproduces the most distinctive scoring component from \texttt{reward.py}. Full setup and reward scripts are released with the data.

\subsubsection{Desktop -- LibreOffice Calc: Padded ID Formula}
\label{app:walk_calc}
A compact, single-domain example. The task asks the agent to populate a column with a TEXT-based padding formula across multiple rows; the reward decomposes credit between the canonical first cell, the propagated formulas, and the cross-cell reference correctness.

\begin{promptlisting}{calc\_afm\_082 -- task instruction}
Build a formula to pad employee IDs with leading zeros so they are
always 5 digits using TEXT.
\end{promptlisting}

\noindent
Reward decomposition (3 components, total $1.0$):
\begin{itemize}
\item \textbf{C1 (0.40)}: \texttt{B2} contains a TEXT formula with the format string \texttt{"00000"}.
\item \textbf{C2 (0.30)}: \texttt{B3:B6} all contain TEXT formulas with the same format string.
\item \textbf{C3 (0.30)}: every formula references the corresponding A-column cell.
\end{itemize}

\begin{pythonlisting}{calc\_afm\_082/initial\_setup.py}
"""
Initial Setup: Pad employee IDs with leading zeros using TEXT formula
Task ID: calc_afm_082
Domain: libreoffice_calc
"""

import os
import shlex
import subprocess
import time
import openpyxl

WORKDIR = '/home/user'
TASK_ID = 'calc_afm_082'
OUTPUT = f'{WORKDIR}/{TASK_ID}.xlsx'


def launch_gui(command: str, delay_sec: float = 1.0):
    """Launch GUI app on VM display without blocking script exit."""
    env = os.environ.copy()
    env["DISPLAY"] = ":0"
    subprocess.Popen(
        shlex.split(command),
        stdout=subprocess.DEVNULL,
        stderr=subprocess.DEVNULL,
        env=env,
    )
    time.sleep(delay_sec)


def create_initial():
    wb = openpyxl.Workbook()

    # --- Sheet: IDs ---
    ws = wb.active
    ws.title = 'IDs'

    # Headers
    ws['A1'] = 'Raw ID'
    ws['B1'] = 'Formatted ID'

    # Data - raw employee IDs (integers)
    raw_ids = [42, 7, 1356, 890, 15]
    for r, val in enumerate(raw_ids, 2):
        ws.cell(row=r, column=1, value=val)

    # B2:B6 intentionally left empty - that's the task for the agent

    # Set reasonable column widths
    ws.column_dimensions['A'].width = 12
    ws.column_dimensions['B'].width = 16

    wb.save(OUTPUT)
    print(f'Initial file created: {OUTPUT}')

    # GUI-ready startup
    launch_gui(f'libreoffice --calc "{OUTPUT}"', delay_sec=2.0)
    print('GUI_READY: launched LibreOffice Calc with DISPLAY=:0')


create_initial()
\end{pythonlisting}

\begin{pythonlisting}{calc\_afm\_082/golden\_patch.py}
"""
Golden Patch: Pad employee IDs with leading zeros using TEXT formula
Task ID: calc_afm_082
Domain: libreoffice_calc
Changes: Add =TEXT(A#,"00000") formula in B2:B6
"""

import openpyxl

WORKDIR = '/home/user'
TASK_ID = 'calc_afm_082'
OUTPUT = f'{WORKDIR}/{TASK_ID}.xlsx'


def create_golden():
    wb = openpyxl.Workbook()

    # --- Sheet: IDs ---
    ws = wb.active
    ws.title = 'IDs'

    # Headers
    ws['A1'] = 'Raw ID'
    ws['B1'] = 'Formatted ID'

    # Data - same raw IDs as initial
    raw_ids = [42, 7, 1356, 890, 15]
    for r, val in enumerate(raw_ids, 2):
        ws.cell(row=r, column=1, value=val)

    # Golden state: TEXT formulas in B2:B6
    for r in range(2, 7):
        ws.cell(row=r, column=2, value=f'=TEXT(A{r},"00000")')

    # Same column widths as initial
    ws.column_dimensions['A'].width = 12
    ws.column_dimensions['B'].width = 16

    wb.save(OUTPUT)
    print(f'Golden-state file created: {OUTPUT}')


create_golden()
\end{pythonlisting}

\begin{pythonlisting}{calc\_afm\_082/reward.py}
"""
Reward Script: Pad employee IDs with leading zeros using TEXT formula
Task ID: calc_afm_082
Domain: libreoffice_calc
Scoring:
  Component 1 (0.4): B2 contains a TEXT formula with "00000" format
  Component 2 (0.3): B3:B6 all contain TEXT formulas with "00000" format
  Component 3 (0.3): All formulas reference the correct A-column cells
"""

import os
import re
import openpyxl

WORKDIR = '/home/user'
TASK_ID = 'calc_afm_082'


def persist_app_state(domain: str):
    """Save any unsaved GUI state before verification."""
    import time
    os.environ["DISPLAY"] = ":0"
    if domain in {"libreoffice_calc", "libreoffice_writer", "libreoffice_impress"}:
        try:
            import pyautogui
            pyautogui.hotkey("ctrl", "s")
            time.sleep(0.8)
            print(f"PERSIST: ctrl+s sent for {domain}")
        except Exception as e:
            print(f"PERSIST_WARN: save hook failed: {e}")


def verify_task(file_path):
    """
    Verify task completion with progressive scoring.
    Returns: float between 0.0 and 1.0
    """
    total_score = 0.0

    try:
        wb = openpyxl.load_workbook(file_path)
    except Exception as e:
        print(f"CRITICAL: Cannot load file {file_path}: {e}")
        print("REWARD: 0.0")
        return 0.0

    # Precondition: 'IDs' sheet must exist
    if 'IDs' not in wb.sheetnames:
        print("FAIL: Sheet 'IDs' not found")
        print("REWARD: 0.0")
        return 0.0

    ws = wb['IDs']

    # Component 1: B2 contains a TEXT formula with "00000" format (0.4 pts)
    try:
        b2_val = ws['B2'].value
        if b2_val and isinstance(b2_val, str):
            b2_upper = b2_val.upper().replace(" ", "")
            if b2_upper.startswith("=TEXT(") and "00000" in b2_val and "A2" in b2_upper:
                print(f"PASS: C1 - B2 has TEXT formula with 00000 format: {b2_val} (0.4 pts)")
                total_score += 0.4
            else:
                print(f"FAIL: C1 - B2 has formula but not correct TEXT/00000/A2 pattern: {b2_val}")
        else:
            print(f"FAIL: C1 - B2 does not contain a formula, found: {repr(b2_val)}")
    except Exception as e:
        print(f"ERROR: C1 - {e}")

    # Component 2: B3:B6 all contain TEXT formulas with "00000" format (0.3 pts)
    # Each cell earns 0.075 (= 0.3 / 4 cells)
    try:
        comp2_score = 0.0
        for row in range(3, 7):
            cell_ref = f'B{row}'
            val = ws.cell(row=row, column=2).value
            if val and isinstance(val, str):
                val_upper = val.upper().replace(" ", "")
                if val_upper.startswith("=TEXT(") and "00000" in val:
                    comp2_score += 0.075
                    print(f"  PASS: {cell_ref} has TEXT/00000 formula: {val}")
                else:
                    print(f"  FAIL: {cell_ref} has formula but not TEXT/00000: {val}")
            else:
                print(f"  FAIL: {cell_ref} does not contain a formula, found: {repr(val)}")
        if comp2_score > 0:
            print(f"PASS: C2 - TEXT formulas in B3:B6 ({comp2_score:.3f} pts)")
            total_score += comp2_score
        else:
            print(f"FAIL: C2 - No TEXT formulas found in B3:B6")
    except Exception as e:
        print(f"ERROR: C2 - {e}")

    # Component 3: Each formula references the correct A-column cell (0.3 pts)
    # Each correct reference earns 0.06 (= 0.3 / 5 cells)
    try:
        comp3_score = 0.0
        for row in range(2, 7):
            cell_ref = f'B{row}'
            expected_ref = f'A{row}'
            val = ws.cell(row=row, column=2).value
            if val and isinstance(val, str):
                val_upper = val.upper().replace(" ", "")
                if expected_ref in val_upper:
                    comp3_score += 0.06
                    print(f"  PASS: {cell_ref} references {expected_ref}")
                else:
                    print(f"  FAIL: {cell_ref} does not reference {expected_ref}: {val}")
            else:
                print(f"  FAIL: {cell_ref} not a formula, cannot check reference")
        if comp3_score > 0:
            print(f"PASS: C3 - Correct A-column references ({comp3_score:.3f} pts)")
            total_score += comp3_score
        else:
            print(f"FAIL: C3 - No correct A-column references found")
    except Exception as e:
        print(f"ERROR: C3 - {e}")

    final_score = round(min(total_score, 1.0), 2)
    print(f"\nScore: {total_score:.3f}/1.0")
    print(f"REWARD: {final_score}")
    return final_score


# Entry point
persist_app_state("libreoffice_calc")

file_path = f'{WORKDIR}/{TASK_ID}.xlsx'
if not os.path.exists(file_path):
    print(f"File not found: {file_path}")
    print("REWARD: 0.0")
else:
    verify_task(file_path)
\end{pythonlisting}

\subsubsection{Desktop -- VS Code: Custom File Associations}
\label{app:walk_vscode}
A code-editor configuration task. The reward parses VS Code's \texttt{settings.json} (with JSONC comment stripping) and verifies the presence and contents of a single configuration key. The verification is entirely file-based; no GUI introspection or screenshot parsing is required. The setup scripts seed a realistic workspace of mixed-extension files so that the agent can verify the highlighting change visually before terminating.

\begin{promptlisting}{vscode\_lp\_081 -- task instruction}
Configure a custom file association so that all *.config files are
treated as JSON and *.tmpl files are treated as HTML for syntax
highlighting.
\end{promptlisting}

\noindent
Reward decomposition (3 components, total $1.0$):
\begin{itemize}
\item \textbf{C1 (0.35)}: the \texttt{files.associations} key exists in \texttt{settings.json} and is a non-empty object.
\item \textbf{C2 (0.35)}: \texttt{*.config} is mapped to \texttt{json}.
\item \textbf{C3 (0.30)}: \texttt{*.tmpl} is mapped to \texttt{html}.
\end{itemize}

\begin{pythonlisting}{vscode\_lp\_081/initial\_setup.py}
"""
Initial Setup: Configure custom file associations for *.config and *.tmpl files
Task ID: vscode_lp_081
Domain: vscode
"""

import json
import os
import shlex
import subprocess
import time

WORKDIR = '/home/user'
TASK_ID = 'vscode_lp_081'
WORKSPACE = f'{WORKDIR}/workspace'
VSCODE_USER = os.path.join(WORKDIR, '.config', 'Code', 'User')
SETTINGS_PATH = os.path.join(VSCODE_USER, 'settings.json')


def launch_gui(command: str, delay_sec: float = 1.0):
    env = os.environ.copy()
    env["DISPLAY"] = ":0"
    subprocess.Popen(
        shlex.split(command),
        stdout=subprocess.DEVNULL,
        stderr=subprocess.DEVNULL,
        env=env,
    )
    time.sleep(delay_sec)


def create_initial():
    os.makedirs(WORKSPACE, exist_ok=True)

    # --- Seed: two .config files (JSON-like content, no highlighting yet) ---
    db_config = """{
  "host": "db.internal.acme.com",
  "port": 5432,
  "database": "inventory_prod",
  "username": "app_service",
  "max_connections": 25,
  "ssl_enabled": true,
  "timeout_ms": 3000,
  "retry_policy": {"max_retries": 3, "backoff_ms": 500}
}"""
    with open(os.path.join(WORKSPACE, 'database.config'), 'w') as f:
        f.write(db_config)

    app_config = """{
  "app_name": "InventoryTracker",
  "version": "2.4.1",
  "environment": "production",
  "logging": {"level": "info", "rotate_size_mb": 50},
  "features": {"enable_caching": true, "cache_ttl_seconds": 300},
  "api_base_url": "https://api.acme.com/v3"
}"""
    with open(os.path.join(WORKSPACE, 'app.config'), 'w') as f:
        f.write(app_config)

    # --- Seed: two .tmpl files (HTML-like content) ---
    email_tmpl = """<!DOCTYPE html>
<html lang="en">
<head><meta charset="UTF-8"><title>Order {{order_id}}</title></head>
<body>
  <div class="header">
    <h1>Order Confirmation</h1>
    <p>Thank you, {{customer_name}}!</p>
  </div>
  <div class="order-details">
    <h2>Order #{{order_id}}</h2>
    <table>
      <tr><th>Item</th><th>Qty</th><th>Price</th></tr>
      {{#items}}
      <tr><td>{{name}}</td><td>{{quantity}}</td><td>${{price}}</td></tr>
      {{/items}}
    </table>
    <p class="total">Total: ${{total_amount}}</p>
  </div>
</body>
</html>"""
    with open(os.path.join(WORKSPACE, 'email_notification.tmpl'), 'w') as f:
        f.write(email_tmpl)

    dashboard_tmpl = """<!DOCTYPE html>
<html lang="en">
<head><title>{{dashboard_title}}</title></head>
<body>
  <nav class="sidebar">
    <ul>
      {{#menu_items}}
      <li><a href="{{url}}">{{label}}</a></li>
      {{/menu_items}}
    </ul>
  </nav>
  <main class="content">
    <h1>{{page_title}}</h1>
    <div class="metrics-grid">
      {{#metrics}}
      <div class="metric-card">
        <span class="metric-value">{{value}}</span>
        <span class="metric-label">{{label}}</span>
      </div>
      {{/metrics}}
    </div>
  </main>
</body>
</html>"""
    with open(os.path.join(WORKSPACE, 'dashboard.tmpl'), 'w') as f:
        f.write(dashboard_tmpl)

    # --- Seed: one .py file for variety ---
    main_py = """#!/usr/bin/env python3
\"\"\"Inventory Tracker -- Main Application Entry Point.\"\"\"
import logging, json, os
CONFIG_PATH = os.path.join(os.path.dirname(__file__), 'app.config')

def load_config(path):
    with open(path) as f: return json.load(f)

def main():
    cfg = load_config(CONFIG_PATH)
    logging.basicConfig(level=getattr(logging, cfg['logging']['level'].upper()))
    logger = logging.getLogger(cfg['app_name'])
    logger.info(f"Starting {cfg['app_name']} v{cfg['version']}")

if __name__ == '__main__':
    main()
"""
    with open(os.path.join(WORKSPACE, 'main.py'), 'w') as f:
        f.write(main_py)

    # --- Ensure VSCode settings exist but WITHOUT files.associations ---
    os.makedirs(VSCODE_USER, exist_ok=True)
    settings = {}
    if os.path.exists(SETTINGS_PATH):
        try:
            with open(SETTINGS_PATH, 'r') as f:
                settings = json.load(f)
        except (json.JSONDecodeError, FileNotFoundError):
            settings = {}
    settings.pop('files.associations', None)   # remove if present
    with open(SETTINGS_PATH, 'w') as f:
        json.dump(settings, f, indent=4)

    print(f'Workspace created at: {WORKSPACE}')
    launch_gui(f'code "{WORKSPACE}"', delay_sec=2.0)
    print('GUI_READY: launched VSCode with DISPLAY=:0')


create_initial()
\end{pythonlisting}

\begin{pythonlisting}{vscode\_lp\_081/golden\_patch.py}
"""
Golden Patch: Configure custom file associations for *.config and *.tmpl files
Task ID: vscode_lp_081
Domain: vscode
Changes:
  - Recreate the workspace with the same seed files (golden env is independent).
  - Add files.associations to settings.json:
      *.config -> json,  *.tmpl -> html
"""

import json
import os

WORKDIR = '/home/user'
TASK_ID = 'vscode_lp_081'
WORKSPACE = f'{WORKDIR}/workspace'
VSCODE_USER = os.path.join(WORKDIR, '.config', 'Code', 'User')
SETTINGS_PATH = os.path.join(VSCODE_USER, 'settings.json')


def create_golden():
    # The workspace seeding is identical to initial_setup.py (omitted here
    # for space; the released script reproduces the same five files verbatim
    # because the golden VM is built independently of the initial VM).
    os.makedirs(WORKSPACE, exist_ok=True)
    # ... write database.config, app.config, email_notification.tmpl,
    #     dashboard.tmpl, main.py (same content as initial_setup) ...

    # --- Apply the golden patch: add files.associations ---
    os.makedirs(VSCODE_USER, exist_ok=True)
    settings = {}
    if os.path.exists(SETTINGS_PATH):
        try:
            with open(SETTINGS_PATH, 'r') as f:
                settings = json.load(f)
        except (json.JSONDecodeError, FileNotFoundError):
            settings = {}

    settings['files.associations'] = {
        '*.config': 'json',
        '*.tmpl':   'html'
    }

    with open(SETTINGS_PATH, 'w') as f:
        json.dump(settings, f, indent=4)

    print(f'Golden workspace created at: {WORKSPACE}')
    print(f'Settings updated with files.associations: {SETTINGS_PATH}')


create_golden()
\end{pythonlisting}

\begin{pythonlisting}{vscode\_lp\_081/reward.py}
"""
Reward Script: Configure custom file associations for *.config (JSON) and *.tmpl (HTML)
Task ID: vscode_lp_081
Domain: vscode
Scoring:
  - Component 1 (0.35): files.associations key exists in settings.json
  - Component 2 (0.35): *.config mapped to json
  - Component 3 (0.30): *.tmpl mapped to html
"""

import os
import json
import re

HOME = '/home/user'
SETTINGS_PATH = os.path.join(HOME, '.config', 'Code', 'User', 'settings.json')
TASK_ID = 'vscode_lp_081'


def load_settings():
    """Load VSCode settings.json, handling JSONC (comments)."""
    try:
        with open(SETTINGS_PATH, 'r') as f:
            content = f.read()
        content = re.sub(r'//.*$', '', content, flags=re.MULTILINE)
        content = re.sub(r'/\*.*?\*/', '', content, flags=re.DOTALL)
        return json.loads(content)
    except (FileNotFoundError, json.JSONDecodeError) as e:
        print(f"ERROR: Cannot load settings.json: {e}")
        return None


def verify_task():
    total_score = 0.0

    settings = load_settings()
    if settings is None:
        print("CRITICAL: settings.json not found or invalid")
        print("REWARD: 0.0")
        return 0.0

    # C1: files.associations key exists and is non-empty (0.35 pts)
    try:
        assoc = settings.get("files.associations")
        if isinstance(assoc, dict) and len(assoc) > 0:
            print(f"PASS: C1 -- files.associations exists with {len(assoc)} entries (0.35 pts)")
            total_score += 0.35
        else:
            print(f"FAIL: C1 -- files.associations missing or empty, found: {assoc}")
    except Exception as e:
        print(f"ERROR: C1 -- {e}")

    # C2: *.config -> json (0.35 pts)
    try:
        assoc = settings.get("files.associations", {})
        config_val = assoc.get("*.config")
        if config_val is not None and str(config_val).lower() == "json":
            print(f"PASS: C2 -- *.config -> {config_val} (0.35 pts)")
            total_score += 0.35
        else:
            print(f"FAIL: C2 -- *.config expected 'json', found: {config_val}")
    except Exception as e:
        print(f"ERROR: C2 -- {e}")

    # C3: *.tmpl -> html (0.30 pts)
    try:
        assoc = settings.get("files.associations", {})
        tmpl_val = assoc.get("*.tmpl")
        if tmpl_val is not None and str(tmpl_val).lower() == "html":
            print(f"PASS: C3 -- *.tmpl -> {tmpl_val} (0.30 pts)")
            total_score += 0.30
        else:
            print(f"FAIL: C3 -- *.tmpl expected 'html', found: {tmpl_val}")
    except Exception as e:
        print(f"ERROR: C3 -- {e}")

    final_score = round(min(total_score, 1.0), 2)
    print(f"\nScore: {final_score}/1.0")
    print(f"REWARD: {final_score}")
    return final_score


verify_task()
\end{pythonlisting}

\subsubsection{Cross-App -- Slack + Impress + PDF: Prioritization Deck}
\label{app:walk_slack}
A cross-application task that combines reading from a web mock (Slack) with desktop authoring (Impress) and a final PDF export. \texttt{initial\_setup.py} injects a populated Slack workspace state into the mock via the unified state API (\S\ref{app:state_api}), then opens both the Slack URL in Chrome and a fresh LibreOffice Impress window. \texttt{golden\_patch.py} reuses the same Slack state on the golden VM, then builds a 5-slide PPTX deck with helper functions and exports it to PDF. The reward inspects the produced PPTX structurally (slide count, text frames, table contents) and verifies the PDF artifact's existence and page count.

\begin{promptlisting}{slack\_impress\_pdf\_36 -- task instruction (excerpt)}
Open Slack and check the #product-feedback channel for the feature
request voting results that Frank Zhang posted after running the
quarterly feature prioritization poll. Customers and internal
stakeholders voted on 10 candidate features. Open LibreOffice Impress
and create a feature prioritization deck with: (1) title slide,
(2) voting methodology, (3) ranked-results table with all 10 features,
(4) top-5 deep-dive with effort estimates, (5) Q2 roadmap commitment.
Export the deck as PDF.
\end{promptlisting}

\noindent
Reward decomposition (7 components, total $1.0$):
\begin{itemize}
\item \textbf{C1 (0.10)}: PPTX has exactly 5 slides.
\item \textbf{C2 (0.10)}: title slide contains "Feature Prioritization" and "Q2 2026".
\item \textbf{C3 (0.10)}: methodology slide mentions 28 customers and 12 internal stakeholders.
\item \textbf{C4 (0.25)}: ranked-results table contains all 10 features with their vote counts.
\item \textbf{C5 (0.15)}: top-5 deep-dive slide includes effort estimates.
\item \textbf{C6 (0.15)}: Q2 roadmap slide lists exactly 3 committed items from the top-5.
\item \textbf{C7 (0.15)}: PDF artifact exists with 5 pages.
\end{itemize}

\begin{pythonlisting}{slack\_impress\_pdf\_36/initial\_setup.py (Slack state injection elided)}
"""
Initial Setup: Set up Slack #product-feedback channel with feature
prioritization poll results, and prepare the desktop for creating a
presentation deck.
Task ID: slack_impress_pdf_36
Domain: cross_app (slack_mock + libreoffice_impress + pdf)
"""
import json, os, shlex, subprocess, time, uuid
import requests

WORKDIR   = '/home/user'
TASK_ID   = 'slack_impress_pdf_36'
SLACK_URL = <self-hosted url>

# --- Generate session id and persist for the agent / reward ---
sid = str(uuid.uuid4())
with open('/tmp/task_web_sid', 'w') as f:
    f.write(sid)

# --- Build Slack workspace state (truncated for space) ---
# The full state object encodes:
#   currentUser, workspace, users[10], channels[5+#product-feedback],
#   messages.product-feedback (with Frank Zhang's voting-results post),
#   threads, dms, settings, notifications.
# In particular, messages.product-feedback contains a long-form post with
# the 10 candidate features and their (customer, internal, total) vote
# counts -- this is the data the agent must extract and turn into a deck.
slack_state = {
    "currentUser": {"userId": "user_1", "fullName": "John Smith",
                    "displayName": "John", "title": "Product Owner",
                    "timeZone": "America/New_York"},
    "workspace":   {"workspaceId": "ws_1", "workspaceName": "Acme Corp"},
    "users":       [ # ... 10 user records ... ],
    "channels":    [ # ... 5 channels including 'product-feedback' ... ],
    "messages": {
        "product-feedback": [
            # ... earlier discussion messages ...
            {"messageId": "msg_fp_001", "senderId": "user_7",
             "content": "Q1 2026 Feature Prioritization Poll Results\n"
                        "(28 customer votes, 12 internal votes)\n"
                        "1. Bulk data export ......... 22 cust + 9 int = 31\n"
                        "2. AI-powered report summaries .. 18 + 11 = 29\n"
                        "3. Mobile app offline mode .. 20 + 6 = 26\n"
                        "4. Slack integration enhancements .. 14 + 10 = 24\n"
                        "5. Custom dashboard layouts .. 17 + 7 = 24\n"
                        "6. API webhook support .. 9 + 11 = 20\n"
                        "7. Dark mode .. 16 + 4 = 20\n"
                        "8. SSO for sub-accounts .. 12 + 7 = 19\n"
                        "9. Audit log export .. 8 + 9 = 17\n"
                        "10. Custom email templates .. 13 + 3 = 16",
             "timestamp": "2026-04-10T08:30:00Z", "threadId": None,
             "reactions": [], "attachments": [], "isEdited": False},
            # ... follow-up reactions / replies ...
        ]
    },
    "threads": {}, "dms": [], "settings": { },
    "notifications": [ {"notificationId": "notif_1",
                        "messageId": "msg_fp_001",
                        "channelId": "product-feedback", "read": False} ]
}

# --- Inject state into the Slack mock (action="set" => initial snapshot) ---
resp = requests.post(f'{SLACK_URL}/post?sid={sid}',
                     json={'action': 'set', 'state': slack_state}, timeout=30)
assert resp.status_code == 200, f'Slack state injection failed: {resp.text}'
print(f'Slack state injected: sid={sid}')

# Verify post-injection state is readable.
go = requests.get(f'{SLACK_URL}/go?sid={sid}', timeout=10).json()
assert go['initial_state'] is not None
print('Verified: Slack initial_state and current_state are set')

# --- Prepare a clean Desktop and remove any pre-existing target files ---
DESKTOP = '/home/user/Desktop'
os.makedirs(DESKTOP, exist_ok=True)
for f in ['feature_priority_q2.pptx', 'feature_priority_q2.pdf']:
    path = os.path.join(DESKTOP, f)
    if os.path.exists(path):
        os.remove(path); print(f'Removed pre-existing {path}')

try:
    if not os.path.exists('/root/Desktop'):
        os.symlink(DESKTOP, '/root/Desktop')
except PermissionError:
    pass

# --- Launch GUIs: Chrome on Slack URL + a blank Impress window ---
def launch_gui(command, delay_sec=1.0):
    env = os.environ.copy(); env['DISPLAY'] = ':0'
    subprocess.Popen(shlex.split(command), stdout=subprocess.DEVNULL,
                     stderr=subprocess.DEVNULL, env=env)
    time.sleep(delay_sec)

launch_gui(f'google-chrome "{SLACK_URL}/?sid={sid}"', delay_sec=3.0)
launch_gui('libreoffice --impress', delay_sec=2.0)
print(f'GUI_READY: Chrome on Slack ({SLACK_URL}/?sid={sid}) + Impress')
\end{pythonlisting}

\begin{pythonlisting}{slack\_impress\_pdf\_36/golden\_patch.py (state-reuse and slide layout elided)}
"""
Golden Patch: Create feature prioritization deck and export as PDF.
Task ID: slack_impress_pdf_36
Domain: cross_app (slack_mock + libreoffice_impress + pdf)
Changes:
  - Slack: keep golden VM's session aligned with the same state used by the
    initial VM, so cross-VM reward inspections agree.
  - Impress: write /home/user/Desktop/feature_priority_q2.pptx with 5 slides.
  - PDF: export /home/user/Desktop/feature_priority_q2.pdf with 5 pages.
"""
import copy, json, os, uuid
import requests
from pptx import Presentation
from pptx.util import Inches, Pt
from pptx.dml.color import RGBColor

DESKTOP     = '/home/user/Desktop'
PPTX_OUTPUT = f'{DESKTOP}/feature_priority_q2.pptx'
PDF_OUTPUT  = f'{DESKTOP}/feature_priority_q2.pdf'
SLACK_URL   = <self-hosted url>

# --- Reuse / inject Slack state under the same sid as initial ---
SID_PATH = '/tmp/task_web_sid'
try:
    sid = open(SID_PATH).read().strip()
except FileNotFoundError:
    sid = str(uuid.uuid4())
    open(SID_PATH, 'w').write(sid)

go = requests.get(f'{SLACK_URL}/go?sid={sid}', timeout=10).json()
if go.get('initial_state') is None:
    # Inject the same workspace state used by initial_setup.py
    # (workspace, users, channels, messages, settings) -- omitted here for
    # space; the released script reproduces the literal verbatim.
    slack_state = { ... }   # see initial_setup.py
    requests.post(f'{SLACK_URL}/post?sid={sid}',
                  json={'action': 'set_current', 'state': slack_state},
                  timeout=30)

# --- Slide-building palette ---
DARK_BLUE  = RGBColor(0x1B, 0x3A, 0x5C)
MEDIUM_BLUE= RGBColor(0x2E, 0x75, 0xB6)
LIGHT_BLUE = RGBColor(0xD6, 0xE4, 0xF0)
LIGHT_GRAY = RGBColor(0xF2, 0xF2, 0xF2)
GREEN      = RGBColor(0x27, 0xAE, 0x60)
ORANGE     = RGBColor(0xE6, 0x7E, 0x22)
WHITE = RGBColor(0xFF, 0xFF, 0xFF); BLACK = RGBColor(0x00, 0x00, 0x00)

# Slide-builder helpers (add_textbox, add_table, add_metric_card) and the
# concrete slide-1..slide-5 layout code construct ~400 lines of slide
# geometry; the released script is the source of truth. Conceptually:
#   slide 1: title block + subtitle "Q2 2026" + author footer
#   slide 2: methodology bullets (28 customers, 12 internal stakeholders,
#            survey window, weighting rule)
#   slide 3: 11x4 table (1 header + 10 features), columns =
#            (rank, feature, customer + internal, total)
#   slide 4: top-5 deep-dive cards (effort sprints, ROI, dependencies)
#   slide 5: Q2 commitment slide (3 committed items + 7 deferred to Q3)
prs = Presentation()
# ... build slides 1-5 here ...

prs.save(PPTX_OUTPUT)
print(f'Golden PPTX saved: {PPTX_OUTPUT}')

# --- Export PPTX to PDF using LibreOffice headless ---
import subprocess
subprocess.run(['libreoffice', '--headless', '--convert-to', 'pdf',
                '--outdir', DESKTOP, PPTX_OUTPUT], check=True, timeout=120)
print(f'Golden PDF saved: {PDF_OUTPUT}')
\end{pythonlisting}

\begin{pythonlisting}{slack\_impress\_pdf\_36/reward.py}
"""
Reward Script: Feature Prioritization Deck from Slack Data
Task ID: slack_impress_pdf_36
Domain: libreoffice_impress + pdf (cross_app)
Scoring:
  C1 (0.10) PPTX has exactly 5 slides
  C2 (0.10) Title slide contains 'Feature Prioritization' and 'Q2 2026'
  C3 (0.10) Methodology slide mentions 28 customers + 12 internal
  C4 (0.25) Ranked-results table with all 10 features and vote counts
  C5 (0.15) Top-5 deep-dive slide with effort estimates
  C6 (0.15) Q2 roadmap commitment slide with 3 committed items
  C7 (0.15) PDF exists with 5 pages
"""
import os

WORKDIR   = '/home/user/Desktop'
PPTX_PATH = os.path.join(WORKDIR, 'feature_priority_q2.pptx')
PDF_PATH  = os.path.join(WORKDIR, 'feature_priority_q2.pdf')

# --- Expected data extracted from the Slack message ---
EXPECTED_FEATURES = [
    ("Bulk data export",                22,  9, 31),
    ("AI-powered report summaries",     18, 11, 29),
    ("Mobile app offline mode",         20,  6, 26),
    ("Slack integration enhancements",  14, 10, 24),
    ("Custom dashboard layouts",        17,  7, 24),
    ("API webhook support",              9, 11, 20),
    ("Dark mode",                       16,  4, 20),
    ("SSO for sub-accounts",            12,  7, 19),
    ("Audit log export",                 8,  9, 17),
    ("Custom email templates",          13,  3, 16),
]
EFFORT_ESTIMATES = {"bulk": 2, "ai": 5, "offline": 6, "slack": 1, "dashboard": 3}
Q2_COMMITTED     = ["bulk", "slack", "dashboard"]


def get_all_text(slide):
    """Get text from a slide, including text frames and table cells."""
    from pptx.enum.shapes import MSO_SHAPE_TYPE
    texts = []
    for shape in slide.shapes:
        if shape.has_text_frame:
            texts.append(shape.text_frame.text)
        if shape.shape_type == MSO_SHAPE_TYPE.TABLE:
            tbl = shape.table
            for r in range(len(tbl.rows)):
                for c in range(len(tbl.columns)):
                    texts.append(tbl.cell(r, c).text)
    return texts


def get_table_data(slide):
    """Extract a table's rows from a slide, if present."""
    from pptx.enum.shapes import MSO_SHAPE_TYPE
    for shape in slide.shapes:
        if shape.shape_type == MSO_SHAPE_TYPE.TABLE:
            tbl = shape.table
            return [[tbl.cell(r, c).text.strip()
                     for c in range(len(tbl.columns))]
                    for r in range(len(tbl.rows))]
    return None


def verify_task():
    total_score = 0.0
    if not os.path.exists(PPTX_PATH):
        print(f"CRITICAL: PPTX not found at {PPTX_PATH}\nREWARD: 0.0")
        return 0.0

    try:
        from pptx import Presentation
        prs = Presentation(PPTX_PATH)
    except Exception as e:
        print(f"CRITICAL: cannot load PPTX: {e}\nREWARD: 0.0"); return 0.0

    # ----------------- C1: 5 slides (0.10) -----------------
    try:
        n = len(prs.slides)
        if n == 5:
            print("PASS: C1 -- PPTX has 5 slides (0.10 pts)")
            total_score += 0.10
        else:
            print(f"FAIL: C1 -- expected 5 slides, found {n}")
    except Exception as e:
        print(f"ERROR: C1 -- {e}")

    # ----------------- C2: Title text (0.10) -----------------
    try:
        combined = " ".join(get_all_text(prs.slides[0])).lower()
        has_feature = "feature prioritization" in combined
        has_q2_2026 = ("q2 2026" in combined or
                       ("q2" in combined and "2026" in combined))
        if has_feature and has_q2_2026:
            print("PASS: C2 -- title slide has 'Feature Prioritization' + 'Q2 2026' (0.10 pts)")
            total_score += 0.10
        else:
            print(f"FAIL: C2 -- title text: {combined[:200]}")
    except Exception as e:
        print(f"ERROR: C2 -- {e}")

    # ----------------- C3: Methodology mentions 28 + 12 (0.10) -----------------
    try:
        combined = " ".join(get_all_text(prs.slides[1])).lower()
        has_28      = "28" in combined
        has_12      = "12" in combined
        has_keyword = any(k in combined for k in
                         ("methodology","voting","method","participants"))
        if has_28 and has_12 and has_keyword:
            print("PASS: C3 -- methodology mentions 28 + 12 (0.10 pts)")
            total_score += 0.10
        else:
            print(f"FAIL: C3 -- has_28={has_28}, has_12={has_12}, "
                  f"has_keyword={has_keyword}")
    except Exception as e:
        print(f"ERROR: C3 -- {e}")

    # ----------------- C4: Ranked results table (0.25) -----------------
    # Up to 0.125 for feature names present, 0.125 for vote totals matching.
    try:
        table_data = get_table_data(prs.slides[2])
        if table_data is None:
            for slide in prs.slides:
                td = get_table_data(slide)
                if td is not None and len(td) >= 10:
                    table_data = td; break

        if table_data is not None and len(table_data) >= 11:  # header + 10
            features_found = 0; votes_correct = 0
            for fname, _, _, total in EXPECTED_FEATURES:
                fl = fname.lower()
                for row in table_data[1:]:
                    row_text = " ".join(row).lower()
                    if fl in row_text:
                        features_found += 1
                        if str(total) in " ".join(row):
                            votes_correct += 1
                        break
            c4 = 0.125 * (features_found / 10) + 0.125 * (votes_correct / 10)
            if c4 > 0:
                print(f"PASS: C4 -- {features_found}/10 features, "
                      f"{votes_correct}/10 votes ({c4:.3f} pts)")
                total_score += c4
            else:
                print("FAIL: C4 -- table found but no features matched")
        else:
            # Fallback: search slide-3 free text for feature names.
            combined = " ".join(get_all_text(prs.slides[2])).lower()
            features_found = sum(1 for f, _, _, _ in EXPECTED_FEATURES
                                 if f.lower() in combined)
            if features_found >= 8:
                c4 = 0.15 * (features_found / 10)
                print(f"PARTIAL: C4 -- no table; {features_found}/10 features "
                      f"in text ({c4:.3f} pts)")
                total_score += c4
            else:
                print(f"FAIL: C4 -- no table and only {features_found}/10 "
                      f"features in text")
    except Exception as e:
        print(f"ERROR: C4 -- {e}")

    # ----------------- C5: Top-5 deep-dive with effort (0.15) -----------------
    try:
        combined = " ".join(get_all_text(prs.slides[3])).lower()
        top5 = [f.lower() for f, _, _, _ in EXPECTED_FEATURES[:5]]
        top5_found = sum(1 for f in top5 if f in combined)

        effort_found = 0
        if "2 sprint" in combined and "bulk" in combined:        effort_found += 1
        if "5 sprint" in combined and ("ai" in combined or "report" in combined):
            effort_found += 1
        if "6 sprint" in combined and ("offline" in combined or "mobile" in combined):
            effort_found += 1
        if "1 sprint" in combined and "slack" in combined:        effort_found += 1
        if "3 sprint" in combined and "dashboard" in combined:    effort_found += 1

        feat_score   = 0.075 * (top5_found / 5)
        effort_score = 0.075 * (effort_found / 5)
        c5 = feat_score + effort_score
        if c5 > 0:
            print(f"PASS: C5 -- {top5_found}/5 features, {effort_found}/5 "
                  f"efforts ({c5:.3f} pts)")
            total_score += c5
        else:
            print("FAIL: C5 -- missing top-5 features or effort estimates")
    except Exception as e:
        print(f"ERROR: C5 -- {e}")

    # ----------------- C6: Q2 roadmap commitment (0.15) -----------------
    try:
        combined = " ".join(get_all_text(prs.slides[4])).lower()
        has_roadmap = any(k in combined for k in ("roadmap","commitment","q2"))
        committed = 0
        if "bulk"     in combined and ("export" in combined or "data" in combined):
            committed += 1
        if "slack"    in combined and ("integration" in combined or "enhancement" in combined):
            committed += 1
        if "dashboard"in combined and ("custom" in combined or "layout" in combined):
            committed += 1
        has_deferred = ("defer" in combined or "q3" in combined)

        c6 = 0.0
        if has_roadmap:  c6 += 0.03
        c6 += 0.10 * (committed / 3)
        if has_deferred: c6 += 0.02
        c6 = min(c6, 0.15)
        if c6 > 0:
            print(f"PASS: C6 -- roadmap={has_roadmap}, {committed}/3 committed, "
                  f"deferred={has_deferred} ({c6:.3f} pts)")
            total_score += c6
        else:
            print("FAIL: C6 -- missing roadmap content")
    except Exception as e:
        print(f"ERROR: C6 -- {e}")

    # ----------------- C7: PDF with 5 pages (0.15) -----------------
    try:
        if os.path.exists(PDF_PATH):
            import pymupdf
            doc = pymupdf.open(PDF_PATH); pages = doc.page_count
            page1 = doc[0].get_text().lower()
            content_ok = "feature" in page1 and "prioritization" in page1
            if pages == 5 and content_ok:
                print("PASS: C7 -- PDF has 5 pages with matching content (0.15 pts)")
                total_score += 0.15
            elif pages == 5:
                print("PARTIAL: C7 -- 5 pages but page-1 text mismatch (0.10 pts)")
                total_score += 0.10
            else:
                print(f"PARTIAL: C7 -- PDF exists with {pages} pages (0.05 pts)")
                total_score += 0.05
            doc.close()
        else:
            print(f"FAIL: C7 -- PDF not found at {PDF_PATH}")
    except Exception as e:
        print(f"ERROR: C7 -- {e}")

    final = min(total_score, 1.0)
    print(f"\nScore: {total_score:.3f}/1.0\nREWARD: {final}")
    return final


verify_task()
\end{pythonlisting}

\subsubsection{Cross-App -- PDF $\to$ Calc: Multi-Currency Expense Analysis}
\label{app:walk_crossapp}
A long-horizon cross-application task. The agent must extract structured tabular data from a PDF, populate a spreadsheet with verification formulas, conditionally format outliers, build a summary sheet with SUMIFS, and add a chart. The reward decomposes credit across six independently-checkable components, of which two depend on cell-level value computation, two on cell formatting, and two on derived sheet structure. \texttt{initial\_setup.py} synthesizes a $50$-row, $5$-page PDF source document from scratch using PyMuPDF; \texttt{golden\_patch.py} writes the post-task workbook with the same tabular content plus the verification formulas, fills, summary sheet, and chart.

\begin{promptlisting}{l3\_pcb1\_040 -- task instruction (excerpt)}
Extract the multi-currency expense data from
~/Documents/global_expenses.pdf (columns: Expense ID, Date, Category,
Local Amount, Currency, Exchange Rate to USD, USD Amount, Approver).
Create ~/Documents/global_expenses.xlsx in Calc. Add: (1) verify USD
Amount using Local Amount x Exchange Rate formulas in column I,
(2) discrepancy column J = ABS(G - I), (3) yellow fill on rows where
discrepancy > $0.01, (4) currency exposure summary sheet with SUMIFS
totals, (5) pie chart of currency exposure.
\end{promptlisting}

\noindent
Reward decomposition (6 components, total $1.0$):
\begin{itemize}
\item \textbf{C1 (0.20)}: file exists with 50 data rows and the expected header set.
\item \textbf{C2 (0.15)}: column I uses verification formulas of the form \texttt{=D*F}.
\item \textbf{C3 (0.15)}: column J uses discrepancy formulas of the form \texttt{=ABS(G-I)}.
\item \textbf{C4 (0.20)}: exactly six rows are flagged with yellow fill.
\item \textbf{C5 (0.15)}: a Currency Exposure Summary sheet exists with SUMIFS aggregations.
\item \textbf{C6 (0.15)}: a pie chart is anchored on the summary sheet.
\end{itemize}

This task exemplifies the design principle that no single component is sufficient: an agent that solves the data-extraction and formula sub-problems but skips the chart will earn $0.85$, an agent that builds the chart but skips the discrepancy column will earn $0.85$, and only an agent that completes the full workflow earns $1.0$.

\begin{pythonlisting}{l3\_pcb1\_040/initial\_setup.py}
"""
Initial Setup: Extract multi-currency expense data from PDF and create xlsx
Task ID: l3_pcb1_040
Domain: multi_apps (PDF + LibreOffice Calc)
"""

import os, shlex, subprocess, time

WORKDIR = '/home/user'
DOCUMENTS = f'{WORKDIR}/Documents'
TASK_ID = 'l3_pcb1_040'
PDF_OUTPUT = f'{DOCUMENTS}/global_expenses.pdf'

def launch_gui(command, delay_sec=1.0):
    env = os.environ.copy(); env["DISPLAY"] = ":0"
    subprocess.Popen(shlex.split(command),
        stdout=subprocess.DEVNULL, stderr=subprocess.DEVNULL, env=env)
    time.sleep(delay_sec)

def create_initial():
    os.makedirs(DOCUMENTS, exist_ok=True)
    import pymupdf

    # ---- Define 50 expense rows across 6 currencies ----
    # 6 rows will have rounding discrepancies > $0.01.
    categories = ['Travel', 'Office Supplies', 'Meals & Entertainment',
                  'Software', 'Consulting', 'Telecom', 'Marketing',
                  'Training', 'Shipping', 'Utilities']
    approvers = ['Sarah Chen', 'Marcus Johnson', 'Elena Rodriguez', 'James Liu',
                 'Priya Sharma', 'David Kim', 'Rachel Foster', 'Thomas Becker']
    exchange_rates = {'USD': 1.0000, 'EUR': 1.0842, 'GBP': 1.2635,
                      'JPY': 0.0067, 'CNY': 0.1382, 'CAD': 0.7415}
    discrepancy_indices = {7, 15, 22, 31, 38, 45}
    currency_cycle = ['USD', 'EUR', 'GBP', 'JPY', 'CNY', 'CAD']

    local_amounts = [
        1250.00, 890.50, 3200.00, 475.25, 15600.00, 2100.75, 680.00,
        45000.00, 1890.30, 520.00, 3450.00, 760.25, 28500.00, 1125.50,
        4200.00, 950.75, 125000.00, 2340.60, 810.00, 6750.00, 1575.25,
        430.00, 89000.00, 3680.40, 1290.50, 2850.00, 615.75, 52000.00,
        1740.20, 5100.00, 985.50, 375000.00, 2960.80, 1450.00, 7200.25,
        1680.75, 540.00, 167000.00, 4120.30, 2250.00, 3150.50, 825.25,
        41500.00, 1960.60, 5800.00, 1340.00, 710.50, 93000.00, 2580.40,
        3900.75,
    ]
    dates = [f'2025-{m:02d}-{d:02d}' for (m, d) in [
        (1,8),(1,12),(1,15),(1,19),(1,22),(1,27),(2,3),(2,7),(2,11),(2,14),
        (2,18),(2,22),(2,26),(3,2),(3,5),(3,9),(3,13),(3,17),(3,20),(3,24),
        (3,28),(4,1),(4,4),(4,8),(4,11),(4,15),(4,19),(4,23),(4,26),(4,30),
        (5,3),(5,7),(5,10),(5,14),(5,18),(5,22),(5,25),(5,29),(6,2),(6,5),
        (6,9),(6,13),(6,16),(6,20),(6,23),(6,27),(7,1),(7,4),(7,8),(7,11),
    ]]

    expenses = []
    for i in range(50):
        exp_id = f'EXP-2025-{1001 + i}'
        currency = currency_cycle[i % 6]
        rate = exchange_rates[currency]
        local_amount = local_amounts[i]
        correct_usd = round(local_amount * rate, 2)
        if i in discrepancy_indices:
            offsets = [0.05, -0.03, 0.08, -0.04, 0.06, -0.02]
            disc_idx = sorted(discrepancy_indices).index(i)
            usd_amount = round(correct_usd + offsets[disc_idx], 2)
        else:
            usd_amount = correct_usd
        expenses.append((exp_id, dates[i], categories[i % 10], local_amount,
                         currency, rate, usd_amount, approvers[i % 8]))

    # ---- Render expenses into a 5-page PDF ----
    doc = pymupdf.open()
    header = ['Expense ID', 'Date', 'Category', 'Local Amount', 'Currency',
              'Exchange Rate', 'USD Amount', 'Approver']
    page_width, page_height = 842, 595   # A4 landscape
    margin_left, margin_top = 40, 60
    col_widths = [85, 70, 110, 80, 60, 80, 80, 100]
    row_height, header_height = 18, 22
    rows_per_page = 24

    def draw_header(pg, y):
        x = margin_left
        for j, h in enumerate(header):
            rect = pymupdf.Rect(x, y, x + col_widths[j], y + header_height)
            pg.draw_rect(rect, color=(0.2,0.2,0.5), fill=(0.2,0.3,0.6))
            pg.insert_textbox(rect, h, fontsize=8, fontname="helv",
                              color=(1,1,1), align=pymupdf.TEXT_ALIGN_CENTER)
            x += col_widths[j]
        return y + header_height

    def draw_row(pg, y, row_data, row_idx):
        x = margin_left
        bg = (0.95,0.95,0.98) if row_idx % 2 == 0 else (1,1,1)
        values = [row_data[0], row_data[1], row_data[2],
                  f'{row_data[3]:,.2f}', row_data[4],
                  f'{row_data[5]:.4f}', f'${row_data[6]:,.2f}', row_data[7]]
        for j, v in enumerate(values):
            rect = pymupdf.Rect(x, y, x + col_widths[j], y + row_height)
            pg.draw_rect(rect, color=(0.8,0.8,0.8), fill=bg)
            align = (pymupdf.TEXT_ALIGN_RIGHT if j in (3,5,6)
                     else pymupdf.TEXT_ALIGN_LEFT)
            pg.insert_textbox(rect, v, fontsize=7, fontname="helv",
                              color=(0.1,0.1,0.1), align=align)
            x += col_widths[j]
        return y + row_height

    # Page 1: title + first 20 rows
    page = doc.new_page(width=page_width, height=page_height)
    page.insert_textbox(pymupdf.Rect(margin_left, 20, page_width-40, 45),
        "Global Multi-Currency Expense Report", fontsize=16, fontname="helv",
        color=(0.1,0.1,0.4))
    page.insert_textbox(pymupdf.Rect(margin_left, 45, page_width-40, 60),
        "Fiscal Year 2025 - Q1/Q2 Summary", fontsize=10, fontname="helv",
        color=(0.3,0.3,0.3))
    y = draw_header(page, 85)
    row_counter = 0
    for i in range(min(20, len(expenses))):
        y = draw_row(page, y, expenses[i], i); row_counter += 1

    # Pages 2-5: remaining rows or notes summary
    remaining = expenses[row_counter:]; page_num = 2
    while remaining:
        page = doc.new_page(width=page_width, height=page_height)
        y = draw_header(page, 30)
        page_rows = remaining[:rows_per_page]
        for i, exp in enumerate(page_rows):
            y = draw_row(page, y, exp, row_counter + i)
        row_counter += len(page_rows); remaining = remaining[len(page_rows):]
        page_num += 1
    while page_num <= 5:
        page = doc.new_page(width=page_width, height=page_height)
        page.insert_textbox(pymupdf.Rect(margin_left, 30, page_width-40, 50),
            "Notes and Reconciliation Summary", fontsize=12, fontname="helv",
            color=(0.1,0.1,0.4))
        page.insert_textbox(pymupdf.Rect(margin_left, 55, page_width-40, 200),
            "This report contains 50 expense entries across 6 currencies. "
            "Any discrepancies between Local Amount times Exchange Rate and "
            "the stated USD Amount should be flagged and reviewed.",
            fontsize=9, fontname="helv", color=(0.2,0.2,0.2))
        page_num += 1

    doc.save(PDF_OUTPUT); doc.close()
    print(f'PDF created: {PDF_OUTPUT} with {len(expenses)} expense rows')

    launch_gui(f'evince "{PDF_OUTPUT}"', delay_sec=2.0)
    launch_gui('libreoffice --calc', delay_sec=2.0)
    print('GUI_READY: launched Evince and LibreOffice Calc with DISPLAY=:0')

create_initial()
\end{pythonlisting}

\begin{pythonlisting}{l3\_pcb1\_040/golden\_patch.py}
"""
Golden Patch: Multi-currency expense xlsx with verification formulas, yellow
flags, summary, and pie chart.
Task ID: l3_pcb1_040
Domain: multi_apps (PDF + LibreOffice Calc)
"""

import openpyxl
from openpyxl.styles import Font, PatternFill, Alignment, Border, Side
from openpyxl.chart import PieChart, Reference

WORKDIR = '/home/user'
TASK_ID = 'l3_pcb1_040'
OUTPUT = f'{WORKDIR}/Documents/{TASK_ID}.xlsx'

def create_golden():
    import os
    os.makedirs(f'{WORKDIR}/Documents', exist_ok=True)
    wb = openpyxl.Workbook()

    # ===== Sheet 1: Expense Data =====
    ws = wb.active; ws.title = "Expense Data"
    headers = ['Expense ID', 'Date', 'Category', 'Local Amount', 'Currency',
               'Exchange Rate to USD', 'USD Amount', 'Approver',
               'Verified USD', 'Discrepancy']
    header_font  = Font(name='Calibri', size=11, bold=True, color='FFFFFF')
    header_fill  = PatternFill(start_color='FF2F5496',
                               end_color='FF2F5496', fill_type='solid')
    header_align = Alignment(horizontal='center', vertical='center',
                             wrap_text=True)
    thin_border  = Border(left=Side(style='thin'), right=Side(style='thin'),
                          top=Side(style='thin'),  bottom=Side(style='thin'))
    for col, h in enumerate(headers, 1):
        c = ws.cell(row=1, column=col, value=h)
        c.font = header_font; c.fill = header_fill
        c.alignment = header_align; c.border = thin_border

    # Seed-data lists (categories, approvers, exchange_rates, currency_cycle,
    # local_amounts, dates, discrepancy_indices) are identical to those in
    # initial_setup.py and are omitted here to avoid duplication; the released
    # script reproduces them verbatim.
    # ... ( ~50 lines of seed data lists ) ...
    yellow_fill = PatternFill(start_color='FFFFFF00',
                              end_color='FFFFFF00', fill_type='solid')
    discrepancy_rows = []

    for i in range(50):
        row = i + 2
        exp_id = f'EXP-2025-{1001 + i}'
        currency = currency_cycle[i % 6]; rate = exchange_rates[currency]
        local_amount = local_amounts[i]
        correct_usd = round(local_amount * rate, 2)
        if i in discrepancy_indices:
            offsets = [0.05, -0.03, 0.08, -0.04, 0.06, -0.02]
            disc_idx = sorted(discrepancy_indices).index(i)
            usd_amount = round(correct_usd + offsets[disc_idx], 2)
            discrepancy_rows.append(row)
        else:
            usd_amount = correct_usd

        ws.cell(row=row, column=1,  value=exp_id)
        ws.cell(row=row, column=2,  value=dates[i])
        ws.cell(row=row, column=3,  value=categories[i % 10])
        c_local = ws.cell(row=row, column=4, value=local_amount)
        c_local.number_format = '#,##0.00'
        ws.cell(row=row, column=5, value=currency)
        c_rate = ws.cell(row=row, column=6, value=rate)
        c_rate.number_format = '0.0000'
        c_usd = ws.cell(row=row, column=7, value=usd_amount)
        c_usd.number_format = '$#,##0.00'
        ws.cell(row=row, column=8, value=approvers[i % 8])
        # Column I: Verification formula = Local Amount * Exchange Rate
        c_verify = ws.cell(row=row, column=9, value=f'=D{row}*F{row}')
        c_verify.number_format = '$#,##0.00'
        # Column J: Discrepancy = ABS(USD Amount - Verified USD)
        c_disc = ws.cell(row=row, column=10, value=f'=ABS(G{row}-I{row})')
        c_disc.number_format = '$#,##0.00'
        for col in range(1, 11):
            ws.cell(row=row, column=col).border = thin_border

    # Yellow fill on the six discrepancy rows.
    for disc_row in discrepancy_rows:
        for col in range(1, 11):
            ws.cell(row=disc_row, column=col).fill = yellow_fill

    ws.freeze_panes = 'A2'
    for col_letter, w in {'A':16,'B':12,'C':22,'D':14,'E':10,
                          'F':18,'G':14,'H':18,'I':14,'J':14}.items():
        ws.column_dimensions[col_letter].width = w

    # ===== Sheet 2: Currency Exposure Summary =====
    ws2 = wb.create_sheet("Currency Exposure Summary")
    for col, h in enumerate(['Currency', 'Total USD Exposure'], 1):
        c = ws2.cell(row=1, column=col, value=h)
        c.font = header_font; c.fill = header_fill
        c.alignment = header_align; c.border = thin_border

    currencies = ['USD', 'EUR', 'GBP', 'JPY', 'CNY', 'CAD']
    for idx, curr in enumerate(currencies):
        row = idx + 2
        ws2.cell(row=row, column=1, value=curr).border = thin_border
        formula = (f"=SUMIFS('Expense Data'!G2:G51,"
                   f"'Expense Data'!E2:E51,A{row})")
        c_total = ws2.cell(row=row, column=2, value=formula)
        c_total.number_format = '$#,##0.00'; c_total.border = thin_border

    total_row = len(currencies) + 2
    ws2.cell(row=total_row, column=1, value='Total').font = Font(bold=True)
    c_grand = ws2.cell(row=total_row, column=2,
                       value=f'=SUM(B2:B{total_row-1})')
    c_grand.number_format = '$#,##0.00'; c_grand.font = Font(bold=True)

    # ===== Pie Chart =====
    pie = PieChart()
    pie.title = "USD Exposure by Currency"; pie.style = 10
    data_ref = Reference(ws2, min_col=2, min_row=1, max_row=7)
    cats_ref = Reference(ws2, min_col=1, min_row=2, max_row=7)
    pie.add_data(data_ref, titles_from_data=True)
    pie.set_categories(cats_ref); pie.width = 18; pie.height = 12
    ws2.add_chart(pie, "D2")

    wb.save(OUTPUT)
    print(f'Golden file created: {OUTPUT}')
    print(f'Discrepancy rows (yellow fill): {discrepancy_rows}')

create_golden()
\end{pythonlisting}

\begin{pythonlisting}{l3\_pcb1\_040/reward.py}
"""
Reward Script: Multi-currency expense xlsx verification.
Task ID: l3_pcb1_040
Domain: multi_apps (PDF -> LibreOffice Calc)
Scoring (6 components, total 1.0):
  C1 (0.20): file with 50 data rows and correct headers
  C2 (0.15): column I has =D*F verification formulas
  C3 (0.15): column J has =ABS(G-I) discrepancy formulas
  C4 (0.20): exactly 6 rows flagged with yellow fill
  C5 (0.15): Currency Exposure Summary sheet with SUMIFS
  C6 (0.15): pie chart on the summary sheet
"""

import os, re

WORKDIR = '/home/user/Documents'
TASK_ID = 'l3_pcb1_040'

def verify_task(file_path):
    total_score = 0.0
    try:
        import openpyxl
        wb = openpyxl.load_workbook(file_path)
    except Exception as e:
        print(f"CRITICAL: Cannot load {file_path}: {e}")
        print("REWARD: 0.0"); return 0.0

    # ----------------- C1: File structure (0.20) -----------------
    try:
        data_sheet = None
        for sn in wb.sheetnames:
            ws_check = wb[sn]
            v = ws_check.cell(row=1, column=1).value
            if v and 'expense' in str(v).lower():
                data_sheet = ws_check; break
        if data_sheet is None:
            data_sheet = wb.worksheets[0]

        expected = ['expense id','date','category','local amount','currency',
                    'exchange rate','usd amount','approver']
        actual = [str(data_sheet.cell(row=1, column=c).value).lower().strip()
                  for c in range(1, data_sheet.max_column + 1)
                  if data_sheet.cell(row=1, column=c).value]
        match = sum(1 for e in expected if any(e in a for a in actual))

        n = sum(1 for r in range(2, data_sheet.max_row + 1)
                if data_sheet.cell(row=r, column=1).value is not None)

        if match >= 6 and 45 <= n <= 55:
            print(f"PASS: C1 -- {n} rows, {match}/8 headers (0.20 pts)")
            total_score += 0.20
        else:
            print(f"FAIL: C1 -- {n} rows (need ~50), {match}/8 headers")
    except Exception as e:
        print(f"ERROR: C1 -- {e}")

    # ----------------- C2: Column I verification formulas (0.15) -----------------
    try:
        verify_col = None
        for c in range(1, data_sheet.max_column + 1):
            h = data_sheet.cell(row=1, column=c).value
            if h and 'verif' in str(h).lower():
                verify_col = c; break
        if verify_col is None:
            sample = data_sheet.cell(row=2, column=9).value
            if sample and isinstance(sample, str) and '=' in sample:
                verify_col = 9
        formula_count = 0
        if verify_col:
            for r in range(2, data_sheet.max_row + 1):
                v = data_sheet.cell(row=r, column=verify_col).value
                if v and isinstance(v, str):
                    u = v.upper().replace(' ', '')
                    if '=' in u and '*' in u:
                        formula_count += 1
        if formula_count >= 45:
            print(f"PASS: C2 -- {formula_count} formulas in col {verify_col} (0.15 pts)")
            total_score += 0.15
        else:
            print(f"FAIL: C2 -- only {formula_count} (need >= 45)")
    except Exception as e:
        print(f"ERROR: C2 -- {e}")

    # ----------------- C3: Column J ABS discrepancy formulas (0.15) -----------------
    try:
        discrep_col = None
        for c in range(1, data_sheet.max_column + 1):
            h = data_sheet.cell(row=1, column=c).value
            if h and 'discrep' in str(h).lower():
                discrep_col = c; break
        if discrep_col is None:
            sample = data_sheet.cell(row=2, column=10).value
            if sample and isinstance(sample, str) and 'ABS' in str(sample).upper():
                discrep_col = 10
        abs_count = 0
        if discrep_col:
            for r in range(2, data_sheet.max_row + 1):
                v = data_sheet.cell(row=r, column=discrep_col).value
                if v and isinstance(v, str) and 'ABS' in v.upper().replace(' ', ''):
                    abs_count += 1
        if abs_count >= 45:
            print(f"PASS: C3 -- {abs_count} ABS formulas in col {discrep_col} (0.15 pts)")
            total_score += 0.15
        else:
            print(f"FAIL: C3 -- only {abs_count} (need >= 45)")
    except Exception as e:
        print(f"ERROR: C3 -- {e}")

    # ----------------- C4: Exactly 6 yellow-filled rows (0.20) -----------------
    try:
        yellow_rows = set()
        for r in range(2, data_sheet.max_row + 1):
            for c in range(1, data_sheet.max_column + 1):
                cell = data_sheet.cell(row=r, column=c)
                try:
                    rgb = (cell.fill.fgColor.rgb or '').upper()
                    if len(rgb) == 8:
                        a, R, G, B = (int(rgb[i:i+2], 16) for i in (0,2,4,6))
                        if a > 0 and R > 200 and G > 200 and B < 100:
                            yellow_rows.add(r); break
                except Exception:
                    pass
        n = len(yellow_rows)
        if n == 6:
            print(f"PASS: C4 -- exactly 6 yellow rows: {sorted(yellow_rows)} (0.20 pts)")
            total_score += 0.20
        elif 4 <= n <= 8:
            print(f"PARTIAL: C4 -- {n} yellow rows, partial credit (0.10 pts)")
            total_score += 0.10
        else:
            print(f"FAIL: C4 -- {n} yellow rows (expected 6)")
    except Exception as e:
        print(f"ERROR: C4 -- {e}")

    # ----------------- C5: Currency Exposure Summary with SUMIFS (0.15) -----------------
    try:
        summary_sheet = None
        for sn in wb.sheetnames:
            if any(k in sn.lower() for k in ('summar','exposure','currency')):
                summary_sheet = wb[sn]; break

        if summary_sheet is None:
            print("FAIL: C5 -- no summary sheet found")
        else:
            need = {'USD','EUR','GBP','JPY','CNY','CAD'}
            found = set(); sumifs_count = 0
            for r in range(1, summary_sheet.max_row + 1):
                v = summary_sheet.cell(row=r, column=1).value
                f = summary_sheet.cell(row=r, column=2).value
                if v and str(v).upper() in need:
                    found.add(str(v).upper())
                    if f and isinstance(f, str) and 'SUMIF' in f.upper():
                        sumifs_count += 1
            if found >= need and sumifs_count >= 5:
                print(f"PASS: C5 -- {len(found)} currencies, {sumifs_count} SUMIFS (0.15 pts)")
                total_score += 0.15
            elif sumifs_count >= 3:
                print(f"PARTIAL: C5 -- {sumifs_count} SUMIFS, {len(found)} currencies (0.08 pts)")
                total_score += 0.08
            else:
                print(f"FAIL: C5 -- {sumifs_count} SUMIFS, {len(found)} currencies")
    except Exception as e:
        print(f"ERROR: C5 -- {e}")

    # ----------------- C6: Pie chart present (0.15) -----------------
    try:
        chart_found = False
        for sn in wb.sheetnames:
            for ch in wb[sn]._charts:
                if 'pie' in ch.__class__.__name__.lower():
                    print(f"PASS: C6 -- PieChart on '{sn}' (0.15 pts)")
                    total_score += 0.15; chart_found = True; break
            if chart_found: break
        if not chart_found:
            for sn in wb.sheetnames:
                if wb[sn]._charts:
                    print(f"PARTIAL: C6 -- non-pie chart on '{sn}' (0.08 pts)")
                    total_score += 0.08; break
            else:
                print("FAIL: C6 -- no chart found")
    except Exception as e:
        print(f"ERROR: C6 -- {e}")

    final = min(round(total_score, 2), 1.0)
    print(f"\nScore: {total_score}/1.0\nREWARD: {final}")
    return final

# Entry point
file_path = f'{WORKDIR}/{TASK_ID}.xlsx'
alt_path  = f'{WORKDIR}/global_expenses.xlsx'
target = file_path if os.path.exists(file_path) else (
         alt_path  if os.path.exists(alt_path) else None)
if target:
    verify_task(target)
else:
    print(f"File not found: {file_path} or {alt_path}\nREWARD: 0.0")
\end{pythonlisting}

\subsection{Full Rollout Trajectory: Shopify Vendor Consolidation}
\label{app:traj_viz}

We render one validation-set rollout in full step-by-step form, drawn from the OSWorld-Verified evaluation of an RL-trained checkpoint on a synthesized Shopify-admin mock (\texttt{shopify\_adv\_005}, terminal reward $1.0$, $19$ turns). The task requires the policy to (i) identify which products belong to two specific vendors (\texttt{BasicWear}, \texttt{HomeGoods}), (ii) update each one's vendor to \texttt{UnifiedBrands}, (iii) append a fixed marketing line to each affected product description, and (iv) leave all other products untouched. Each step shows the screen the policy observes, the natural-language reasoning preamble it emits before its tool calls, and the verbatim \texttt{<tool\_call>} blocks. The trajectory illustrates several patterns called out in the main text: deterministic action chaining within a single turn (steps 13--14, 16--18), per-product completion checks (step 15), and the recovery pattern of \S\ref{sec:multi_action} after a navigation slip (step 17 returns to the product list after the previous click failed to open the target page).

\begin{promptlisting}{Task instruction}
I want to do a vendor consolidation. Change the vendor for all products
currently under 'BasicWear' and 'HomeGoods' to 'UnifiedBrands'. Then
update the product descriptions for those items to include 'Now part of
the UnifiedBrands family' at the end. Don't touch products from other
vendors.
\end{promptlisting}

\trajstep{1}
\begin{center}\trajimg{01}\end{center}
\thought{I can see the Shopify Mock Admin dashboard. Let me navigate to the Products section to see all products.}
\begin{actionbox}
<tool_call>
<function=computer_use>
<parameter=action>left_click</parameter>
<parameter=coordinate>[72, 270]</parameter>
</function>
</tool_call>
\end{actionbox}

\trajstep{2}
\begin{center}\trajimg{02}\end{center}
\thought{I can see the Products page with 4 products: \emph{Classic T-Shirt} (BasicWear), \emph{Leather Wallet} (LeatherCo), \emph{Running Shoes} (SportStep), and \emph{Ceramic Mug} (HomeGoods). I need to update vendors for the BasicWear and HomeGoods products only. Let me first click on \emph{Classic T-Shirt} to edit it.}
\begin{actionbox}
<tool_call>
<function=computer_use>
<parameter=action>left_click</parameter>
<parameter=coordinate>[422, 393]</parameter>
</function>
</tool_call>
\end{actionbox}

\trajstep{3}
\begin{center}\trajimg{03}\end{center}
\thought{I can see the Classic T-Shirt product page. The vendor is \texttt{BasicWear}\,---\,I need to change it to \texttt{UnifiedBrands}. Let me select the vendor field.}
\begin{actionbox}
<tool_call>
<function=computer_use>
<parameter=action>triple_click</parameter>
<parameter=coordinate>[741, 491]</parameter>
</function>
</tool_call>
\end{actionbox}

\trajstep{4}
\begin{center}\trajimg{04}\end{center}
\thought{The vendor field \texttt{BasicWear} is now selected. Let me type \texttt{UnifiedBrands} to replace it.}
\begin{actionbox}
<tool_call>
<function=computer_use>
<parameter=action>type</parameter>
<parameter=text>UnifiedBrands</parameter>
</function>
</tool_call>
\end{actionbox}

\trajstep{5}
\begin{center}\trajimg{05}\end{center}
\thought{Now I need to update the description to add the marketing line at the end. Let me click on the description field.}
\begin{actionbox}
<tool_call>
<function=computer_use>
<parameter=action>left_click</parameter>
<parameter=coordinate>[490, 393]</parameter>
</function>
</tool_call>
\end{actionbox}

\trajstep{6}
\begin{center}\trajimg{06}\end{center}
\thought{I need to select all the text in the description field and replace it with the original text plus the new line. Let me use Ctrl+A to select all.}
\begin{actionbox}
<tool_call>
<function=computer_use>
<parameter=action>key</parameter>
<parameter=keys>["ctrl", "a"]</parameter>
</function>
</tool_call>
\end{actionbox}

\trajstep{7}
\begin{center}\trajimg{07}\end{center}
\thought{Good, the text is selected. Now type the new description with the original copy plus the marketing line.}
\begin{actionbox}
<tool_call>
<function=computer_use>
<parameter=action>type</parameter>
<parameter=text><p>Comfortable cotton t-shirt</p><p>Now part of the UnifiedBrands family</p></parameter>
</function>
</tool_call>
\end{actionbox}

\trajstep{8}
\begin{center}\trajimg{08}\end{center}
\thought{Both fields are updated. Let me click \emph{Save} to commit this product.}
\begin{actionbox}
<tool_call>
<function=computer_use>
<parameter=action>left_click</parameter>
<parameter=coordinate>[802, 190]</parameter>
</function>
</tool_call>
\end{actionbox}

\trajstep{9}
\begin{center}\trajimg{09}\end{center}
\thought{Classic T-Shirt has been saved with vendor \texttt{UnifiedBrands}. Now let me click on \emph{Ceramic Mug} (the other affected product) to update it next.}
\begin{actionbox}
<tool_call>
<function=computer_use>
<parameter=action>left_click</parameter>
<parameter=coordinate>[420, 571]</parameter>
</function>
</tool_call>
\end{actionbox}

\trajstep{10}
\begin{center}\trajimg{10}\end{center}
\thought{I can see the Ceramic Mug product page with vendor \texttt{HomeGoods}. Let me update the vendor field first.}
\begin{actionbox}
<tool_call>
<function=computer_use>
<parameter=action>triple_click</parameter>
<parameter=coordinate>[741, 491]</parameter>
</function>
</tool_call>
\end{actionbox}

\trajstep{11}
\begin{center}\trajimg{11}\end{center}
\thought{Type \texttt{UnifiedBrands} into the selected vendor field.}
\begin{actionbox}
<tool_call>
<function=computer_use>
<parameter=action>type</parameter>
<parameter=text>UnifiedBrands</parameter>
</function>
</tool_call>
\end{actionbox}

\trajstep{12}
\begin{center}\trajimg{12}\end{center}
\thought{Now click on the description field to update Ceramic Mug's description.}
\begin{actionbox}
<tool_call>
<function=computer_use>
<parameter=action>left_click</parameter>
<parameter=coordinate>[490, 393]</parameter>
</function>
</tool_call>
\end{actionbox}

\trajstep{13}
\begin{center}\trajimg{13}\end{center}
\thought{Now select-all, type the new description, and save\,---\,emit the entire chain in one turn.}
\begin{actionbox}
<tool_call>
<function=computer_use>
<parameter=action>key</parameter>
<parameter=keys>["ctrl", "a"]</parameter>
</function>
</tool_call>
<tool_call>
<function=computer_use>
<parameter=action>type</parameter>
<parameter=text><p>Hand-crafted ceramic mug</p><p>Now part of the UnifiedBrands family</p></parameter>
</function>
</tool_call>
<tool_call>
<function=computer_use>
<parameter=action>key</parameter>
<parameter=keys>["ctrl", "s"]</parameter>
</function>
</tool_call>
<tool_call>
<function=computer_use>
<parameter=action>left_click</parameter>
<parameter=coordinate>[802, 190]</parameter>
</function>
</tool_call>
\end{actionbox}

\trajstep{14}
\begin{center}\trajimg{14}\end{center}
\thought{A save dialog appeared from the Ctrl+S\,---\,let me cancel it and use the page's Save button instead.}
\begin{actionbox}
<tool_call>
<function=computer_use>
<parameter=action>left_click</parameter>
<parameter=coordinate>[318, 356]</parameter>
</function>
</tool_call>
<tool_call>
<function=computer_use>
<parameter=action>left_click</parameter>
<parameter=coordinate>[802, 190]</parameter>
</function>
</tool_call>
\end{actionbox}

\trajstep{15}
\begin{center}\trajimg{15}\end{center}
\thought{Both products have been updated. The product list now shows \emph{Classic T-Shirt} and \emph{Ceramic Mug} on \texttt{UnifiedBrands}, while \emph{Leather Wallet} (LeatherCo) and \emph{Running Shoes} (SportStep) are unchanged. Let me re-open Classic T-Shirt to verify the change persisted.}
\begin{actionbox}
<tool_call>
<function=computer_use>
<parameter=action>left_click</parameter>
<parameter=coordinate>[422, 393]</parameter>
</function>
</tool_call>
<tool_call>
<function=computer_use>
<parameter=action>screenshot</parameter>
</function>
</tool_call>
\end{actionbox}

\trajstep{16}
\begin{center}\trajimg{16}\end{center}
\thought{Classic T-Shirt looks good\,---\,vendor is \texttt{UnifiedBrands} and the description includes the marketing line. Now verify Ceramic Mug.}
\begin{actionbox}
<tool_call>
<function=computer_use>
<parameter=action>key</parameter>
<parameter=keys>["esc"]</parameter>
</function>
</tool_call>
<tool_call>
<function=computer_use>
<parameter=action>left_click</parameter>
<parameter=coordinate>[420, 571]</parameter>
</function>
</tool_call>
<tool_call>
<function=computer_use>
<parameter=action>screenshot</parameter>
</function>
</tool_call>
\end{actionbox}

\trajstep{17}
\begin{center}\trajimg{17}\end{center}
\thought{The previous click did not navigate to Ceramic Mug. Let me go back to the product list.}
\begin{actionbox}
<tool_call>
<function=computer_use>
<parameter=action>left_click</parameter>
<parameter=coordinate>[342, 190]</parameter>
</function>
</tool_call>
<tool_call>
<function=computer_use>
<parameter=action>screenshot</parameter>
</function>
</tool_call>
\end{actionbox}

\trajstep{18}
\begin{center}\trajimg{18}\end{center}
\thought{The product list is back. All four products show the expected vendors. Let me click into Ceramic Mug to confirm its description.}
\begin{actionbox}
<tool_call>
<function=computer_use>
<parameter=action>left_click</parameter>
<parameter=coordinate>[420, 571]</parameter>
</function>
</tool_call>
<tool_call>
<function=computer_use>
<parameter=action>screenshot</parameter>
</function>
</tool_call>
\end{actionbox}

\trajstep{19}
\begin{center}\trajimg{19}\end{center}
\thought{Both products have been verified successfully. \emph{Ceramic Mug}: vendor = \texttt{UnifiedBrands}, description includes \texttt{Now part of the UnifiedBrands family}. \emph{Leather Wallet} and \emph{Running Shoes} were not touched, as required. The task is complete; the trajectory terminates with reward $1.0$.}

\paragraph{Notes on the trajectory.}
The trajectory exhibits the multi-action emergence documented in \S\ref{sec:multi_action}: steps 1--12 emit a single tool call per turn, but from step 13 onward the policy starts batching deterministic action chains (\texttt{select-all $\to$ type $\to$ save}) into single turns. Step 14 illustrates a clean recovery: after the previous turn's Ctrl+S surfaced an unexpected ``save page as'' dialog, the policy issued a two-action correction (close dialog, click in-page Save) rather than abandoning the chain. Steps 15--18 are the verification chain; the policy interleaves a navigation action and a \texttt{screenshot} call within the same turn so it can re-observe state without spending an extra turn. The final step (19) terminates with no tool call, signalling task completion.

\subsection{Reward Hacking Cases Caught by the Information Barrier}
\label{app:hack_cases}

We illustrate three categories of reward-hacking failure that the information barrier (\S\ref{app:info_barrier}) and forbidden-pattern static scan (\S\ref{app:forbidden_patterns}) are designed to prevent. Each category corresponds to a distinct exploit class observed in early prototyping where the Discriminator was given access to the Generator's setup scripts; in the released pipeline, every example below is rejected at write-time before the loop accepts the tuple.

\paragraph{Hack A -- direct Boolean assignment.}
The Discriminator inspects the Generator's \texttt{golden\_patch.py}, observes that the agent need only modify a single sheet attribute, and writes a reward that asserts a flag without computing it.

\begin{pythonlisting}{Hack A: direct Boolean assignment (REJECTED by static scan)}
# FORBIDDEN PATTERN: directly asserts a flag without verifying.
chart_verified = True            # constant assignment
if chart_verified:               # always true
    score += 0.5
\end{pythonlisting}

\paragraph{Hack B -- bare existence scoring.}
The reward awards credit purely on file existence, sidestepping any inspection of contents. This passes for an agent that creates an empty file at the right path.

\begin{pythonlisting}{Hack B: bare existence scoring (REJECTED by static scan)}
# FORBIDDEN PATTERN: scores file existence without inspecting contents.
if os.path.exists('/home/user/report.xlsx'):
    score += 0.3
\end{pythonlisting}

In each case the offending tuple is rejected at the loop's static-scan stage, the Discriminator receives the matched-pattern name as feedback, and re-emits a reward that performs actual verification.

\subsection{Adversarial Loop Iteration Trace}
\label{app:loop_trace}

We trace one production task through two adversarial rounds. The task (\texttt{calc\_gao\_008}) requires the agent to add two named scenarios to a workbook, where each scenario stores a distinct $5$-tuple of cell values; the reward parses the workbook's underlying XLSX XML and asserts both scenario names plus value vectors.

\paragraph{Round 1 (FAIL).}
The Generator's first \texttt{golden\_patch.py} introduces both scenarios but accidentally populates the \emph{Optimistic} scenario with the \emph{Pessimistic} values (a copy-paste regression). The Discriminator detects the mismatch when scoring the golden-state reward and emits a structured \texttt{REVIEW.md} with the failing component identified.

\begin{plainlisting}{Round 1 REVIEW.md (synthesized from Round-2 feedback trace)}
# Round 1 Review

## Verdict: FAIL

## Agreement Conditions
| # | Condition                    | Status | Details                                          |
| 1 | initial_setup runs           | PASS   | xlsx present with B2:B6 values, no scenarios     |
| 2 | golden_patch runs            | PASS   | xlsx present with two scenarios named correctly  |
| 3 | reward(golden) == 1.0        | FAIL   | Score: 0.65 -- Optimistic values do not match    |
| 4 | reward(initial) == 0.0       | PASS   | Score: 0.0                                       |
| 5 | No forbidden patterns        | PASS   | Clean                                            |

## Feedback for Setup-Gen
The Optimistic scenario currently stores B2:B6 = (350000, 140000, 100000,
60000, 30000), which are the same values as the Pessimistic scenario.
The task requires Optimistic to use the increased values (650000, 280000,
200000, 140000, 75000). Please fix golden_patch.py to inject distinct
inputCells for Optimistic.
\end{plainlisting}

\paragraph{Round 2 (PASS).}
The Generator addresses the round-1 feedback by writing distinct \texttt{inputCells} for the two scenarios. The Discriminator re-scores both endpoint states and accepts the tuple.

\begin{plainlisting}{Round 2 REVIEW.md (verbatim from production run)}
# Round 2 Review

## Verdict: PASS

## Agreement Conditions
| # | Condition                    | Status | Details                                          |
| 1 | initial_setup runs           | PASS   | xlsx present with B2:B6 values, no scenarios     |
| 2 | golden_patch runs            | PASS   | xlsx contains both Optimistic and Pessimistic    |
| 3 | reward(golden) == 1.0        | PASS   | Score: 1.0                                       |
| 4 | reward(initial) == 0.0       | PASS   | Score: 0.0                                       |
| 5 | No forbidden patterns        | PASS   | Clean                                            |

## Scoring Breakdown -- Golden File
| Component                                     | Points | Status |
| Two scenarios named Optimistic and Pessimistic | 0.30   | PASS   |
| Optimistic values (650k, 280k, 200k, 140k, 75k)| 0.35   | PASS   |
| Pessimistic values (350k, 140k, 100k, 60k, 30k)| 0.35   | PASS   |
| Total                                          | 1.00   |        |

## Feedback for Setup-Gen
No issues found. The Round 1 feedback was addressed: the Optimistic
scenario now correctly contains the increased values instead of
duplicating the Pessimistic values. Both scenarios are properly stored
in the XLSX XML with distinct inputCells.
\end{plainlisting}

This pattern -- a tightly-scoped diagnostic in Round 1 followed by a targeted fix in Round 2 -- accounts for the majority of multi-round convergences in the released dataset.

\subsection{Reward Function Gallery}
\label{app:reward_gallery}

To complement the four walkthroughs above, we sample three reward-function patterns whose verification primitives differ qualitatively from the file-and-formula pattern of \S\ref{app:walk_calc}. Together with the walkthroughs they cover the principal verification primitives used across the released dataset.

\paragraph{Chart introspection (LibreOffice Calc).}
Charts are stored as drawing-anchored objects on a sheet; the reward enumerates them and inspects type, data-range, title, and axis labels via \texttt{openpyxl}'s chart abstractions.

\begin{pythonlisting}{Chart introspection: type, data range, title, axis labels}
# C1: chart exists and is a column/bar (0.30 + 0.20 pts)
charts = ws._charts
if len(charts) >= 1:
    score += 0.30
    chart = charts[0]
    if isinstance(chart, openpyxl.chart.BarChart) and chart.type == 'col':
        score += 0.20

# C3: chart references the 12-month range (0.20 pts)
ref = chart.series[0].val.numRef.f       # e.g., 'MonthlySales!$B$2:$B$13'
if re.search(r'\$B\$2:\$B\$13', ref):
    score += 0.20

# C4 / C5: chart and axis titles present (0.15 + 0.15 pts)
if get_title_text(chart.title):
    score += 0.15
if get_title_text(chart.x_axis.title) and get_title_text(chart.y_axis.title):
    score += 0.15
\end{pythonlisting}

\paragraph{Image-property verification (LibreOffice Calc).}
A page-setup task that includes inserting a watermark image. The reward inspects orientation, margins, and the image inventory of the target sheet, treating the page-size and order as preconditions (LibreOffice defaults that exist in any saved \texttt{.xlsx}) rather than scoring components.

\begin{pythonlisting}{Image-property verification: orientation, margins, watermark}
# C1: orientation = landscape (0.25 pts)
if ws.page_setup.orientation == 'landscape':
    score += 0.25

# C2: all four margins set to 1.5cm = ~0.591 inches (0.35 pts)
INCHES_PER_CM = 1 / 2.54
target = 1.5 * INCHES_PER_CM
m = ws.page_margins
if all(abs(getattr(m, k) - target) < 0.02
       for k in ('left', 'right', 'top', 'bottom')):
    score += 0.35

# C3: at least one image anchored on the Print sheet (0.40 pts)
if any(hasattr(img, 'anchor') for img in ws._images):
    score += 0.40
\end{pythonlisting}

\paragraph{Filesystem hierarchy verification (multi-app OS task).}
A file-classification task: the agent must move six generically-named files into category-specific subfolders. The reward checks both presence in the target folder \emph{and} absence from the Desktop root, ensuring that copy-without-delete failures are not awarded credit.

\begin{pythonlisting}{Filesystem hierarchy: per-file move verification}
FILE_CLASSIFICATION = {
    'doc1.pdf':  'Invoices',
    'doc5.xlsx': 'Invoices',
    'doc2.docx': 'Resumes',
    'doc4.odt':  'Resumes',
    'doc3.pdf':  'Reports',
    'doc6.txt':  'Reports',
}
POINTS_PER_FILE = 1.0 / len(FILE_CLASSIFICATION)

for fname, expected_dir in FILE_CLASSIFICATION.items():
    in_target = os.path.exists(os.path.join(DESKTOP, expected_dir, fname))
    not_in_root = not os.path.exists(os.path.join(DESKTOP, fname))
    if in_target and not_in_root:
        score += POINTS_PER_FILE
\end{pythonlisting}

The three primitives shown here -- structural object introspection, page-property and image-anchor checks, and filesystem hierarchy assertions -- complement the cell-formula, JSON-key, and document-content checks already illustrated in \S\ref{app:walkthroughs}. The full released dataset's reward functions span these primitives in proportion to their target-task distribution.

% =====================================================
\section{Dataset Statistics}
\label{app:dataset_stats}

\subsection{Distributional Breakdowns}
\label{app:distributions}

\subsubsection{Tasks per Environment}
\label{app:tasks_per_env}
\ourwork{}'s \ourdata{} verified tuples are distributed across \ourenv{}
environments, with the per-coarse-category counts summarized in
Table~\ref{tab:tasks_per_scenario}. The distribution is moderately
skewed: knowledge-work-heavy scenarios (Spreadsheet, Document Writing,
Presentation, Software Engineering, System Administration) absorb the
top five slots and together cover $84\%$ of the corpus, with the
remaining mass spread across PDF, communication, project-management,
and business-operations scenarios. No single scenario exceeds $21\%$ of
the total, consistent with the balance constraint of
\S\ref{app:coverage}. Within each scenario, tasks are spread across
multiple constituent applications (e.g., the \emph{Spreadsheet \& Data}
scenario covers \texttt{libreoffice\_calc}, the data-analysis mock
websites, and the cross-application combinations involving Calc).

\begin{table}[h]
\centering
\footnotesize
\setlength{\tabcolsep}{8pt}
\renewcommand{\arraystretch}{1.0}
\begin{tabular}{@{}lrr@{}}
\toprule
\textbf{Scenario} & \textbf{Tasks} & \textbf{\%} \\
\midrule
Spreadsheet \& Data            & 6{,}684 & 20.8 \\
Document Writing               & 6{,}314 & 19.7 \\
Presentation Design            & 5{,}328 & 16.6 \\
Software Engineering           & 4{,}434 & 13.8 \\
System Administration          & 4{,}225 & 13.2 \\
PDF \& Publishing              & 2{,}055 &  6.4 \\
Communication \& Email         &    939 &  2.9 \\
Project \& Task Management     &    879 &  2.7 \\
Business Operations            &    840 &  2.6 \\
Creative \& Media              &    412 &  1.3 \\
\bottomrule
\end{tabular}
\caption{Task counts per scenario in the released \ourwork{} corpus.}
\label{tab:tasks_per_scenario}
\end{table}

\subsubsection{Domain $\times$ Difficulty Matrix}
\label{app:cat_diff_matrix}
Table~\ref{tab:domain_diff} cross-tabulates task counts by domain and
difficulty. Two patterns are visible. First, the corpus is skewed
toward harder regimes overall: $44.6\%$ of tasks are labeled
\emph{hard} and $37.7\%$ \emph{medium}, with only $17.7\%$
\emph{easy}---matching the design target articulated in
\S\ref{sec:env_synthesis}. Second, this skew is most pronounced in
multi-application domains: \emph{Cross-Desktop} is $80.1\%$ hard,
\emph{Cross-Web} is $87.6\%$ hard, and \emph{Desktop $\times$ Web} is
$95.8\%$ hard, reflecting the additional state-tracking and
context-switching demands those tasks impose. Single-application
domains exhibit a gentler difficulty gradient (e.g.,
\emph{Spreadsheet}: $17.8\%$ easy, $47.2\%$ medium, $35.0\%$ hard).

\begin{table}[h]
\centering
\footnotesize
\setlength{\tabcolsep}{6pt}
\renewcommand{\arraystretch}{1.0}
\begin{tabular}{@{}lrrrr@{}}
\toprule
\textbf{Domain} & \textbf{Easy} & \textbf{Medium} & \textbf{Hard} & \textbf{Total} \\
\midrule
Cross-Desktop                  &   300 & 1{,}198 & 6{,}030 & 7{,}528 \\
Spreadsheet                    &   993 & 2{,}627 & 1{,}952 & 5{,}572 \\
Document Editing               &   881 & 2{,}070 & 1{,}165 & 4{,}116 \\
Presentation                   &   889 & 1{,}541 &    737 & 3{,}167 \\
System \& OS                   & 1{,}110 & 1{,}388 &    311 & 2{,}809 \\
Code Editing                   &   690 & 1{,}130 &    660 & 2{,}480 \\
Desktop $\times$ Web           &     0 &     84 & 1{,}917 & 2{,}001 \\
PDF                            &   113 &    946 &    757 & 1{,}816 \\
Cross-Web                      &     0 &     53 &    376 &    429 \\
E-Commerce                     &   166 &    191 &     27 &    384 \\
Cloud \& Productivity          &    93 &    166 &    116 &    375 \\
Social Media                   &   127 &    191 &     21 &    339 \\
Project Management             &    82 &    135 &     58 &    275 \\
Communication                  &    65 &    120 &     37 &    222 \\
Business \& CRM                &    37 &     87 &     71 &    195 \\
Media                          &    80 &     68 &      2 &    150 \\
Development Tools              &    28 &     55 &     65 &    148 \\
Image Editing                  &    30 &     53 &     23 &    106 \\
\midrule
\textbf{Total}                 & 5{,}684 & 12{,}103 & 14{,}325 & 32{,}112 \\
\bottomrule
\end{tabular}
\caption{Task counts by domain and difficulty across the released \ourwork{} corpus. Rows omit two domains with $<\!10$ tasks each (\emph{Other}, \emph{Web Browsing}).}
\label{tab:domain_diff}
\end{table}

\subsubsection{Action Verb Frequency}
\label{app:verb_freq}
The first imperative verb of each task instruction (skipping
filler clauses like ``I want to\ldots'' or ``Please\ldots'') summarizes
the action vocabulary the agent must learn to ground. The top-20
imperatives are listed in Table~\ref{tab:verb_freq}. Surface-edit
verbs (\texttt{open}, \texttt{create}, \texttt{set}, \texttt{add},
\texttt{insert}, \texttt{change}, \texttt{remove}) dominate the head;
mid-frequency verbs cover composition (\texttt{build},
\texttt{configure}, \texttt{apply}), extraction
(\texttt{extract}, \texttt{export}, \texttt{convert}), and
verification (\texttt{check}, \texttt{find}, \texttt{read}). The long
tail (not shown) extends to verbs that appear in fewer than $50$
tasks each, such as \texttt{merge}, \texttt{flatten},
\texttt{normalize}, \texttt{reconcile}, and \texttt{annotate}.

\begin{table}[h]
\centering
\footnotesize
\setlength{\tabcolsep}{8pt}
\begin{tabular}{@{}lrlrlr@{}}
\toprule
\textbf{Verb} & \textbf{Count} & \textbf{Verb} & \textbf{Count} & \textbf{Verb} & \textbf{Count} \\
\midrule
open      & 4{,}585 & insert    &   562 & remove  & 227 \\
create    & 2{,}728 & build     &   440 & export  & 183 \\
set       & 1{,}448 & read      &   427 & check   & 170 \\
add       & 1{,}028 & change    &   361 & find    & 139 \\
use       &    906 & write     &   290 &        &     \\
extract   &    891 & convert   &   265 &        &     \\
configure &    728 &           &       &        &     \\
apply     &    605 &           &       &        &     \\
\bottomrule
\end{tabular}
\caption{Top imperative verbs across the \ourwork{} task corpus, grouped by frequency band.}
\label{tab:verb_freq}
\end{table}

\subsubsection{Cross-App Pair Co-occurrence}
\label{app:crossapp_matrix}
Of the $12{,}311$ \emph{cross-app} tasks ($38.3\%$ of the corpus,
\S\ref{app:coverage}), the most frequent application pairs are
listed in Table~\ref{tab:crossapp_pairs}. Pairs are detected by
keyword matches in the task instruction (e.g., ``\texttt{.xlsx}'' or
``Calc'' for \texttt{calc}); a single instruction may contribute to
multiple pairs when it mentions three or more applications. Document
production triplets (\texttt{calc}+\texttt{writer}+\texttt{pdf} and
\texttt{impress}+\texttt{writer}+\texttt{pdf}) account for the bulk
of cross-app traffic, reflecting the empirical reality that
knowledge-work pipelines frequently terminate in a PDF artifact.
Developer pairs (\texttt{terminal}+\texttt{vscode},
\texttt{files}+\texttt{vscode}) form a smaller but distinctive
cluster; web-mock pairs are concentrated in the
salesforce/gmail/slack/notion family (not shown).

\begin{table}[h]
\centering
\footnotesize
\setlength{\tabcolsep}{6pt}
\begin{tabular}{@{}lr@{\quad}lr@{\quad}lr@{}}
\toprule
\textbf{Pair} & \textbf{Count} & \textbf{Pair} & \textbf{Count} & \textbf{Pair} & \textbf{Count} \\
\midrule
pdf + writer       & 3{,}173 & impress + pdf      & 1{,}381 & files + pdf      & 316 \\
calc + writer      & 3{,}129 & calc + pdf         & 1{,}003 & files + terminal & 315 \\
impress + writer   & 2{,}056 & files + writer     &    534 & calc + files     & 313 \\
calc + impress     & 1{,}576 & terminal + vscode  &    412 & calc + vscode    & 294 \\
                   &       & vscode + writer    &    364 & terminal + writer&  272 \\
                   &       &                    &       & files + vscode   & 268 \\
\bottomrule
\end{tabular}
\caption{Top application pairs co-occurring in cross-app task instructions. Counts are per-pair (a four-way instruction increments all six pairs).}
\label{tab:crossapp_pairs}
\end{table}

\subsubsection{Instruction Length Distribution}
\label{app:len_distributions}
Task instruction length (in whitespace-separated word tokens) is
moderately right-skewed: mean $54.1$, median $41$, with $90\%$ of
tasks fitting under $108$ words and a long tail extending to $390$
words for the most elaborate cross-app workflows. The bucketed
distribution is reported in Table~\ref{tab:instr_len}. Instructions
under $30$ words are typically single-step actions on a familiar
artifact (e.g., ``open \texttt{report.xlsx} and freeze the first row''),
while instructions of $100$+ words encode multi-stage workflows with
several conditional branches and explicit acceptance criteria.

\begin{table}[h]
\centering
\footnotesize
\setlength{\tabcolsep}{8pt}
\begin{tabular}{@{}lrr@{}}
\toprule
\textbf{Length (words)} & \textbf{Tasks} & \textbf{\%} \\
\midrule
$0$--$29$        & 10{,}988 & 34.2 \\
$30$--$59$       & 11{,}010 & 34.3 \\
$60$--$99$       &  6{,}073 & 18.9 \\
$100$--$149$     &  2{,}775 &  8.6 \\
$150$--$249$     &  1{,}042 &  3.2 \\
$250$+           &    234 &  0.7 \\
\midrule
\multicolumn{3}{@{}l}{\emph{Summary statistics:} mean $54.1$; p10 $17$; p25 $24$; p50 $41$;} \\
\multicolumn{3}{@{}l}{\hspace{6em}p75 $69$; p90 $108$; p99 $237$; max $390$.} \\
\bottomrule
\end{tabular}
\caption{Word-length distribution of \ourwork{} task instructions.}
\label{tab:instr_len}
\end{table}

\subsection{Generation Cost}
\label{app:gen_cost}

\subsubsection{Per-Tuple Token Consumption}
\label{app:tokens}
Aggregating across the four LLM-driven pipeline stages
(Task-Gen, Generator, Discriminator, Filter), a single accepted
tuple consumes on average $\sim\!10{,}000$ input tokens and
$\sim\!5{,}000$ output tokens of Claude Sonnet~4.6. The input tally
is dominated by the Generator and Discriminator system prompts plus
their REVIEW.md context windows; output is dominated by the
Discriminator's reward script and the Generator's setup/golden
patches. At the public Claude Sonnet pricing of $\$3$ / million
input tokens and $\$15$ / million output tokens~\citep{claude_sonnet},
this works out to $0.030 + 0.075 \approx \$0.11$ of LLM spend per
verified tuple. Scaled to the released corpus of \ourdata{} tuples,
total LLM cost is approximately $\$3{,}300$ to $\$3{,}500$ depending
on the per-task acceptance ratio.

\subsubsection{VM-Hours per Verified Tuple}
\label{app:vm_hours}
The dual-VM execution stage (provisioning the
\texttt{initial\_env}/\texttt{golden\_env} pair, executing
\texttt{initial\_setup.py} and \texttt{golden\_patch.py}, running the
reward function on each, tearing down) takes approximately $45$
minutes of wall-clock time per accepted tuple, run on Aliyun ECS
\texttt{g8i.xlarge} instances ($4$ vCPU, $16$ GB RAM, on-demand
pricing $\sim\!\$0.21$ per instance-hour at the time of writing).
Two instances are held for the duration of the verification, so
amortized VM cost is $0.75 \times 2 \times \$0.21 \approx \$0.32$
per verified tuple; the released corpus thus carries $\sim\!\$10$K
of cumulative VM compute. VM-hours dominate the marginal cost of a
verified tuple by roughly $3\!\times$ over LLM cost. Concrete
breakdown: snapshot restore ($\sim\!2$\,min), \texttt{initial\_setup}
execution ($\sim\!10$\,min on average, with a long tail driven by
data-fixture downloads), \texttt{golden\_patch} execution
($\sim\!8$\,min), reward execution including post-config
($\sim\!15$\,min, dominated by document re-render and assertion
parsing), and teardown ($\sim\!10$\,min including disk-image reset).

\iffalse  % yield-rate breakdown: deferred (numbers pending pipeline-log audit).
\subsubsection{Pipeline Yield Rate}
\label{app:yield_rate}
\stub{End-to-end yield: tasks generated $\to$ loop-accepted $\to$ filter-passed.}
\fi

% =====================================================
\section{Prompts Catalog}
\label{app:prompts}

\subsection{Prompt Engineering Notes}
\label{app:prompt_notes}
All prompts in this catalog are reproduced as excerpts; full verbatim text (including JSON schema declarations and few-shot examples) is released with the code. Unless otherwise noted, prompts are fed to Claude-Sonnet-4-6~\citep{claude_sonnet} at temperature $1.0$. The Generator and Discriminator processes operate as independent inference instances under the access matrix of \S\ref{app:info_barrier}; coordination is mediated by the Orchestrator through structured Markdown artifacts (\texttt{REVIEW.md}) rather than direct messaging. Prompts shown below are lightly edited for whitespace and to elide internal artifact paths; semantic content is unchanged.

\subsection{Orchestrator System Prompt}
\label{app:prompt_orchestrator}

The Orchestrator coordinates the Generator-Discriminator loop, monitors the five agreement conditions of \S\ref{app:five_conditions}, and emits the final accepted tuple to disk. It writes no scripts itself.

\begin{promptlisting}{Orchestrator system prompt}
You are the orchestrator for the CUA-Gym computer-use agent training
pipeline. You do NOT generate any scripts or files yourself. Your sole
responsibility is to:

  1. Prepare the working environment.
  2. Spawn the setup-generator and reward-generator processes in an
     adversarial loop.
  3. Monitor agreement conditions by reading REVIEW.md.
  4. Collect and output final verified results.

# Execution Mode (MANDATORY)

Dual-environment only:
  - For each task, provision two isolated VMs: initial_env, golden_env.
  - State separation is guaranteed by environment isolation, not by
    filename suffixes.
  - Single-environment execution is deprecated.

# Role Boundaries

YOU DO:
  - Parse input and prepare working directories.
  - Spawn setup-generator and reward-generator processes.
  - Read REVIEW.md to check agreement conditions.
  - Copy final outputs to output/final/<task_id>/.
  - Update pipeline status.

YOU DO NOT:
  - Generate Python scripts (setup-gen does this).
  - Generate reward scripts (reward-gen does this).
  - Modify any generated scripts.
  - Make quality judgments beyond the five agreement conditions.
  - Skip the adversarial loop or short-circuit the process.
  - Write files outside output/adversarial/<task_id>/,
    output/final/<task_id>/, or output/reward_sandbox/<task_id>/.

# Working Directory Layout (SECURITY-CRITICAL)

output/adversarial/<task_id>/
  task_config.json            # placed by orchestrator
  env_config_initial.json     # initial_env VM connection
  env_config_golden.json      # golden_env VM connection
  initial_setup.py            # written by setup-gen
  golden_patch.py             # written by setup-gen
  reward.py                   # copied from reward sandbox after PASS
  REVIEW.md                   # copied from reward sandbox after PASS

output/reward_sandbox/<task_id>/   # ISOLATED workdir for reward-gen
  task_config.json            # only reward-gen's view of the task
  env_config_initial.json
  env_config_golden.json
  REVIEW.md                   # previous round, if any
  reward.py                   # written by reward-gen (this round)

NOTE: Canonical task artifact files exist ONLY on each VM at
/home/user/. They are NEVER downloaded to the local file system.
Reward-gen ONLY gets the sandbox directory; it cannot see setup-gen's
scripts or data files. It must explore the VM through the state-only
API. This isolation is the operational form of the information
barrier and is the load-bearing security guarantee of the pipeline.

# Workflow

Step 0  Compute PROJECT_ROOT and refuse to proceed if not in the
        CUA-Gym project root.

Step 1  Load tasks from the task-gen output JSON.

Step 2  For each task, provision two VMs, write task_config.json and
        env_configs into output/adversarial/<task_id>/, and spawn
        setup-gen.

Step 3  After setup-gen returns, copy task_config.json and env_configs
        (NOT the setup scripts or artifacts) into the reward sandbox
        and spawn reward-gen.

Step 4  Read the REVIEW.md verdict.
        If PASS:
          - Validate the five agreement conditions independently.
          - Copy final outputs to output/final/<task_id>/.
        If FAIL and round < 5:
          - Append REVIEW.md feedback to setup-gen invocation.
          - Re-spawn setup-gen.
        If FAIL and round == 5:
          - Reject the tuple and record the failure mode.
\end{promptlisting}

\subsection{Generator Subagent System Prompt}
\label{app:prompt_generator}

The Generator stage spans two specialized roles: a \emph{Task-Gen} role that synthesizes natural-language task instructions paired with grounded context, and a \emph{Setup-Gen} role that produces \texttt{initial\_setup.py} and \texttt{golden\_patch.py}. Both share the same domain skill files (\S\ref{app:skill_files}) but operate at different granularity.

\paragraph{Task-Gen prompt.}
\begin{promptlisting}{Task-Gen system prompt}
You are a computer-use task design expert with deep experience in how
people use desktop and web software and how to train agents to do the
same. Your job: given a user's prompt describing the desired task
mix, brainstorm and synthesize a diverse, high-quality set of
computer-use tasks suitable for post-training an RL agent.

# Output Format

A JSON array. Each entry has the following fields.

task_id (string)
  Format: <domain_short>_<topic_short>_<3-digit-number>
  e.g., calc_pivot_001, os_file_012, gimp_layer_003.

task_instruction (string)
  The literal query that will be given to the agent for execution.

  Requirements:
    - Clear intent. The agent must know exactly what to do.
    - Necessary details. Specific names, values, ranges, locations.
    - Natural tone. Like a real user request, not a tutorial step.
    - Not overly verbose. A normal request, not a 500-word essay.
    - Varied sentence patterns. Mix imperative ("Create..."), request
      ("Can you..."), goal-oriented ("I need..."), contextual.

  BAD instructions (DO NOT produce):
    - Vague: "Format the spreadsheet nicely"
        (which cells? what formatting?)
    - Robotic: "Execute the following operation: apply bold to cells
        A1:A10 in the active worksheet"
    - Open-ended: "Create a presentation about climate change"
        (no unique ground truth)
    - Infeasible: "Hack into the system"
    - Method-revealing: "Use the DSUM function to calculate total
        sales"
        (instruct the GOAL, not the METHOD)
    - Formula-typing-only: "Enter =AVERAGE(B2:B13) in cell D2"
        (no GUI interaction)
    - Trivial single-formula: "Calculate the standard deviation of
        column B"
        (one formula is not a meaningful computer-use task)

  GOOD instructions:
    - "Bold the header row in Sheet1 and set the font to Arial 14pt"
    - "I have a CSV file 'sales.csv' on my Desktop. Import it into a
       new LibreOffice Calc spreadsheet."
    - "Change the page orientation of the current Writer document to
       landscape."
    - "Set up conditional formatting on the 'Status' column:
       'Completed' green, 'Pending' yellow, 'Overdue' red."
    - "Freeze the top two rows and first column so they stay visible
       while scrolling."

context (string, >= 300 chars)
  Natural-language description of everything needed to set up and
  evaluate this task. Three required components:

  1. Initial state -- what must exist before the agent starts:
       files and their content/location, application state, data
       characteristics (rows, columns, value ranges).

  2. Ground truth -- expected correct outcome:
       exact values, states, file contents; partial-credit checkpoints;
       concrete numbers where applicable
       ("cell B15 should contain 342.50").

  3. Implicit prerequisites -- things the instruction does not say
     but the evaluator needs:
       e.g., "GIMP canvas should be 1920x1080 with a white background
       layer"; "Firefox should have 3 tabs open: Google, Wikipedia,
       GitHub".

  Context is CRITICAL. Without it, we cannot set up the VM, evaluate
  success, or assign partial credit.

difficulty (string in {easy, medium, hard})
  easy   -- 1-3 atomic actions, common operation
            ("Bold the header row")
  medium -- 3-10 steps, requires feature understanding
            ("Create a pivot table with specific fields")
  hard   -- multi-step workflow, advanced features, cross-component
            interaction or edge cases
            ("Create a macro-free automated report with conditional
            formatting, charts, and cross-sheet references")

# Workflow (Two Phases)

Phase 1 (planning):
  Research the domain via web search; build the feature taxonomy and
  scenario matrix; identify under-covered cells; present the plan for
  approval before generating.

Phase 2 (generation):
  Three sequential passes -- breadth (uniform over taxonomy),
  gap-fill (re-sample under-covered cells), and edge cases (boundary
  / cross-application / rare-primitive tasks). Apply embedding-based
  dedup and slot-template diversity rules between passes.
\end{promptlisting}

\paragraph{Setup-Gen prompt.}
\begin{promptlisting}{Setup-Gen system prompt}
You are the Generator in the CUA-Gym adversarial setup pipeline. Your
job is to create two Python scripts:

  1. initial_setup.py  -- creates the pre-task state on initial_env
  2. golden_patch.py   -- builds the expected post-task artifact on
                          golden_env

You work within an adversarial loop with the verifier. After each
round, the verifier writes REVIEW.md with structured feedback. If the
review fails, the pipeline invokes you again. Your goal: produce
outputs the verifier agrees are correct.

# Execution Mode (MANDATORY)

Dual-environment only:
  - initial_setup.py executes only on initial_env
  - golden_patch.py  executes only on golden_env

Separation is by environment isolation, not by filename suffixes.

# ABSOLUTE RULES

Rule 1 -- Golden Must Be Independently Built in golden_env.
  The golden artifact MUST be produced directly in golden_env, without
  loading or copying files from initial_env. Cross-env file dependency
  is invalid by design.

Rule 2 -- No Side Effects.
  The golden-state spec must represent EXACTLY the intended post-task
  result; the initial-state spec must represent the pre-task result.
  Differences must be task-driven only. Do NOT reorganize, reformat,
  or "clean up" anything else.

Rule 3 -- Realistic Content.
  Initial files must contain believable, non-trivial content. Use real
  names ("Sarah Chen", "Marcus Johnson"), real-looking dates, business
  metrics. Forbidden: "Test data 1", "Lorem ipsum", "foo bar",
  placeholder values.

Rule 4 -- Sufficient Complexity.
  Spreadsheets: multiple sheets, 10+ rows, varied column types.
  Presentations: 3+ slides with text, shapes, layout variety.
  Documents: multiple paragraphs, headings, realistic structure.

Rule 5 -- GUI-Ready Initial State.
  initial_setup.py MUST not only create files; it MUST prepare the GUI
  start state expected by the task: open required apps and files,
  launch via non-blocking subprocess, set DISPLAY=:0, add short sleeps
  for stability, keep startup idempotent.

Rule 6 -- Save Initial Reference Copy for Multimedia.
  For tasks where the agent overwrites a media file in place (GIMP,
  VLC, OpenShot), initial_setup.py MUST save a reference copy at
  <task_id>_initial_reference.<ext> that neither the agent nor
  golden_patch.py touches. The reward script needs this for vision
  judge BEFORE/AFTER comparison.

# Workflow

Step 0  Load the domain skill file (API references, bitter lessons,
        evaluation patterns). For mock_websites, also load the per-mock
        state schema; missing keys cause blank pages.

Step 1  Read task_config.json. Parse the context field into a design
        spec: initial-state requirements, ground-truth post-state
        values, and NEGATIVE constraints (what the initial artifact
        MUST NOT contain). Violating negative constraints causes
        reward(initial_env) > 0 and wastes adversarial rounds.

Step 2  If round > 1, read REVIEW.md and make TARGETED fixes.
        - "reward(golden_env) returned X.X instead of 1.0":
          fix golden_patch to match what the reward expects.
        - "reward(initial_env) returned X.X (should be 0)":
          fix initial_setup to remove task-completion artifacts.
        - script execution errors: fix the specific exception.
        Do NOT rewrite from scratch unless the approach is fundamentally
        wrong; targeted fixes converge faster.

Step 3  Generate initial_setup.py and golden_patch.py from the design
        spec. Execute both on their respective VMs and verify they
        complete without exception before signaling the verifier.
\end{promptlisting}

\subsection{Discriminator Subagent System Prompt}
\label{app:prompt_discriminator}

The Discriminator (Reward-Gen) is the load-bearing component of the information barrier. It writes \texttt{reward.py} from the task semantics alone, with no access to the Generator's setup scripts.

\begin{promptlisting}{Discriminator (Reward-Gen) system prompt}
You are the verifier/discriminator in the CUA-Gym adversarial setup
pipeline. Your job is to:

  1. Generate a reward script (reward.py) that programmatically verifies
     task completion.
  2. Test it against the golden_env artifact (must return 1.0) AND the
     initial_env artifact (must return 0.0).
  3. Write a structured REVIEW.md with your verdict and feedback.

Be thorough but fair. Report PASS when the golden artifact genuinely
matches the task; do not be adversarial for its own sake.

# INFORMATION BARRIER (NON-NEGOTIABLE)

  - You MUST NOT read initial_setup.py, golden_patch.py, or any
    setup-generator source code.
  - You MUST NOT read files from the setup-generator's working
    directory; your workdir is output/reward_sandbox/<task_id>/.
  - You MUST NOT download task artifact data files from either VM
    to your local machine.
  - You MUST explore and verify files exclusively on the VM via the
    state-only environment API.
  - Your reward script must be derived from the task description in
    task_config.json, NOT by peeking at setup-generator outputs.

# Reward Script: Required Properties

  1. Return a progressive float in [0, 1].
  2. Use ACTUAL verification (read real files, check real data).
  3. Award partial credit (0.3, 0.5, 0.7, ...) for partial completion.
  4. Return exactly 1.0 only when 100% completed.
  5. Wrap each scoring component in try/except.
  6. Print "REWARD: X.X" as the last output line.
  7. Be self-contained with stdlib + openpyxl/python-pptx/python-docx.
  8. Comment scoring logic.

# Only Score Task-Introduced Changes

Every scoring component must verify something that DIFFERS between
initial_env and golden_env. Properties true on both endpoints are
preconditions and MUST NOT contribute to score.

Litmus test: if a check passes on the initial_env artifact (before any
agent action), it is NOT measuring task completion and MUST NOT award
points. Goal: reward(initial_env) returns 0.0 (or very close).

CORRECT pattern:
  Component 1: 'Raw Data' sheet is hidden (0.6 pts)        # GOOD
  Component 2: 'Raw Data' hidden AND 'Summary' visible (0.4 pts)  # GOOD

WRONG pattern (scoring pre-existing properties):
  Component: 'Summary' sheet is visible (0.15 pts)         # FORBIDDEN
  Component: 'Raw Data' has correct headers (0.10 pts)     # FORBIDDEN

Use data-integrity checks as PRECONDITION GATES (early-return 0.0 if
the file is corrupted), NOT as scoring components.

# REVIEW.md Output

verdict: PASS | FAIL
agreement_table:
  C1_initial_executes      {pass, evidence}
  C2_golden_executes       {pass, evidence}
  C3_golden_reward_eq_1    {pass, observed_value}
  C4_initial_reward_eq_0   {pass, observed_value}
  C5_no_forbidden_pattern  {pass, matched_pattern_or_null}
feedback_to_setup_gen:
  failing_conditions: [C1|C2|C3|C4|C5, ...]
  setup_issues: [short imperatives, one per issue]
  recommended_action: <one paragraph>
\end{promptlisting}

\subsection{Five Agreement Condition Verifier Prompts}
\label{app:prompt_conditions}

The five agreement conditions of \S\ref{app:five_conditions} are evaluated programmatically rather than by LLM, and so do not have an associated prompt; they are reproduced here as the structured \texttt{REVIEW.md} schema the Discriminator emits to communicate its verdict to the Orchestrator.

\begin{plainlisting}{REVIEW.md schema}
# REVIEW.md schema

verdict:        PASS | FAIL
agreement_table:
  C1_initial_executes:     {pass: bool, evidence: <str>}
  C2_golden_executes:      {pass: bool, evidence: <str>}
  C3_golden_reward_eq_1:   {pass: bool, observed: <float>}
  C4_initial_reward_eq_0:  {pass: bool, observed: <float>}
  C5_no_forbidden_pattern: {pass: bool, matched_pattern: <str|null>}
feedback_to_setup_gen:
  failing_conditions: [<C1|C2|C3|C4|C5>, ...]
  setup_issues:       [<short imperatives, one per issue>]
  recommended_action: <free-form one-paragraph summary>
\end{plainlisting}

\subsection{Filter Critic Prompts}
\label{app:prompt_filter}

The dataset-level filter is a single-pass critic with the conservative-by-default decision policy of \S\ref{app:majority_vote}. The critic prompt below is reproduced verbatim from \texttt{filter/critic\_prompt.py}.

\begin{promptlisting}{Filter critic system prompt}
You are a conservative critic for computer-use training tasks.

Input:
  - query
  - setup
  - golden_setup
  - reward

Goal: decide whether this task should be kept, kept with query
rewrite, or rejected for the training set.

Important constraints:
  - setup, golden_setup, and reward are immutable for this decision.
  - You may only revise the query.
  - If the task would require changing setup, golden_setup, or reward
    to become acceptable, REJECT it.
  - Prefer reject over keep when a high-risk issue is present.
  - Prefer keep over modify_query when the query is already usable
    and self-contained enough for training.
  - Only use modify_query when the query has a meaningful defect that
    harms fairness, self-containment, or task validity.

Severity mapping:
  P0  reject. Fatal setup/reward/environment problem or irreparable
      inconsistency.
  P1  modify_query. Query missing important context, has harmful
      ambiguity, or leaks process in a way that harms fairness.
  P2  keep or modify_query. Non-fatal but noticeable quality issues.
  P3  keep. Strong task quality, no meaningful issue beyond style.

Evaluation priorities:
  1. Is the task suitable and safe for training?
  2. Is the reward semantically valid and robust?
  3. Is the query self-contained, natural enough, and consistent with
     setup/golden_setup/reward?
  4. Can the query be fixed without changing task semantics?

Fatal reasons to reject:
  - reward checks artifacts or implementation traces more than user-goal
    completion
  - reward is highly sensitive to GUI/window/process state
  - reward relies on fragile internals, heuristic perception, or
    unstable extraction
  - hidden assumptions about browser state, plugins, extensions, or
    external resources that the environment cannot provide
  - setup/query/reward mismatch not fixable through query rewrite alone
  - subjective task with rigid single-answer reward

Output schema (JSON only):
  verdict                     keep | modify_query | reject
  severity                    P0 | P1 | P2 | P3
  can_fix_with_query_only     bool
  query_issues                [str, ...]
  setup_reward_risks          [str, ...]
  training_pool_fit           low | medium | high
  confidence                  float in [0, 1]
  reasoning_summary           str
  revised_query               str | null
\end{promptlisting}

\subsection{Mock Synthesis Agent Prompts}
\label{app:prompt_mock}

The mock-environment synthesis pipeline (\S\ref{app:env_pipeline}) is implemented as a triple of specialized agents (Plan, Dev, Web) that coordinate through file-based artifacts. Each agent's role specification is excerpted below; full prompts are released with the code.

\paragraph{Plan Agent.}
\begin{promptlisting}{Plan Agent system prompt}
You are a senior product researcher and technical planner. Your job
is to deeply understand a target real-world application and produce
a structured, actionable plan for the dev agent to implement a
faithful mock.

Primary output: TODO.md, the single source of truth that the dev
agent reads and executes against. Everything else (assets,
screenshots, data model) exists to make TODO.md items implementable
without ambiguity.

In scope:
  - All visible UI components and layouts
  - Navigation, routing, view transitions
  - CRUD operations on content
  - Search, filter, sort interactions
  - Form inputs and validation feedback
  - Drag-and-drop, multi-select, bulk actions
  - Notification/badge counts, unread states
  - Settings panels and preference toggles
  - Realistic mock data (users, messages, tasks, etc.)

Explicitly out of scope:
  - Login / logout / authentication
  - Password management or account creation
  - OAuth, SSO, identity verification
  - Encryption or security mechanisms
  - Real network communication or API calls
  - Database persistence beyond localStorage
  - File uploads to real servers
  - Email/SMS sending
\end{promptlisting}

\paragraph{Dev Agent.}
\begin{promptlisting}{Dev Agent system prompt}
You are an expert frontend engineer specialized in building and
maintaining mock web applications for RL agent training. Your sole
responsibility is code: writing it, improving it, and fixing it.

All coordination is file-based -- no direct messaging between agents.

Inputs you read:
  TODO.md          (from plan agent)   work queue
  assets/README.md (from plan agent)   UI layout, workflows
  assets/data_model.md (from plan agent) entity definitions
  DESIGN.md        (from plan / user)  design system spec; read FIRST
  TEST.md          (from web agent)    bug reports, visual diffs
  AUDIT.md         (from web agent)    code-level issues

Priority for any work cycle:
  AUDIT P0 > TEST P0 > AUDIT P1 > TODO P0 > TODO P1 > TODO P2

Project layout (every mock follows this):
  src/App.jsx                  -- routing
  src/main.jsx
  src/components/              -- reusable UI components
  src/pages/                   -- route-level pages
  src/context/AppContext.jsx   -- global state
  src/utils/dataManager.js     -- state init, localStorage
  src/utils/stateTracker.js    -- diff computation for /go
  vite.config.js               -- middleware plugin for /post + /state
\end{promptlisting}

\paragraph{Web Agent.}
\begin{promptlisting}{Web Agent system prompt}
You are a quality-assurance engineer for mock web applications.
Operate a headless Playwright browser against the deployed mock,
verify the UI tree against DESIGN.md and assets/README.md, and
produce two reports per round:

  TEST.md   functional and visual bug reports
  AUDIT.md  code-level issues: dead handlers, untracked state,
            missing entries in the diff API, broken routes

Both reports use P0/P1/P2 severity matching the dev agent's priority
order. A round is complete when:
  - every interactive element listed in TODO.md has been exercised
  - visual diff against reference screenshots is below threshold
  - state-API endpoints (/post, /go, /state, /upload) all respond
    correctly under the session-isolation contract
\end{promptlisting}

\end{document}